\documentclass{article} % For LaTeX2e
\usepackage{iclr2027_conference,times}

\usepackage{amsmath,amsfonts,bm}

\def\eqref#1{equation~\ref{#1}}
\def\1{\bm{1}}

\DeclareMathAlphabet{\mathsfit}{\encodingdefault}{\sfdefault}{m}{sl}
\SetMathAlphabet{\mathsfit}{bold}{\encodingdefault}{\sfdefault}{bx}{n}

\usepackage{hyperref}
\usepackage{url}

\usepackage{amsmath}
\usepackage{amssymb}
\usepackage{amsthm}
\usepackage{caption}

\usepackage[most]{tcolorbox}

\newtheorem{theorem}{Theorem}[section]
\newtheorem{lemma}[theorem]{Lemma}
\newtheorem*{lemma*}{Lemma}

\newtheorem{proposition}[theorem]{Proposition}
\newtheorem*{proposition*}{Proposition}
\newtheorem{corollary}[theorem]{Corollary}
\newtheorem*{corollary*}{Corollary}

\tcbset{
  theorem box/.style={
    enhanced,
    breakable,
    colback=blue!3,
    colframe=blue!50!black,
    boxrule=0.6pt,
    arc=2mm,
    left=8pt,
    right=8pt,
    top=6pt,
    bottom=6pt,
    before skip=10pt,
    after skip=10pt
  }
}

\usepackage{algorithm}
\usepackage{algpseudocode}

\usepackage[dvipsnames]{xcolor}
\usepackage[table]{xcolor}
\usepackage{pifont}
\newcommand{\cmark}{\textcolor{green!60!black}{\ding{51}}} % green check
\newcommand{\xmark}{\textcolor{Red!20!red}{\ding{55}}}   % red cross
\usepackage{diagbox}

\usepackage{graphicx}
\usepackage{subfigure}
\usepackage{booktabs} % for professional tables
\usepackage{multirow}
\usepackage{wrapfig}
\usepackage{caption}

\title{Symmetry-Breaking De Novo Crystal Generation via Markovian Jump Diffusion}

\author{%
   Van Khoa Nguyen \\
   HES-SO Geneva, \\
   University of Geneva \\
   \texttt{van-khoa.nguyen@etu.unige.ch}
  \And
  Alexandros Kalousis\\
  HES-SO Geneva \\
  \texttt{alexandros.kalousis@hesge.ch}
}

\iclrfinalcopy % Uncomment for camera-ready version, but NOT for submission.
\begin{document}

\maketitle

\begin{abstract}
Generating crystals has recently attracted significant interest due to their broad applications in materials science. However, existing generative models struggle to produce complete crystallographic specifications, limiting their ability to capture full crystal symmetries and structural dependencies. In particular, current state-of-the-art approaches generate crystals only up to site symmetries and rely on sampling space groups from empirical distributions during inference. Inspired by \emph{spontaneous symmetry breaking} in physics, where crystals break symmetries of the underlying system, we propose a novel diffusion-based framework that generates full crystallographic specifications by reversing from the lowest-symmetry priors. Our method leverages a Markovian jump-diffusion process to model these symmetry-breaking dynamics, enabling it to traverse different space groups in a physically motivated manner. Our model, dubbed \emph{Symmetry-breaking Crystal Diffusion} (SbCD), introduces a principled approach to explicitly incorporate inter-space-group transitions into the generative process. In de novo crystal generation experiments, SbCD substantially outperforms its symmetry-preserving counterpart and { performs competitively} on thermodynamic stability on MPTS-52, offering a promising perspective for generative modeling of crystalline materials. {Our code and model weights are publicly available \href{https://github.com/v18nguye/SbCD}{\underline{here}}.}
\end{abstract}

\section{Introduction}

\begin{table}[h]
\centering
\caption{Comparison of crystallographic structure specifications modeled by existing crystal generative frameworks. \cmark denotes the modeling of a crystallographic structure specification, while \xmark denotes its absence. Structure elements enclosed in braces $\{\}$ denote equivalent crystal representations.} %Note that all approaches generate the atom-type feature $\mathbf{A}$; therefore, it is omitted here.
\label{tab:compare}
\resizebox{1.\linewidth}{!}{
\begin{tabular}{l cccc c c}
\toprule
\multirow{2}{*}{Method} & \multicolumn{4}{c}{\textsc{Structure specification}} & \textsc{Asymmetric unit} & \textsc{Symmetry breaking} \\
& $\mathbf{\{k,L\}}$ & $\mathbf{\{F,X\}}$ & $\mathbf{A}$ & $\mathbf{S}$ & & ${G}$ \\
\midrule
CDVAE \citep{xie2021crystal} & \cmark & \cmark & \cmark & \xmark & \xmark & \xmark\\
DiffCSP \citep{jiao2023crystal}, FlowMM \citep{miller2024flowmm} & \cmark & \cmark & \cmark & \xmark & \xmark & \xmark\\
DiffCSP++ \citep{jiao2024space}, SGFM \citep{puny2025space} & \cmark & \cmark & \cmark & \xmark & \xmark & \xmark \\
SymmCD \citep{levy2025symmcd}, SGEquiDiff \citep{chang2026space} & \cmark & \cmark & \cmark & \cmark & \cmark & \xmark \\
\midrule
SbCD (ours) & \cmark & \cmark & \cmark & \cmark & \cmark & \cmark \\
\bottomrule
\end{tabular}
}
\end{table}

% crystal intro in science & acceleration
Crystals play a fundamental role in materials science, underpinning advances across diverse fields including electrochemical energy storage \citep{collins20162d}, semiconductor hardware design \citep{bakaul2016single}, and pharmaceuticals \citep{feig2025self}. Traditionally, discovering a useful crystal with targeted properties has required years or even decades of trial-and-error experimentation \citep{cheetham2022chemical}. Accelerating this discovery process \citep{szymanski2023autonomous,merchant2023scaling} thus holds immense potential for unlocking superior technologies.
% generative models & for crystals
Generative models \citep{kingma2013auto,ho2020denoising,song2020denoising,austin2021structured,lipman2022flow} have emerged as a highly effective paradigm for the in silico screening of structured data with desired properties. Pioneering work by \citet{xie2021crystal} introduced latent diffusion models for raw crystal-structure representations. Subsequent approaches \citep{jiao2023crystal,miller2024flowmm,okhotin2025miad} operate on more compact unit cells and model fractional coordinates on Riemannian manifolds, better respecting the underlying geometry.
% inductive biases
More recently, \citet{jiao2024space} further constrained the generative process to specific space groups, enabling template-conditioned generation of crystals with desirable properties. Notably, \citet{levy2025symmcd} and \citet{chang2026space} proposed learning crystal site symmetries, yielding improved generalization to non-trivial symmetry point groups. Table~\ref{tab:compare} compares these approaches. 

% current limitation scope
%% Wyckoff & space groups
Despite this progress, many methods rely heavily on space groups and Wyckoff positions derived from empirical distributions during generation, limiting their ability to capture full crystal symmetries and structural dependencies. Specifically, \citet{puny2025space} showed that models not explicitly learning Wyckoff positions generalize poorly to unseen configurations. In addition, \citet{chang2025space} invoked Neumann’s principle \citep{neumann1885vorlesungen}, which states that physical properties must be invariant under a crystal’s point group symmetries, underscoring the importance of accurately modeling symmetry through Wyckoff positions and their associated space groups. In this work, we bridge this gap by learning to directly generate complete crystallographic specifications, including Wyckoff positions and space groups, rather than sampling from empirical distributions. 
%% spontaneous symmetry-breaking
To achieve this, we draw inspiration from \emph{spontaneous symmetry breaking} (SSB) in physics, where ordered phases emerge through spontaneous breaking of a higher symmetry. For instance, vanadium dioxide ($\mathrm{VO}_2$) undergoes a reduction in crystal symmetry from the high-symmetry rutile phase ($\mathrm{P4_{2}/mnm}$) to lower-symmetry monoclinic phases such as M1 ($\mathrm{P2_{1}/c}$) or M2 ($\mathrm{C2/m}$) \citep{muller2011symmetryrelationships}. % under varying external conditions \citep{muller2011symmetryrelationships}.
%% jump diffusion
We then leverage Markovian jump diffusion \citep{Feller1949OnTT, campbell2022a} to learn space-group transitions, where forward processes correspond to simulating SSB processes that reduce crystal symmetry to lower space groups, while reverse (generative) processes sample crystals directly from the lowest-symmetry priors.  

% contributions
Here, we summarize our contributions: (i) We theoretically derive a variational bound objective that unifies the structural dependencies among crystal components, in which (ii) we leverage Markovian jump diffusion to model space-group distributions. We then (iii) derive novel symmetry-breaking diffusion processes on continuous and discrete state spaces, which adaptively enforce space-group constraints and admit analytical forms. To model site symmetries, (iv) we propose a simplified representation that facilitates capturing site-symmetry distributions and yields more stable generated crystal structures. Our model, dubbed \emph{Symmetry-breaking Crystal Diffusion} (SbCD), represents the first generative framework capable of generating complete crystallographic specifications.

% structure
The remainder of this paper is organized as follows. Section~\ref{sec:jump} reviews the background on Markovian jump diffusion. Section~\ref{sec:sgawre} revisits existing symmetry-preserving crystal diffusion frameworks under space-group constraints. Building on these foundations, Section~\ref{sec:symbreak} develops a novel theoretical symmetry-breaking generative diffusion framework for crystalline materials. Finally, Section~\ref{sec:exp} presents promising results on de novo crystal generation tasks.

\section{Markovian jump diffusion}
\label{sec:jump}
% continuous time jump diffusion
Many natural phenomena can be modeled via discontinuous Markov processes, such as Poisson processes. These contrast with diffusion processes, a class of continuous-time Markov processes where changes occur incrementally over infinitesimal time intervals \citep{anderson1982reverse, oksendal2003stochastic}. Here, we consider processes that evolve via discrete jumps: the system remains in a given state for a random duration before transitioning abruptly to another state.  
% definition, instantaneous rate, infinitesimal transition probability, total jump (exit) rate.
Following \citet{Feller1949OnTT}, we introduce the instantaneous rate matrix $\mathbf{\Lambda}_t(\widetilde{x},x)$ from state $\widetilde{x}$ to state $x$ as follows:
 % \widetilde{x} appears before x in the forward run as Campell; when h moves -> t moves -> applied Feller
\begin{align}
    \mathbf{\Lambda}_t(\widetilde{x},x) \triangleq \lim_{\Delta t\rightarrow 0} \frac{q_{t|t-\Delta t}(x|\widetilde{x}) -\delta_{\widetilde{x}x}}{\Delta t} = \lambda_t(\widetilde{x})\mathbf{\Pi}_{t}(\widetilde{x},x) - \delta_{\widetilde{x}x}\lambda_t(\widetilde{x})
\end{align} % + for right-hand side limit.
where $q_{t|t-\Delta t}(x|\widetilde{x})$ denotes the infinitesimal transition probability, $\sum_{x} q_{t|t-\Delta t}(x|\widetilde{x})=1$; $\lambda_t(\widetilde{x})$ is the total jump (exit) rate out of state $\widetilde{x}$ at the time $t$, how fast the process leaves the state; $\mathbf{\Pi}_{t}(\widetilde{x},x)$ is the probability that, given a jump from state $\widetilde{x}$  occurs at the time $t$, the process jumps to state $x$, $\mathbf{\Pi}_t(\widetilde{x},\widetilde{x})=0$, denoting where the process goes after jumping; and, $ \mathbf{\Lambda}_t(\widetilde{x},\widetilde{x}) < 0$.    
% forward & backward Kolmogorov equation
Then, the forward and backward Kolmogorov equations for a continuous-time Markov chain are given by \citet{Feller1949OnTT}:
\begin{align}
\label{eq:backford}
    \frac{\partial}{\partial t} q_{t|\widetilde{t}}(x|\widetilde{x})=\sum_{x'}\mathbf{\Lambda}_t(x',x)q_{t|\widetilde{t}}(x'|\widetilde{x}), \quad \frac{\partial}{\partial \widetilde{t}} q_{t|\widetilde{t}}(x|\widetilde{x})=- \sum_{x'}\mathbf{\Lambda}_{\widetilde{t}}(\widetilde{x},x')q_{t|\widetilde{t}}(x|x') 
\end{align}

We now define the time-reversal of the forward process with reverse transition rate $\hat{\mathbf{\Lambda}}_t (x,\widetilde{x})$. \citet{campbell2022a} further derive its relation to the forward transition rate by manipulating the forward and backward equations above, yielding:
\begin{align}
    \hat{\mathbf{\Lambda}}_t (x,\widetilde{x}) = \mathbf{\Lambda}_t (\widetilde{x},x)\frac{q_t(\widetilde{x})}{q_t(x)} = \mathbf{\Lambda}_t (\widetilde{x},x) \sum_{x_0} \frac{q_{t|0}(\widetilde{x}|x_0)}{q_{t|0}(x|x_0)}q_{0|t}(x_0|x) 
\end{align}
% intracibility, non-analytical formula, parametric form
However, computing this reverse transition rate involves marginalizing $q_{0|t}(x_0|x)$ over $x_0$, which is generally intractable. \citet{campbell2022a} proposes estimating this term via a parametric neural network $p^{\theta}_{0|t}(x_0|x)$, and learning the reverse transition rate $\hat{\mathbf{\Lambda}}_t^{\theta} (x,\widetilde{x})$ by optimizing the continuous-time evidence bound, which we formulate in the following lemma:
% continuous-time kl
% lemma, continuous-time ELBO, training with Campbell's framework
\begin{lemma}
\label{lemma:jump}
    Following \citet{campbell2022a}, the negative log-likelihood of a continuous-time jump diffusion process admits the variational upper bound.
    \begin{align}
        &\mathbb{E}_{p_{\mathrm{data}}(x_0)}\left[-\log p_0^{\theta}(x_0)\right] \leq \mathcal{L}_{\mathrm{jump}}(\theta) + \mathrm{Const.}\nonumber\\
        \mathcal{L}_{\mathrm{jump}}(\theta) &\triangleq T\mathbb{E}_{t\sim\mathcal{U}(0,T)\widetilde{x}\sim q_{t|0}(\widetilde{x}|x_0)x\sim\Pi_t(\widetilde{x},x)}\left[-\hat{\mathbf{\Lambda}}_t^{\theta} (\widetilde{x},\widetilde{x}) + \mathbf{\Lambda}_t (\widetilde{x},\widetilde{x}) \log (\hat{\mathbf{\Lambda}}_t^{\theta} (x,\widetilde{x}))\right] \nonumber
    \end{align}
\end{lemma}
% explain each term & tau leaping
where the first term acts as a regularizer that penalizes the total rate of leaving state $\widetilde{x}$, thereby preventing the reverse process from excessive jumping, while the second term maximizes the log-rate of the correct reverse transition pair $(x, \widetilde{x})$, scaled by the total rate of leaving $\widetilde{x}$. Sampling can then be implemented efficiently using $\tau$-leaping algorithms \citep{gillespie2001approximate,wilkinson2018stochastic}. % with the learned reverse rate $\hat{\mathbf{\Lambda}}_t^{\theta} (\cdot)$.
\begin{lemma}
\label{lem:tau}
    Consider a finite-state continuous-time Markov chain with $K$ ordinal states,
    $x_t\in\{1,\dots,K\}$, and learned reverse transition-rate matrix
    $\hat{\mathbf{\Lambda}}_t^\theta$. Assume that for $i\neq j$,
    $\hat{\mathbf{\Lambda}}_t^\theta(i,j)\ge 0$, and $\hat{\mathbf{\Lambda}}_t^\theta(i,i) = -\sum_{j\neq i}\hat{\mathbf{\Lambda}}_t^\theta(i,j)$.
   Then the next state $x_{t-\Delta t}$ after a $\tau$-leap of size $\Delta t$ is updated via $x_{t-\Delta t} = x_t + \sum_{j=1}^K \mathbf{P}_j\,(j - x_t)$, where $\mathbf{P}$ is a vector drawn from  independent Poisson random variables $\mathbf{P}_{j} \sim \mathrm{Poisson}\bigl(\hat{\mathbf{\Lambda}}_t^{\theta}(x_t, j)\,\Delta t\bigr),  \; \forall j \neq x_t$.
\end{lemma}

\section{Space-group-aware crystal generation}
\label{sec:sgawre}
\paragraph{Crystal symmetry and representation.}
 %% define space group, point group
 % The corresponding \textit{point group} is defined as $P \triangleq \{\mathbf{O} | (\mathbf{O}, \mathbf{t}) \in G\}$, which results from applying the homomorphism that projects onto the rotational component.
%% representation & asymmetric unit
To characterize crystal symmetries, we define a \textit{space group} $G \triangleq \{(\mathbf{O}, \mathbf{t}) \in \mathbf{E}(3) \mid \mathbf{x} \mapsto \mathbf{O}\mathbf{x} + \mathbf{t}\}$ as a discrete subgroup of the Euclidean group $\mathbf{E}(3)$ leaving a crystal invariant, and its \textit{point group} $P \triangleq \{\mathbf{O} \mid (\mathbf{O}, \mathbf{t}) \in G\}$ by projecting out translations.  In total, there are 230 space groups and 32 crystallographic point groups, the complete sets of which we denote by $\mathcal{G}$ and $\mathcal{P}$, respectively.
Due to the periodic arrangement of atoms, crystal structures admit multiple equivalent representations. Prior works typically model either the full crystal \citep{xie2021crystal} or its \textit{unit cell} \citep{jiao2023crystal, jiao2024space}. Here, we adopt the more compact \textit{asymmetric unit} \citep{levy2025symmcd}, the minimal non-redundant subset that generates the full crystal via symmetry operations.
Moreover, an asymmetric unit can be mathematically described by the tuple $(\mathbf{F}, \mathbf{A}, \mathbf{L}, \mathbf{S})$, where $\mathbf{F} \in \mathbb{R}^{M\times 3}$ and $\mathbf{A} \in \mathbb{R}^{M \times Z}$ represent the fractional coordinates and atom types (over $Z$ chemical species) of $M$ representative atoms, respectively; $\mathbf{L} \in \mathbb{R}^{3\times 3}$ denotes the crystal lattice, from which the raw Cartesian atomic coordinates can be obtained as $\mathbf{X}=\mathbf{L}\mathbf{F}$;
 and $\mathbf{S} \in \mathcal{P}^{M}$ specifies the site-symmetry group of each atom. These components are used in prior space-group-aware, symmetry-preserving crystal generation methods.

\begin{figure}[t]
\begin{center}
\includegraphics[width=.9\linewidth]{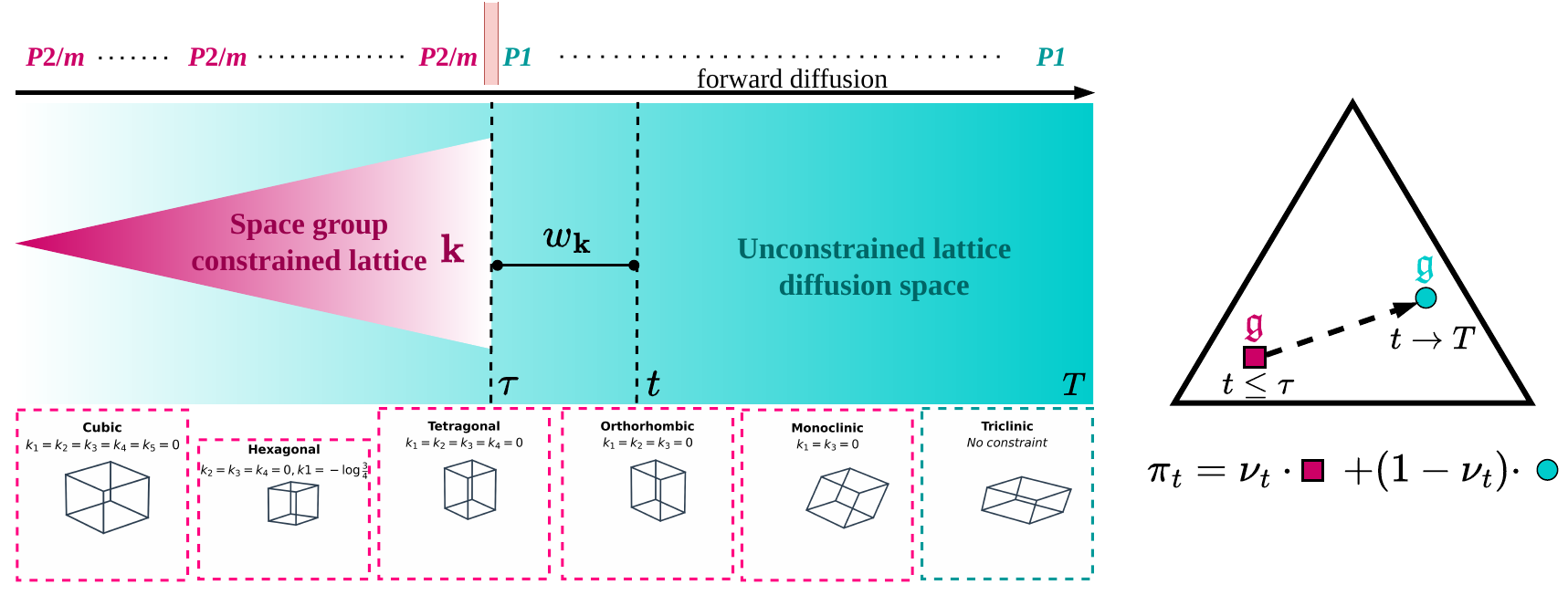}
\end{center}
\caption{SbCD Conceptual Framework. (\textbf{Left}, \emph{continuous state space}) A symmetry-breaking diffusion process that undergoes a reduction in symmetry from a {\color{VioletRed}higher} space group $\color{VioletRed}\mathrm{P2/m}$ to a {\color{BlueGreen}lower} space group $\color{BlueGreen}\mathrm{P1}$ at time step $\tau$. The colored regions correspond to specific space-group constraints applied to the lattice component $\mathbf{k}$. The upper panel depicts space-group transitions, while the lower panel visualizes lattice constraints for six crystal families. (\textbf{Right}, \emph{discrete state space}) For $t>\tau$, the induced site-symmetry prior $\boldsymbol{\pi}_t$ evolves monotonically along the straight line segment connecting $\color{VioletRed}\boldsymbol{\mathfrak{g}}$, the marginal site-symmetry prior of $\color{VioletRed}\mathrm{P2/m}$, and $\color{BlueGreen}\boldsymbol{\mathfrak{g}}$, the marginal site-symmetry prior of $\color{BlueGreen}\mathrm{P1}$.}
\label{fig:main}
\end{figure}

\paragraph{Space group constraints.} 
%% space group constraint on k
The crystal representation can be further constrained depending on its crystal family; the 230 space groups are classified into 6 crystal families. \citet{jiao2024space} derive conditions on the lattice using the polar decomposition \citep{hall2013lie}. Concretely, any invertible matrix $\mathbf{L}$ admits a unique decomposition $\mathbf{L}=\mathbf{T}\exp(\mathbf{U})$, where $\mathbf{T} \in \mathbb{R}^{3 \times 3}$ is orthogonal, and $\mathbf{U} \in \mathbb{R}^{3 \times 3}$ is symmetric. This decomposition yields an \textit{invariant} representation: applying an orthogonal transformation to $\mathbf{L}$ affects only $\mathbf{T}$, leaving $\mathbf{U}$ unchanged. Furthermore, $\mathbf{U}$ can be expanded over a fixed set of symmetric basis matrices $\{\mathbf{B}_i\}_{i=1}^6$ as $\mathbf{U}=\sum_{i=1}^6 k_i \mathbf{B}_i$. Thus, the coefficient vector $\mathbf{k}=(k_i)_{i=1}^6$ provides a complete, invariant lattice representation. %Specific space-group constraints then translate into conditions on $\mathbf{k}$. Following \citet{jiao2024space}, we summarize the construction of $\{\mathbf{B}_i\}$ and the associated constraints on $\mathbf{k}$ in Appendix~\ref{ap:proofk}. We then introduce constrained diffusion processes \citep{ho2020denoising} that respect these lattice conditions through a space-group-informed masking mechanism, as formalized below.
We then introduce a masking mechanism that respects lattice constraints during both forward and reverse diffusion processes: 
% lemma k
\begin{lemma}
    \label{lemma:k}
    Let $\mathbf{k} \in \mathbb{R}^6$ denote the invariant crystal-lattice representation, whose values are constrained by a given crystal family. A discrete-time, continuous-state diffusion process \citep{ho2020denoising} satisfies the masking mechanism defined below to remain within the specified crystal family:
    \begin{align}
        \mathbf{k}_t = \mathfrak{m} \odot \bigl(\sqrt{\bar{\alpha}_t}\,\mathbf{k}_0 + \sqrt{1-\bar{\alpha}_t}\,\boldsymbol{\epsilon}\bigr) + \mathfrak{m_b}, \qquad \boldsymbol{\epsilon} \sim \mathcal{N}(0,I),
    \end{align}
\end{lemma}
where $\bar{\alpha}_t$ denotes the noise schedule from \citet{ho2020denoising}, $\odot$ is the element-wise product, and $(\mathfrak{m}, \mathfrak{m_b}) \in \{0,1\}^6 \times \mathbb{R}^6$ is the space-group-informed mask--bias pair, e.g., $\mathfrak{m} = [0,0,0,0,1,1]$ and $\mathfrak{m_b} = [-\log\tfrac{3}{4}, 0,0,0,0,0]$ correspond to the mask--bias  pair for the hexagonal family (Figure~\ref{fig:main}).

\paragraph{Symmetry preserving.}
%site symmetry, Wynckoff definition.
% A Wyckoff position = an orbit of points. All points in that orbit are symmetry-equivalent, have stabilizers that conjugate -> have the same type of site symmetry. A stabilizer is a set of symmetry ops that leave the site unchanged, stab(x) = {g \in G, gx=x}. Two stabilizers (site symmetry groups) H1 & H2 that are conjugate, if H2 = gH1g^-1.
We characterize the local symmetry at each atomic site via its \textit{site symmetry}, i.e., the set of symmetry operations that leave the site invariant. Sites that are symmetry-equivalent under the action of the space group $G$ form an orbit corresponding to a \textit{Wyckoff position}. The site-symmetry group is a point group isomorphic to a subgroup of the point group $P$.
\citet{levy2025symmcd} encode the site symmetries $\mathbf{S}$ using a binary representation of oriented site-symmetry symbols \citep{hahn1983international}, which specify generators of the site-symmetry group along 15 possible crystallographic axes (including body and face diagonals), with up to 13 distinct symmetry operations per axis. To preserve symmetry within a given space group, they employ categorical diffusion \citep{austin2021structured} with space-group-specific marginal priors over site symmetries during both training and generation.
\begin{lemma}
\label{lemma:S}
    Let $s\in\{1,\cdots,15\}$ index the crystallographic axes, and $\mathbf{s} \in \{0,1\}^{13} \subset \mathbf{S}$ denote the site-symmetry representation along the $s$-th axis. Let $\mathfrak{g} \in [0,1]^{13}$ be the marginal prior of site symmetry associated with the space group $G$ over the $s$-th axis. A discrete-time, discrete-state diffusion process \citep{austin2021structured} that preserves crystal symmetry is defined  via the forward transition:
    $$
    \mathbf{s_t} \sim \mathrm{Cat}(\mathbf{s_t}; \mathbf{s}_0 \mathbf{\bar{Q}}_{s,t}), \qquad \mathbf{\bar{Q}}_{s,t} = (1-\bar{\beta}_t)\mathbf{I} + \bar{\beta}_t \mathbf{1}\mathfrak{g}
    $$
\end{lemma}
% do not leave the site symmetry of the space group
where $\bar{\beta}_t$ denotes the categorical noise schedule from \citet{austin2021structured}. By construction, this space-group-conditioned prior ensures that generated site-symmetry states are consistent with the symmetry constraints imposed by $G$, which consequently restricts the candidate Wyckoff positions of representative atoms. If multiple candidates share the same site-symmetry state, we assign the Wyckoff position whose symmetry-equivalent sites are closest to generated fractional coordinates.

\section{SbCD: Symmetry-breaking crystal diffusion}
\label{sec:symbreak}
\subsection{Structure-informed variational bound}
% structure challenging & dependent assumption 
Crystals are inherently heterogeneous structures comprising multiple data modalities. This challenge necessitates moving beyond a single generative modeling paradigm or data representation. Here, we unify diffusion-based training objectives across continuous and discrete time and state spaces into a single formulation. We recall the parametrization of an asymmetric unit as $\mathcal{M} = (\mathbf{F}, \mathbf{A}, \mathbf{k}, \mathbf{S}, G)$. As defined in Section~\ref{sec:sgawre}, the site symmetries $\mathbf{S}$ and the lattice representation $\mathbf{k}$ are explicitly constrained by the space group $G$. Under these structural dependencies, we formulate the variational upper bound.
% Proposition KL decomposition
\begin{proposition}
\label{prop:path_kl_decomp}
Let $\mathbb{Q}$, $\hat{\mathbb{Q}}$, and $\mathbb{P}^{\theta}$ denote the forward, exact reverse, and learned reverse path measures of the continuous-time Markov chain. The negative log-likelihood on crystal structures $\mathcal{M}_0$ is bounded above by the expected Kullback--Leibler divergence over the reverse trajectory:
\begin{equation}
    -\log p_{\theta}(\mathcal{M}_0) \leq \mathbb{E}_{q_{T|0}}\Bigl[ D_{\mathrm{KL}}\bigl( \hat{\mathbb{Q}}^{\mathcal{M}_T \rightarrow \mathcal{M}_0} \,\big\|\, \mathbb{P}^{\theta, \mathcal{M}_T} \bigr) \Bigr] + \mathrm{Const.} \nonumber
\end{equation}
The divergence decomposes additively into space group, lattice, site symmetry, atom type, and fractional coordinate components:
\begin{align}
    D_{\mathrm{KL}}\bigl( \hat{\mathbb{Q}}^{\mathcal{M}_T \rightarrow \mathcal{M}_0} \,&\big\|\, \mathbb{P}^{\theta, \mathcal{M}_T} \bigr) = \underbrace{\colorbox{white!10}{$ D_{\mathrm{KL}}\bigl( \hat{\mathbb{Q}}_{\mathbf{A}} \,\big\|\, \mathbb{P}^{\theta}_{\mathbf{A}} \bigr) 
    + D_{\mathrm{KL}}\bigl( \hat{\mathbb{Q}}_{\mathbf{F}} \,\big\|\, \mathbb{P}^{\theta}_{\mathbf{F}} \bigr)$}}_{\textsc{I}:~\text{Atom Types \& Fractional Coordinates}} \nonumber \\
    &+ \underbrace{\colorbox{white!20}{$ D_{\mathrm{KL}}\bigl( \hat{\mathbb{Q}}_{G} \,\big\|\, \mathbb{P}^{\theta}_G \bigr)$}}_{\textsc{II}:~\text{Space Group}}
    + \underbrace{\colorbox{white!20}{$ \mathbb{E}_{\hat{\mathbb{Q}}_G}\Bigl[ 
        D_{\mathrm{KL}}\bigl( \hat{\mathbb{Q}}_{\mathbf{k}\mid G} \,\big\|\, \mathbb{P}^{\theta}_{\mathbf{k}\mid G} \bigr) 
        + D_{\mathrm{KL}}\bigl( \hat{\mathbb{Q}}_{\mathbf{S}\mid G} \,\big\|\, \mathbb{P}^{\theta}_{\mathbf{S}\mid G} \bigr)
    \Bigr]$}}_{\textsc{III}:~\text{Lattice \& Site Symmetry}} \nonumber
\end{align}
\end{proposition}
% explain each KL term, highlight the contribution to G
Here, the conditioning superscripts have been omitted on the right-hand side for brevity. The notation $\hat{\mathbb{Q}}^{\mathcal{M}_T\rightarrow\mathcal{M}_0}_{\cdot}$ denotes a bridge measure pinned at both endpoints $\mathcal{M}_T$ and $\mathcal{M}_0$, and $\mathbb{P}^{\theta, \mathcal{M}_T}_{\cdot}$ denotes a path measure conditioned on $\mathcal{M}_T$. While most existing de novo crystal generation frameworks neglect learning the space-group distribution (term~II) and assume it is sampled from data during inference, our method explicitly models $G$ as part of the generative process. Moreover, during the diffusion process, the space group transitions, resulting from term~II, directly modify the constraints imposed on the lattice and site symmetry (term~III), which we address in the following sections. Finally, the space-group-independent attributes (term~I) can be modeled using standard techniques from prior work of~\citet{jiao2023crystal, austin2021structured, levy2025symmcd}.

\subsection{Physics-inspired space group modeling via jump diffusion}
% jump diffusion & simplest scenario & minimize the first kl term
Taking inspiration from \emph{spontaneous symmetry breaking}, in which physical systems spontaneously break symmetry into lower-symmetry states, we learn the space-group distribution by mirroring this phenomenon via Markovian jump-diffusion. In the forward process, a crystal remains in a given space group ${\widetilde{G}}$ for a random duration before transitioning to a lower-symmetry space group $G$, denoted by ${G} \prec {\widetilde{G}}$. The transition dynamics are governed by a time-dependent rate matrix $\mathbf{\Lambda}_t({\widetilde{G}}, {G})$. The generative process approximates the reverse transitions by learning a parametric rate $\hat{\mathbf{\Lambda}}_t^{\theta}({G},{\widetilde{G}})$. For simplicity, we fix the terminal state of every jump to $G = \mathrm{P1}$, the lowest-symmetry space group (triclinic system) that contains only the identity operation. In the following proposition, we introduce the transition rate matrix such that (i) the forward process mixes rapidly toward a reference distribution and (ii) the Kolmogorov forward equation admits a closed-form solution.

% Proposition on G & transition matrix & commutative property & forward kernel 
\begin{proposition}
\label{prop:jump}
Consider the discrete state space of $230$ crystallographic space groups drawn from the six crystal families. Let $\mathbf{\Pi}, \mathbf{\Lambda}_t \in \mathbb{R}^{230 \times 230}$ denote the time-invariant transition probability matrix and the instantaneous rate matrix, respectively, defined element-wise as follows:
\begin{align}
\mathbf{\Pi}(\widetilde{G},G)=
\begin{cases}
0 & \widetilde{G} = \mathrm{P1} \\
0 & G = \widetilde{G} \\
1 & G = \mathrm{P1} \neq  \widetilde{G} \\
0 & \text{otherwise}.
\end{cases}
\qquad
\mathbf{\Lambda}_t(\widetilde{G},G)=
\begin{cases}
0 & \widetilde{G} = \mathrm{P1} \\
-\lambda(t) & G = \widetilde{G} \\
\lambda(t) & G = \mathrm{P1} \neq \widetilde{G} \\
0 & \text{otherwise}. \nonumber
\end{cases}
\end{align}
   
Then, $\mathbf{\Lambda}_t$ and $\mathbf{\Lambda}_s$ commute for all $t,s$. Consequently, the forward transition probability of being in state $G$ at time $t$, given state $\widetilde{G}$ at time $\widetilde{t} < t$, $q_{t|\widetilde{t}}(G|\widetilde{G})$, admits the matrix exponential form:
\begin{align}
q_{t|\widetilde{t}}(G|\widetilde{G}) = \left(\mathbf{D}\exp\Bigr[\mathbf{\Sigma}\int_{\widetilde{t}}^t\lambda(s)ds\Bigl]\mathbf{D}^{-1}\right)_{\widetilde{G},G} \quad \text{with} \quad
\frac{\mathbf{\Lambda}_t}{\lambda(t)} =
\mathbf{D}\mathbf{\Sigma}\mathbf{D}^{-1} \nonumber
\end{align}
\end{proposition}
We defer the proof to the Appendix. Given the analytical forward jump kernel above, we learn the reverse rate $\hat{\mathbf{\Lambda}}_t^{\theta}({G},{\widetilde{G}})$ by minimizing the variational upper bound component corresponding to $D_{\mathrm{KL}}(\mathbb{\hat{Q}}_{G}^{\mathcal{M}_T\rightarrow\mathcal{M}_0}\|\mathbb{P}^{\theta, \mathcal{M}_T}_G)$ in Proposition~\ref{prop:path_kl_decomp}.
\paragraph{Posterior holding-time $\tau$ and symmetry-breaking time window $w$.}
Optimising the jump diffusion loss $\mathcal{L}_{\mathrm{jump}}(\theta)$ via Lemma~\ref{lemma:jump} first requires sampling a diffusion timestep $t \sim \mathcal{U}(0, T)$, followed by sampling the corresponding space group according to the analytical forward kernel $q_{t|0}$ from Proposition~\ref{prop:jump}. For large diffusion times, $t \rightarrow T$, the forward process has already reached the space group $\mathrm{P1}$ and remains there thereafter. In the following corollary, we define the posterior holding time $\tau$ as the duration for which the process remains in its initial space group before jumping, and the symmetry-breaking time window $w$ as the elapsed time spent in the space group $\mathrm{P1}$ up to the current diffusion timestep. Figure~\ref{fig:main} depicts these induced time variables.
\begin{corollary}
\label{cor:tau}
Let a crystal initially reside in a space group $\widetilde{G}$ and be observed in a lower-symmetry space group $G \prec \widetilde{G}$ at time $t$. Consider a constant transition rate, $\lambda(t) = \lambda \; \forall t$, then the posterior distribution of the holding time $\tau \in [0, t]$ is $\tau = -\frac{1}{\lambda}\log(1 - u(1 - e^{-\lambda t}\bigr)), \;\text{where} \; u \sim \mathcal{U}(0,1)$. Consequently, the symmetry-breaking time window can be calculated as $w = t - \tau$.
\end{corollary}

\subsection{Symmetry-breaking diffusion in continuous-state space}
We extend the space-group-constrained diffusion framework of \citet{jiao2024space} to learn the crystal-lattice distribution. In their framework, the space group is fixed during both forward and reverse diffusion, and an element-wise masking mechanism on the lattice representation $\mathbf{k}$ (see Lemma~\ref{lemma:k}) ensures it remains confined to the space-group-constrained subspace. In contrast, our framework permits the space group to evolve throughout the diffusion process. We therefore introduce a symmetry-breaking diffusion process over the continuous-state lattice space, which adaptively enforces the lattice constraints of the prevailing crystal family whenever the process transitions to a new space group. Below, we provide the analytical forms of the forward and reverse sampling kernels, which correspond to the variational bound term $\mathbb{E}_{\hat{\mathbb{Q}}_G}\bigl[ D_{\mathrm{KL}}\bigl( \hat{\mathbb{Q}}_{\mathbf{k}\mid G} \,\big\|\, \mathbb{P}^{\theta}_{\mathbf{k}\mid G} \bigr) \bigr]$ in Proposition~\ref{prop:path_kl_decomp}.
\begin{proposition}
\label{prop:k}
Consider a forward diffusion process with a symmetry-breaking event at time $\tau$, transitioning from space group $\widetilde{G}$ to $G$. From Lemma~\ref{lemma:k}, let $(\mathfrak{m}, \mathfrak{m_b})$ and $(\widetilde{\mathfrak{m}}, \mathfrak{\widetilde{m}_b})$ denote the mask--bias pairs associated with $G$ and $\widetilde{G}$, respectively. Then, the continuous-state forward diffusion kernel applied to the lattice parameters $\mathbf{k}$ is given analytically by:
\begin{align}
    \mathbf{k}_t = \widetilde{\mathfrak{m}}\odot&\left(\sqrt{\bar{\alpha}_t}\,\mathbf{k}_0+\sqrt{1-\bar{\alpha}_t}\,\widetilde{\boldsymbol{\epsilon}}\right) +\mathfrak{\widetilde{m}_b} + \mathfrak{m}\odot\mathfrak{\widetilde{m}_b}\sqrt{\frac{\bar{\alpha}_t}{\bar{\alpha}_{t-w_{\mathbf{k}}}}}\nonumber \\
    &-\left(\widetilde{\mathfrak{m}}\odot\sqrt{1-\frac{\bar{\alpha}_t}{\bar{\alpha}_{t-w_{\mathbf{k}}}}}\,\widetilde{\boldsymbol{\epsilon}} + \mathfrak{\widetilde{m}_b}\right) + \left(\mathfrak{m}\odot\sqrt{1-\frac{\bar{\alpha}_t}{\bar{\alpha}_{t-w_{\mathbf{k}}}}}\,\boldsymbol{\epsilon} + \mathfrak{m_b} \right), \nonumber
\end{align}
where $w_{\mathbf{k}} = t - \tau$.
The corresponding reverse-time sampling step is simulated as:
\begin{align}
    \mathbf{k}_{t-1} = \widetilde{\mathfrak{m}}\odot\Bigl(\frac{\sqrt{\bar{\alpha}_{t-1}}}{1-\bar{\alpha}_t}(1-\alpha_t)\,\hat{\mathbf{k}}^{\theta}_0(\mathcal{M}_t) + \mathfrak{m}\odot\frac{\sqrt{\alpha_t}}{1-\bar{\alpha}_t}(1-\bar{\alpha}_{t-1})\,\mathbf{k}_t + \mathfrak{m_b} + \sigma_t\boldsymbol{\epsilon}\Bigr) + \mathfrak{\widetilde{m}_b}. \nonumber
\end{align}
%
% with $\sigma_t=\sqrt{(1-\bar{\alpha}_{t-1})(1-\alpha_t)/(1-\bar{\alpha_t})}$ following the parameterization of \citet{song2020denoising}.
%
\end{proposition}
This novel masking mechanism respects the lattice constraints both before and after symmetry breaking occurs. Moreover, our formulation directly generalizes the prior work of \citet{jiao2024space}, which corresponds to the special case of no symmetry breaking in Lemma~\ref{lemma:k}, i.e., $\widetilde{G}=G$. Figure~\ref{fig:main} (left) illustrates how lattice constraints adaptively change during a space-group transition.

\subsection{Symmetry-breaking diffusion in discrete-state space}
\paragraph{Novel site-symmetry representation learning.} \citet{levy2025symmcd} proposed learning site symmetries by using a binary matrix of size $15 \times 13$ ($15$ canonical symmetry axes and $13$ possible symmetry operations per axis; see Lemma~\ref{lemma:S}), based on the oriented site-symmetry symbols \citep{hahn1983international}. However, this encoding scheme does not distinguish Wyckoff positions sharing the same oriented site-symmetry symbol. Due to this ambiguity, the scheme may complicate the learning process and often requires projecting generated site symmetries onto the nearest valid point group \citep{levy2025symmcd}. For instance, in space group $\mathrm{Immm}$, the two Wyckoff positions $\mathrm{4i}$ and $\mathrm{4j}$ both possess the oriented site-symmetry symbol $\mathrm{mm2}$ and are therefore represented by the same $15 \times 13$ binary matrix. Here, we directly use a one-hot vector representation to model the oriented site-symmetry symbols. We identify $81$ distinctive oriented site-symmetry symbols, inferred across all $230$ space groups (see Appendix~\ref{app:orient_ss} for the list).

% diffusion symmetry breaking
In the \emph{symmetry-preserving} framework of \citet{levy2025symmcd}, for a fixed space group, site-symmetry states are corrupted via a discrete-state diffusion process \citep{austin2021structured}, with transition dynamics defined by the marginal site-symmetry prior induced by that space group (Lemma~\ref{lemma:S}). In contrast, in our symmetry-breaking process, the ambient space group itself may transition from one group to another. This change alters the induced marginal prior over site symmetries, and therefore requires an interpolation between site-symmetry priors across space groups. In the following proposition, we introduce such an interpolant to enable learning of site symmetries in the symmetry-breaking diffusion process.
\begin{proposition}
\label{prop:marginal_s}
  Let $\mathfrak{\widetilde{g}}, \mathfrak{g} \in \Delta^{81-1}$ be the marginal site-symmetry priors associated with the space groups $\widetilde{G}$ and $G$, respectively. Let $\bar{\alpha}_t$ be a strictly decreasing noise schedule with symmetry-breaking time $\tau$ satisfying $0<\bar{\alpha}_t, \bar{\alpha}_{\tau} \leq 1$. For $t>\tau$, define
    \begin{align}
        \boldsymbol{\pi}_t = \nu_t\mathfrak{\widetilde{g}} + (1-\nu_t)\mathfrak{g}=\frac{\bar{\alpha}_t(1-\bar{\alpha}_{t- w_{\mathbf{S}}})\mathfrak{\widetilde{g}} + (\bar{\alpha}_{t- w_{\mathbf{S}}}-\bar{\alpha}_t)\mathfrak{g}}{\bar{\alpha}_{t- w_{\mathbf{S}}} (1- \bar{\alpha}_t)} \mathrm{\quad where\;} w_{\mathbf{S}}=t -\tau,\; \nu_t \in [0,1] \nonumber
    \end{align}
    Then, $\boldsymbol{\pi}_t$ lies in the convex hull of the two priors, $\boldsymbol{\pi}_{\tau}=\mathfrak{\widetilde{g}},\; \lim_{t\rightarrow T}\boldsymbol{\pi}_{t}=\mathfrak{g}$ , and the trajectory $\{\boldsymbol{\pi}_t\}_{t>\tau}$ moves monotonically along the straight line segment connecting $\mathfrak{\widetilde{g}}$ to $\mathfrak{g}$ inside the simplex.
\end{proposition}

% Fig1. train and sampling
Figure~\ref{fig:main} (right) visualizes the interpolated prior trajectory. Below, we derive corresponding forward and reverse discrete-state diffusion kernels and summarize the sampling of SbCD in Algorithm~\ref{alg:sampleSbCD}.
\begin{corollary}
\label{coro:s}
Let the discrete-state diffusion transition matrices of \citet{austin2021structured} be redefined under the symmetry-breaking formulation as:
\begin{align}
\mathbf{Q}_{t} = \alpha_tI + (1-\alpha_t)\mathbf{1}(\mathfrak{g})^T, \qquad
 \mathbf{\bar{Q}}_{t-1} = \bar{\alpha}_{t-1}I + (1-\bar{\alpha}_{t-1})\mathbf{1}(\mathfrak{\widetilde{g}})^T \nonumber\\
 \bar{\mathbf{Q}}_{t}^{\widetilde{G} \to G} = \bar{\alpha}_tI + (1-\bar{\alpha}_t)\mathbf{1}(\boldsymbol{\pi}_t)^T \mathrm{\; with\;} \mathfrak{g}, \mathfrak{\widetilde{g}}, \boldsymbol{\pi}_t \mathrm{\; in\; Proposition~\ref{prop:marginal_s}\;} \nonumber
\end{align}
Then, the forward and reverse sampling processes on the site symmetry $\mathbf{S}$ can be derived as follows:
\begin{align}
\mathbf{S}_{t} \sim \mathrm{Cat}(\mathbf{S}_{t}; \mathbf{S}_{0}\bar{\mathbf{Q}}_{t}^{\widetilde{G} \to G}) \qquad \mathbf{S}_{t-1} \sim \mathrm{Cat}(\mathbf{S}_{t-1}; \frac{\mathbf{S}_{t}\mathbf{Q}^T_{t}\odot\mathbf{\hat{S}}_{0}^{\theta}(\mathcal{M}_t)\mathbf{\bar{Q}}_{t-1}}{\mathbf{\hat{S}}_{0}^{\theta}(\mathcal{M}_t)\bar{\mathbf{Q}}_{t}^{\widetilde{G} \to G} \mathbf{S}_{t}^T}) \nonumber 
\end{align}
\end{corollary}

\section{Experiments}
\label{sec:exp}

\paragraph{Task and metrics.}
Following \citet{levy2025symmcd}, we conduct de novo crystal generation experiments on the MP-20 dataset, a subset of the Materials Project \citep{10.1063/1.4812323}, and the more challenging MPTS-52 dataset \citep{baird2024matbench}, which contains up to 52 atoms per primitive unit cell. We evaluate our results using standard metrics: (i) \emph{Validity}, including structure ($\mathrm{Struct.}$), requiring a minimum pairwise atomic distance of $0.5$\r{A}, and composition ($\mathrm{Comp.}$), requiring overall charge neutrality; (ii) \emph{Coverage}, measuring similarity between generated and test materials with respect to structure and composition cutoff distances; and (iii) \emph{Property statistics}, measuring the Wasserstein-1 distance between property distributions, including density ($\mathrm{d_{\rho}}$, unit $\mathrm{g/cm^3}$) and the number of unique elements ($\#\mathrm{elem.}$), and the Jensen Shannon distance between space group distributions ($\mathrm{d_{sg}}$). Finally, we assess the thermodynamic stability ($\mathrm{Stable}$) of generated materials and compute their novelty and uniqueness ($\mathrm{S.U.N}$), following \citet{miller2024flowmm}. As in \citet{levy2025symmcd}, we employ the pretrained CHGNet model \citep{deng2023chgnet} to perform structural relaxations before calculating the energy above the hull ($E^{\text{hull}}$). We evaluate all methods on 10,000 sampled crystals, except that property statistics are computed on 1,000 valid structures, following prior works \citep{xie2021crystal, levy2025symmcd}. We defer the experimental details to Appendix~\ref{app:exp}.

\paragraph{Baselines.} 
We benchmark SbCD against diffusion-based methods, including CDVAE~\citep{xie2021crystal}, { UniMat~\citep{yang2024scalable}, MatterGen~\citep{zeni2025generative}, ADiT~\citep{joshi2025all}}, DiffCSP~\citep{jiao2023crystal}, DiffCSP++~\citep{jiao2024space}, and SGEquiDiff~\citep{chang2025space} as well as flow-based methods, including FlowMM~\citep{miller2024flowmm}, SGFM~\citep{puny2025space}, { MCFlow~\citep{seong2026multimodal}, and SymADiT~\citep{karmush2026symadit}}. {We also include two recent state-of-the-art LLM-based baselines, FlowLLM~\citep{sriram2024flowllm} and CrysLLMGen~\citep{khastagir2025llm}.} In particular, we compare SbCD with its symmetry-preserving counterpart, SymmCD~\citep{levy2025symmcd}. We classify baselines by using raw-crystal ($\mathrm{R.}$), unit-cell ($\mathrm{Unit.}$), or asymmetric-unit ($\mathrm{Asym.}$) representations. 

In addition, we ablate various SbCD variants: (1) $\text{SbCD}^{\sharp}$, which learns site symmetries using the $15 \times 13$ binary matrix introduced by \citet{levy2025symmcd}(as described in Lemma~\ref{lemma:S}); (2) $\text{SbCD}^{\diamondsuit}$, which has a smaller model architecture similar to that of DiffCSP/++~\citep{jiao2023crystal, jiao2024space}; (3)  $\text{SbCD}^{\diamondsuit}$ (w/o SB), a non symmetry-breaking $\text{SbCD}^{\diamondsuit}$ ablation, which explicitly conditions on empirical space groups in both training and inference.  

\begin{table}[t]
\centering
\caption{De novo crystal generation on MP-20.  Baseline results are taken from  \citet{levy2025symmcd}, \citet{puny2025space}, { and \citet{khastagir2025llm}. Bold indicates the best result. Highlighting indicates the better-performing method between SbCD and its symmetry-preserving counterpart, SymmCD.} Hyphens ($-$) indicate unreproducible results.}
\label{tab:res}
\resizebox{.95\linewidth}{!}{
\begin{tabular}{ll cc cc| cc| cc | cc}
\toprule
& \multirow{2}{*}{Method} & \multicolumn{2}{c}{\textbf{Validity} (\%)($\uparrow$)} & \multicolumn{2}{c}{\textbf{Coverage} (\%)($\uparrow$)} & \multicolumn{2}{c}{\textbf{Property statistics} ($\downarrow$)} & \multicolumn{2}{c}{\textbf{CHGNet} $E^{\text{hull}} < 0$ (\%)($\uparrow$)} & \multicolumn{2}{c}{\textbf{CHGNet} $E^{\text{hull}} < 0.1$ (\%)($\uparrow$)}\\
\cmidrule(lr){3-6} \cmidrule(lr){7-8} \cmidrule(lr){9-10} \cmidrule(lr){11-12}
& & $\mathrm{Struct.}$ & $\mathrm{Comp.}$ & $\mathrm{Recall}$ & $\mathrm{Precision}$ & $\mathrm{d_{\rho}}$ &  $\#\; \mathrm{elem.}$ & $\mathrm{Stable}$ & $\mathrm{S.U.N}$ &  $\mathrm{MetaStable}$ &  $\mathrm{M.S.U.N}$ \\
\midrule
\midrule
\multirow{1}{*}{\rotatebox{90}{$\mathrm{R.}$}} & CDVAE & $99.97$  & $85.61$ & $99.31$ & $99.47$ & $0.70$ & $1.28$ &  $4.42$ & $4.26$ & $23.58$ & $21.99$ \\
\midrule
\multirow{11}{*}{\rotatebox{90}{$\mathrm{Unit.}$}} & DiffCSP & $97.43$ & $82.50$ & $99.55$ & $98.73$ & $0.18$ & $0.56$ & $ 11.32$ & $8.92$ & $35.04$ & $32.19$\\
& FlowMM & $96.67$ & $83.25$ & $99.49$ & $99.58$ & $0.23$ & $0.08$ & $9.05$ &$6.49$ & $31.64$ & $22.46$\\
 &   MatterGen &  $\mathbf{100}$ &   $86.34$ &   $99.59$  & $99.45$  &   $0.459$ &  $0.254$ &   $-$ &  $-$ & $-$ & $-$\\
  &   UniMat &    $97.2$ &   $89.4$ &    $99.8$ &  $99.7$ &   $0.088$ &  $0.056$ &   $-$ &  $-$ & $-$ & $-$\\
&   ADiT &    $99.58$ &   $90.46$ &    $-$ &  $-$ &   $-$ &  $-$ &   $-$ &  $-$ & $-$ & $-$\\
\cmidrule(lr){2-12}
& DiffCSP++ (empirical) & $99.44$ & $86.50$ & $99.72$ & $99.61$ & $0.12$ & $0.33$ & $11.36$ &$8.62$ & $42.39$ & $30.56$\\
& DiffCSP++ (Wyformer) & $99.66$ & $80.34$ & $-$ & $-$ & $0.67$ & $0.10$ & $-$ & $-$ & $-$ & $-$\\
& SGFM (empirical) & $99.87$ & $86.81$ & $-$ & $-$ & $\mathbf{0.075}$ & $0.181$ & $-$ &$-$ & $-$ & $-$\\
& SGFM (Wyformer) & $99.87$ & $84.76$ & $-$ & $-$ & $0.24$ & $0.23$ & $-$ & $-$ & $-$ & $-$\\
\cmidrule(lr){2-12}
 &    FlowLLM (70B)&   $99.94$ &   $90.84$ &    $96.95$  & $99.82$  &   $1.142$ &  $0.150$ &   $ -$ &  $-$ & $-$ & $-$\\
 &   CrysLLMGen (7B) &    $99.94$ &   $\mathbf{93.55}$ &    $98.52$ &  $99.84$ &   $0.972$ &  $0.272$ &   $\mathbf{16.79}$ &  $\mathbf{9.21}$ & $\mathbf{62.02}$ & $\mathbf{35.94}$\\
\midrule
\multirow{6}{*}{\rotatebox{90}{$\mathrm{Asym.}$}}
& SymmCD & \cellcolor{Green!30} $90.34$ & $85.81$ & $99.58$ & $97.76$ & $0.23$ & $0.40$ & $9.34$ &$6.86$ & $40.01$ & $31.69$\\
& $\text{SymmCD}^{\ddag}$ (retrain) & $87.37$ & $86.30$ & $99.46$ & $97.21$ & $0.41$ & $0.28$ & $7.40$ &$6.79$ & $33.54$ & $30.61$\\
& SbCD & $89.80$ & \cellcolor{Green!30} $90.32$ & $99.44$ & $97.91$ & $0.18$ & \cellcolor{Green!30} $\mathbf{0.03}$ & \cellcolor{Green!30} $11.28$ & $7.15$ & \cellcolor{Green!30} $59.74$ &  $34.00$\\
& $\text{SbCD}^{\sharp}$ & $ 89.81$ & $88.32$ & $99.51$ & $97.31$ & \cellcolor{Green!30} $0.09$ & $0.04$ & $10.74$ & $7.31$ & $56.25$ & $35.66$\\
\cmidrule(lr){2-12}
&  $\text{SbCD}^{\diamondsuit}$& $88.25$ & $87.47$ & $99.44$ & $97.71$ &  $0.16$ & $0.07$ &  $10.23$ & \cellcolor{Green!30} $7.67$ & $50.91$ & \cellcolor{Green!30} $35.82$\\
&  $\text{SbCD}^{\diamondsuit}$(w/o SB)& $78.22$ & $84.93$ & $99.76$ & $96.33$ &  $0.15$ & $0.21$ & $8.90$ & $6.19$ & $44.87$ & $28.49$\\
\bottomrule
\end{tabular}
}
% \vspace{-20pt}
\end{table}

\begin{minipage}[t]{0.5\textwidth}
 % \vspace{-21pt}
\begin{center}
        \includegraphics[width=.8\linewidth]{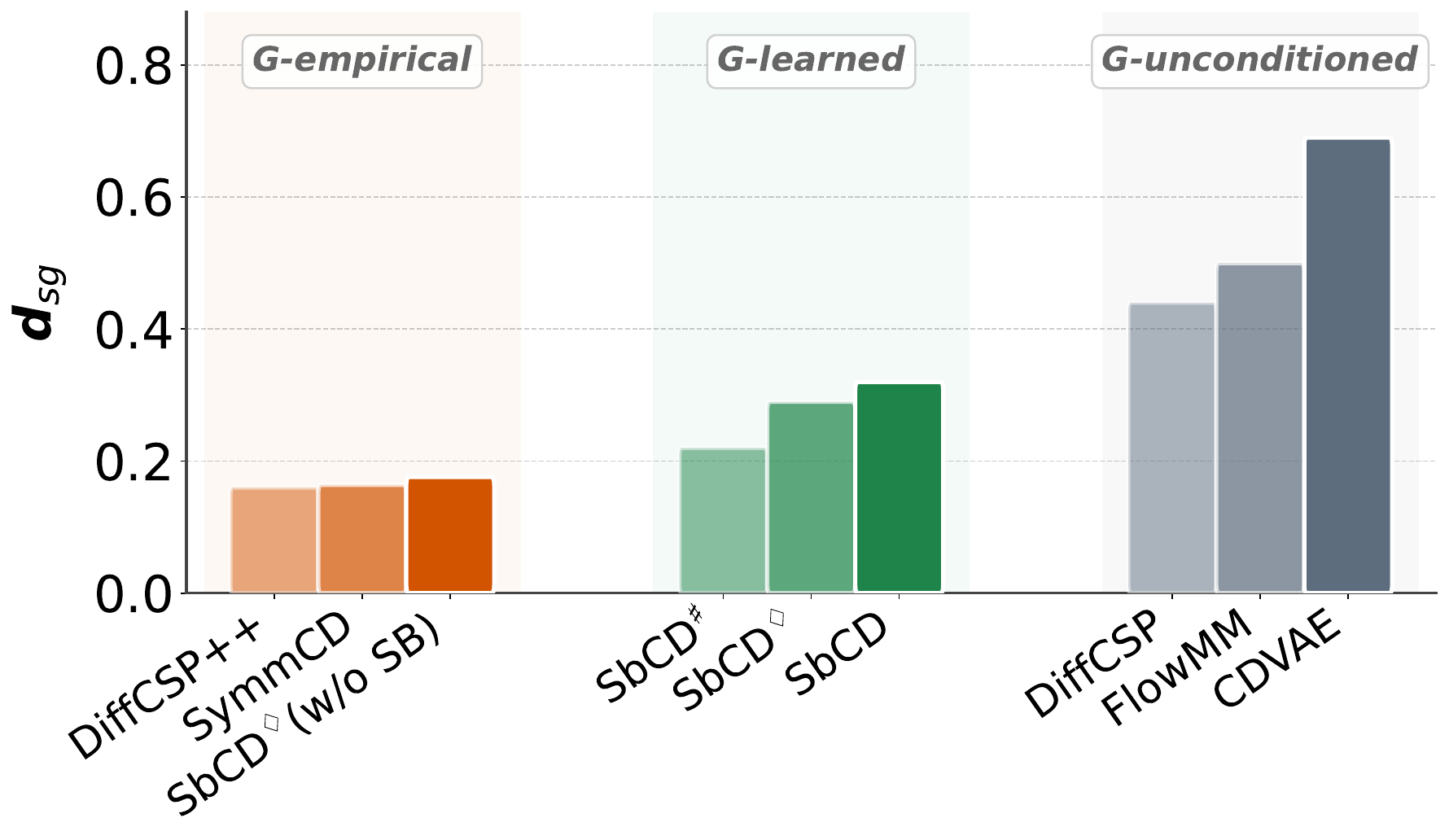}
        \captionof{figure}{Results on the JS space-group distance between generated and test distributions on MP20. $\mathrm{G\text{-}empirical}$ and $\mathrm{G\text{-}learned}$ rely on empirical and learned distributions, while $\mathrm{G\text{-}unconditioned}$ denotes no space-group conditioning.}
        \label{fig:dsg}
    \end{center}
\end{minipage}
\hfill
\begin{minipage}[t]{0.45\textwidth}
    % \vspace{-28pt}
    \begin{center}
        \includegraphics[width=.8\linewidth]{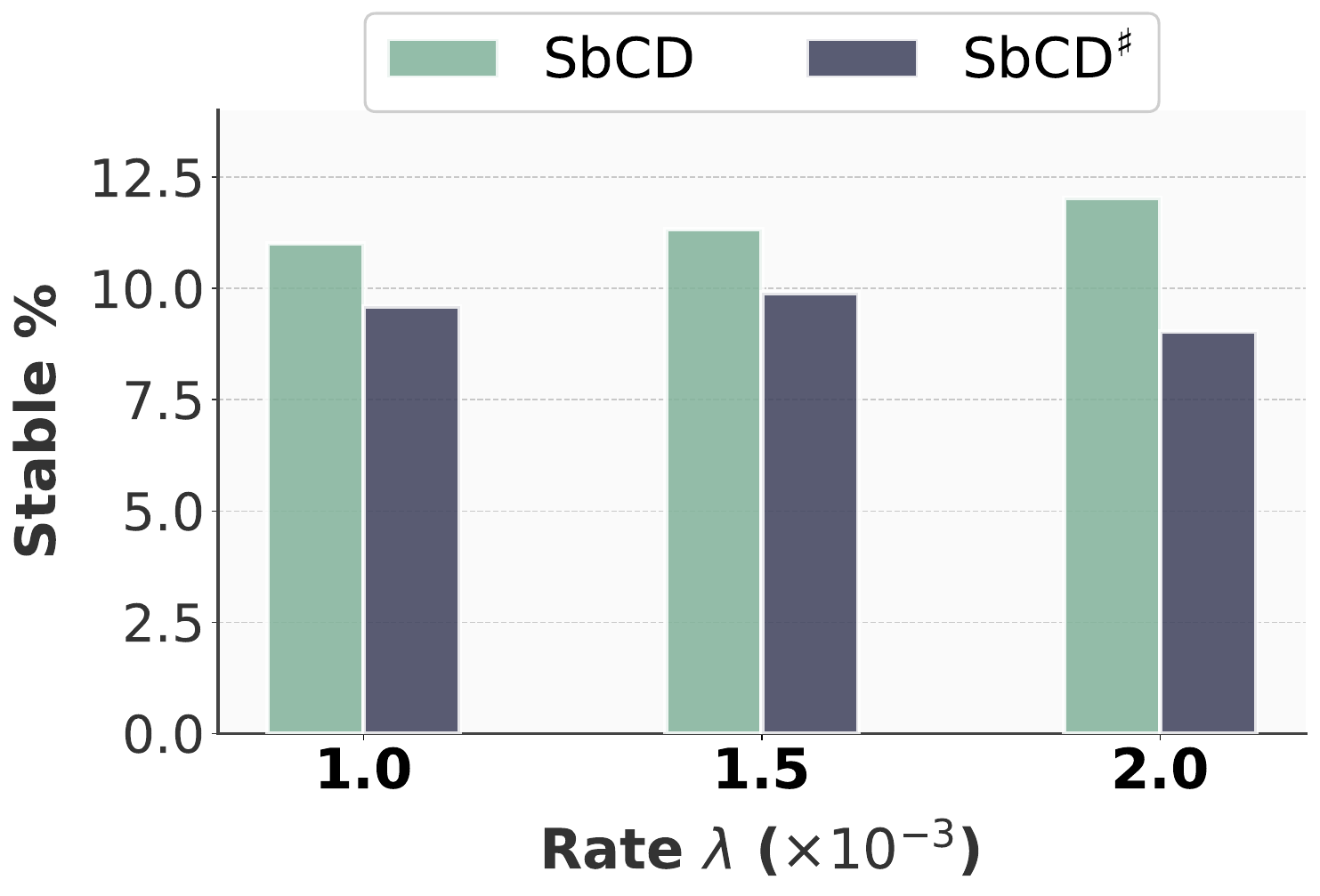}
        \captionof{figure}{Thermodynamic stability results ($E^{\text{hull}} < 0$) across increasing transition rates $\lambda$: Comparison between the novel site-symmetry representation, SbCD, versus the canonical representation on symmetry axes and operations~\citep{levy2025symmcd}, $\text{SbCD}^{\sharp}$.}
    \label{fig:stable}
    \end{center}
\end{minipage}

\subsection{De novo crystal generation results}
%
% best comp. metric result
As shown in Table~\ref{tab:res}, { compared with non-LLM baselines,} SbCD generated crystals, which exhibit a { competitive} proportion of charge-neutral compositions and remain substantially more similar to the test materials in terms of the number of unique elements, indicating a better capture of the underlying compositional distribution.
% thermodynamic stability
On the thermodynamic stability metric, SbCD performs competitively with DiffCSP and DiffCSP++, despite the latter additionally employing space-group constraints on unit-cell fractional coordinates.
% compared to symmcd & ablation no symmetry breaking
Notably, SbCD variants demonstrate clear improvements over their symmetry-preserving SymmCD counterparts across most evaluation criteria, underscoring the effectiveness of incorporating symmetry-breaking transitions into the diffusion process.
% highlight the first generative approach without relying on empirical Wyckoff positions and space groups.
Moreover, existing methods typically depend on external Wyckoff positions for generating crystalline materials. Table~\ref{tab:res} demonstrates that most models suffer performance drops when replacing \textit{empirical} Wyckoff positions with unseen ones generated by \textit{Wyformer} \citep{kazeev2025wyckoff}. 
In contrast, SbCD represents the first framework capable of generating complete crystallographic structure specifications (see Table~\ref{tab:compare}), while outperforming the generated Wyckoff position baselines across various metrics. 

{ The LLM-informed baselines FlowLLM and CrysLLMGen achieve the strongest results for crystal validity and stability on MP-20~\footnote{ In Table~\ref{tab:stat}, the SbCD best checkpoint slightly samples more M.S.U.N crystals than CrysLLMGen.}. Their strong performance stems from how they divide the modeling task. The LLM excels at handling discrete atom types and generates an initial structure, while the diffusion or flow model refines the continuous lattice parameters and atomic coordinates \citep{khastagir2025llm}. Additionally, general-purpose large-scale pretraining of the base LLMs may also capture chemical priors beyond those available in the MP-20 fine-tuning set, making LLM-generated structures a more informative starting point for refinement than pure noise. Using LLM-generated structures as inputs to SbCD could further improve crystal validity and stability by taking advantage of SbCD's ability to model crystalline symmetry.}

\begin{table}[t]
\centering
\caption{De novo crystal generation on MPTS-52. We scale SbCD at different sizes: -S, -L, -XL.}
\label{tab:res_mpts52}
\resizebox{1.\linewidth}{!}{
\begin{tabular}{lcc cc| ccc| cc| cc}
\toprule
\multirow{2}{*}{Method} & \multicolumn{2}{c}{\textbf{Validity} (\%)($\uparrow$)} & \multicolumn{2}{c}{\textbf{Coverage} (\%)($\uparrow$)} & \multicolumn{3}{c}{\textbf{Property statistics} ($\downarrow$)} & \multicolumn{2}{c}{\textbf{CHGNet} $E^{\text{hull}} < 0$ (\%)($\uparrow$)} & \multicolumn{2}{c}{\textbf{CHGNet} $E^{\text{hull}} < 0.1$ (\%)($\uparrow$)}\\
\cmidrule(lr){2-5} \cmidrule(lr){6-8} \cmidrule(lr){9-10} \cmidrule(lr){11-12}
 & $\mathrm{Struct.}$ & $\mathrm{Comp.}$ & $\mathrm{Recall}$ & $\mathrm{Precision}$ & $\mathrm{d_{\rho}}$ &  $\#\; \mathrm{elem.}$ & $\mathrm{d_{sg}}$ & $\mathrm{Stable}$ & $\mathrm{S.U.N}$  &  $\mathrm{MetaStable}$ &   $\mathrm{M.S.U.N}$ \\
\midrule
\midrule
SGEquiDiff (params=5.5M) & \cellcolor{Green!30} $97.93$ & $80.43$ & $99.90$ & $96.15$ & $0.870$ & $0.559$ & $0.306$ & $5.79$ & $3.90$ & $28.10$ & $20.28$\\
SymmCD (params=61M) & $90.10$ & $79.20$ & $99.60$ & $96.30$ & $0.844$ & \cellcolor{Green!30} $0.317$ & \cellcolor{Green!30} $0.274$ & $ 5.97$ & $4.62$ & $-$ & $-$\\
$\text{SymmCD}^{\ddag}$ (params=61M)(retrain) & $85.92$ & $80.22$ & $99.74$ & $95.24$ & $1.104$ & $0.381$ & $0.297$ & $5.26$ & $4.60$ & $23.73$ & $20.66$\\
$   \text{MCFlow (params=464M)}$ &   $97.34$ & \cellcolor{Green!30}  $85.21$ &   $-$ &   $-$ &    $-$ &   $-$ &    $-$ &    $-$ &   $-$ & $-$ & $-$\\
$   \text{SymADiT (params=450M)}$  &   $94.34$ &    $84.23$ &   $-$ &   $-$ &    $-$ &   $0.4$ &    $-$ &    $-$ &   $-$ & $-$ & $-$\\
\midrule
SbCD-S (params=14.8M) & $90.42$ & $78.62$ & $99.57$ & $99.57$ & $0.818$ & $0.585$ &  $0.358$ & $\mathbf{6.88}$ & $\mathbf{5.01}$ & $\mathbf{31.61}$ & $\mathbf{22.44}$\\
SbCD-L (params=47.9M) & $88.02$ &  $81.13$ & $99.53$ & $97.02$ & \cellcolor{Green!30} $0.665$ & $0.567$ &  $0.368$ & $\mathbf{8.23}$ & $\mathbf{5.29}$ & $\mathbf{36.38}$ & $\mathbf{22.82}$\\
SbCD-XL (params=62.1M) & $90.55$ & $80.08$ & $99.65$ & $97.10$ & $0.760$ & $0.654$ &  $0.343$ & \cellcolor{Green!30} $\mathbf{8.36}$ & \cellcolor{Green!30} $\mathbf{5.68}$ & \cellcolor{Green!30} $\mathbf{39.95}$ & \cellcolor{Green!30} $\mathbf{24.79}$\\
\bottomrule
\end{tabular}
} % fixed bugs on the novelty by changing MP-20 to MPTS-52 reference
\end{table}

% space group covering
In Figure~\ref{fig:dsg}, SbCD variants narrow the performance gap with methods directly conditioned on space groups from empirical distributions, and outperform methods without space-group conditioning. 
% symmetry breaking
In parallel, $\text{SbCD}^{\diamondsuit}$ outperforms its variant without learning space group distributions through symmetry-breaking modeling, $\text{SbCD}^{\diamondsuit}$ (w/o SB).
These results highlight SbCD's ability to effectively capture a wide range of space-group distributions.
% challenging dataset and scaling
Eventually, Table~\ref{tab:res_mpts52} shows SbCD's performance on more challenging, larger MPTS-52 crystalline materials. Despite generating lower structural-validity crystals compared to SGEquiDiff, SbCD effectively scales and achieves the best performance on thermodynamic stability metrics. This result emphasizes that high structural validity can not necessarily guarantee thermodynamic-stability improvement on generated materials.

\subsection{Ablation studies}

\begin{wrapfigure}{r}{0.4\linewidth} %this figure will be at the right
    \centering
    \vspace{-40pt}
    \centerline{\includegraphics[width=1.\linewidth]{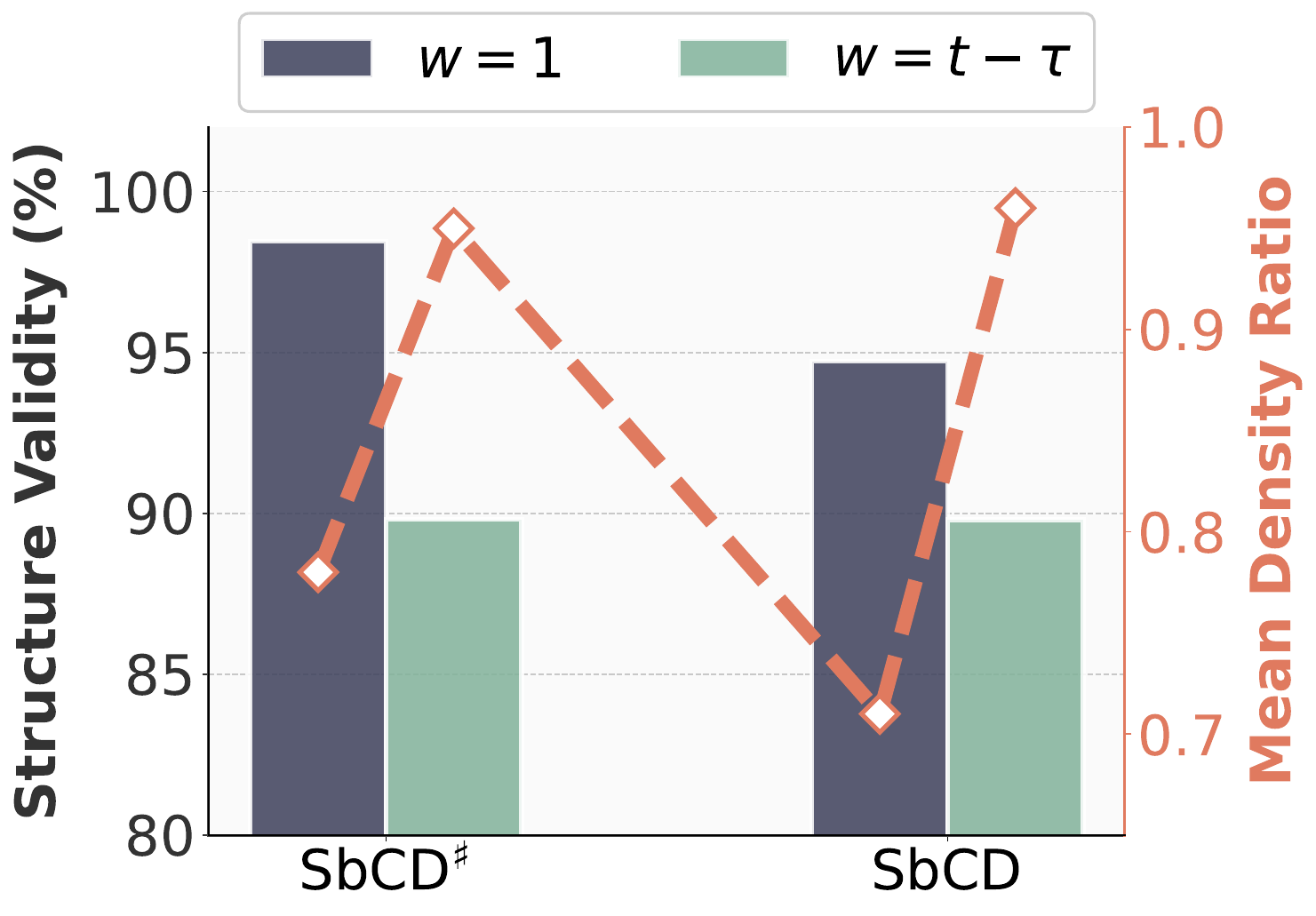}}
    \caption{
      Results on structural validity and mean density ratio w.r.t. the test set for different symmetry-breaking time window sizes, $w$, on MP-20.
    }
    \label{fig:wsize}
    \vspace{-10pt}
\end{wrapfigure}
\paragraph{Symmetry-breaking window revealing the structural validity pitfalls.}
%
% structure validity & explain why is high
Figure~\ref{fig:wsize} visualizes the structural validity results under a fixed time window ($w=1$). We observe that SbCD variants significantly generate more structurally valid crystals ($\sim98\%$). % due to xxx
% mean density ratio
However, their mean density ratio relative to the training data decreases dramatically, indicating that low-density crystals tend to exhibit valid structures more readily. %(minimum pairwise atomic distance above $0.5$\,\r{A}).
% undesirable structures, CDVAE, ref
In Table~\ref{tab:res}, CDVAE, { MatterGen, FlowLLM, CrysLLMGen}, while achieving the strongest performance on structure validity, generate low-density crystals ($\mathrm{d_{\rho}}$ is high), whose structures are loosely packed. This can also be observed in generated structures with low-symmetry space groups, predominantly P1 from \citet{levy2025symmcd}. In practice, we would prioritize obtaining more compact crystal structures, which SbCD variants (with the theoretical window size from Corollary~\ref{cor:tau}) can straightforwardly satisfy, as demonstrated by their mean density ratio results.

\begin{wraptable}{r}{0.5\textwidth}
\begin{center}
\vspace{-17pt}
\captionof{table}{ LeMat-GenBench results.}
\label{tab:lemat}
\resizebox{\linewidth}{!}{
        \begin{tabular}{l cc | cc}
            \toprule
            \multirow{1}{*}{Method}
            & $\mathrm{\mathbf{Stable}}$ ($\%$) & $\mathrm{\mathbf{S.U.N}}$ ($\%$)[count]  & $\mathrm{\mathbf{MetaStable}}$ ($\%$) & $\mathrm{\mathbf{M.S.U.N}} ($\%$)$ \\
            \midrule
            MatterGen & $0.5$ & $0.08$ [$2$] & $36$ & \cellcolor{Green!30}$15.8$\\
            ADiT & $0.1$ & $0.00$  [$0$] & $37.1$ & $1.$\\
            DiffCSP & $1.1$ & $0.04$ [$1$] & $32.2$ & $9.3$\\
            DiffCSP++ &   $0.7$ &  $0.16$ [$4$] & $27.6$ & $5.3$\\
            SymmCD  & $0.5$ & $0.04$ [$1$] & $19.4$ & $2.5$\\
            \midrule
            SbCD & \cellcolor{Green!30} $1.96$ & \cellcolor{Green!30} $0.24$ [$6$] &  \cellcolor{Green!30} $60.93$ & $6.37$\\
            \bottomrule
        \end{tabular}
}
\vspace{-20pt}
\end{center}
\end{wraptable}
\paragraph{ Benchmarking on LetMat GenBench.}
%intro
We evaluate SbCD on MP-20 using LetMat GenBench~\citep{betala2025lemat}, a recent, highly standardized benchmark for crystal generation. To evaluate thermodynamic stability, \citet{betala2025lemat} compute the energy above the convex hull, $E_{\text{hull}}$, using an ensemble of advanced MLIPs comprising MACE-MP~\citep{batatia2025foundationmodelatomisticmaterials}, UMA~\citep{wood2025uma}, and Orb-v3~\citep{rhodes2025orb}. They also use LetMat-Bulk~\citep{siron2025lemat}, which contains approximately $5$ million structures and provides the broadest coverage of competing phases relaxed under consistent DFT settings. The same database serves as a reference for accurately assessing the novelty of generated structures. As recommended by \citet{betala2025lemat}, we generated $2500$ structures and pre-relaxed them to assess their thermodynamic viability better. We report the mean across three runs alongside leaderboard results~\citep{betala2025lemat} for the baselines listed in Table~\ref{tab:res}.

In Table~\ref{tab:lemat}, we observe that SbCD outperforms all the baselines on multiple metrics. SbCD generates significantly higher numbers of stable and metastable structures and also yields more S.U.N. structures. Note that novelty is assessed against $5$ million structures from LetMat-Bulk, so sampling a S.U.N. structure is particularly meaningful. Given the promising results on LetMat GenBench and the availability of our open-source code and model weights, we hope that SbCD, with its principled approach to modeling symmetry breaking, can make meaningful contributions toward addressing real-world materials discovery problems.

\begin{wrapfigure}{r}{0.45\linewidth} %this figure will be at the right
    \centering
    \vspace{-15pt}
    \centerline{\includegraphics[width=1.\linewidth]{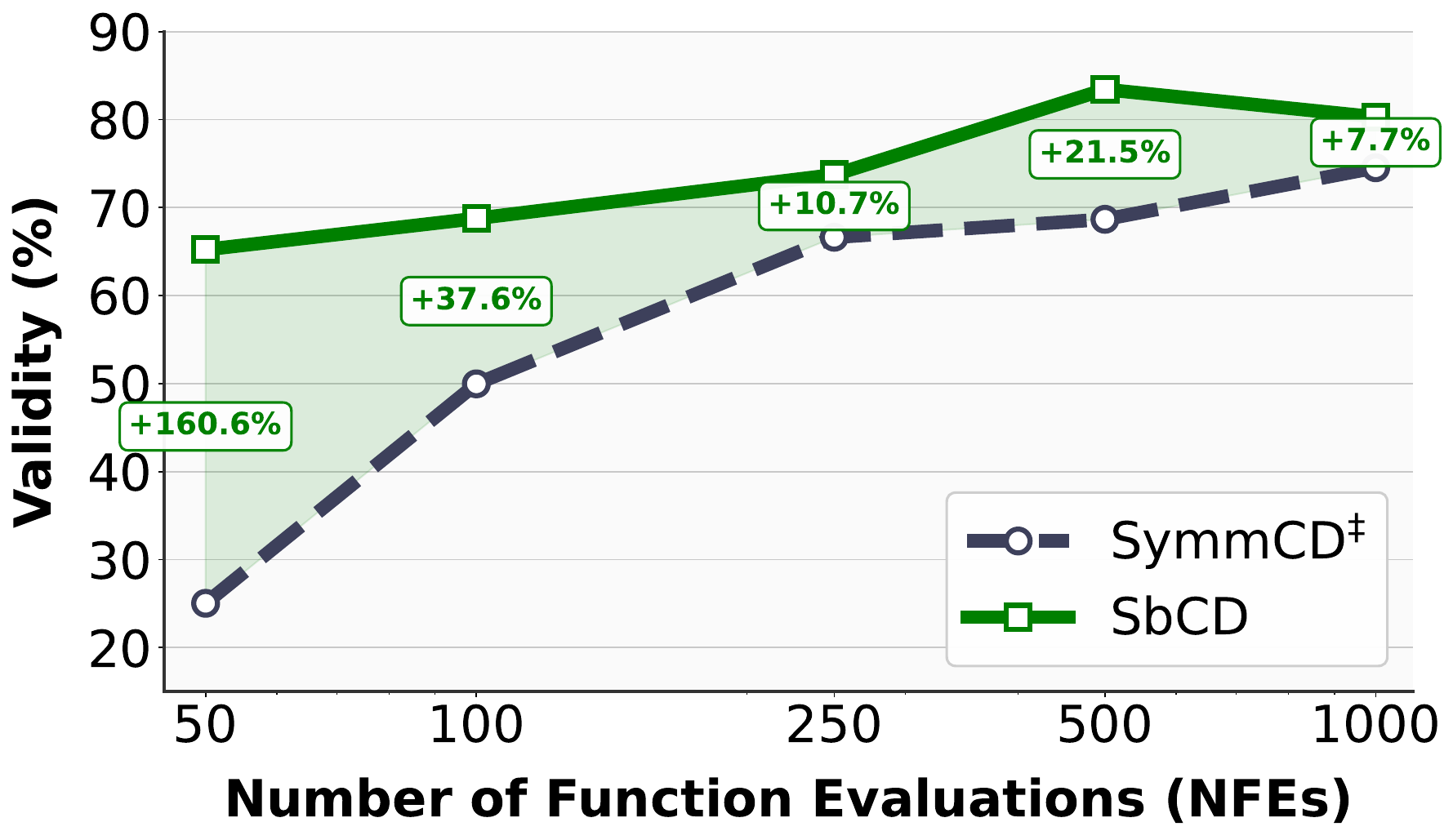}}
    \caption{
      Comparison of crystal validity, defined as the product of structural and compositional validity, between SbCD and the symmetry-preserving model $\text{SymmCD}^{\ddag}$ \citep{levy2025symmcd} across different NFE budgets on MP-20. The NFE budget on the $x$-axis is shown on a logarithmic scale.
    }
    \label{fig:speed}
    % \vspace{-20pt}
\end{wrapfigure}
\paragraph{Robustness of site-symmetry learning under varying transition rates $\lambda$ and NFEs.}
% site symmetry representation 
Rates determine how quickly crystals transition toward the least-constrained space group $\mathrm{P1}$. Therefore, picking a suitable rate is essential so that the diffusion process has enough time to denoise in original space groups while still converging to the prior distributions. Figure~\ref{fig:stable} shows that SbCD achieves greater robustness than $\text{SbCD}^{\sharp}$, even at high transition rates, which underscores the advantage of our simplified site-symmetry representation in alleviating the challenges of modeling the complex site symmetries from \citet{levy2025symmcd}. In terms of inference speed, Figure~\ref{fig:speed} shows that SbCD significantly outperforms its symmetry-preserving counterpart $\text{SymmCD}^{\ddag}$, particularly in low-NFE regimes, which highlights SbCD's potential for enabling few-step de novo crystal generation.

\section{Conclusions}
\label{sec:conclu}
We propose SbCD, a novel symmetry-breaking crystal diffusion model that enables sampling full crystallographic structure specifications from minimal symmetry assumptions. As such, SbCD marks an important step for de novo crystal generation. SbCD takes inspiration from spontaneous symmetry breaking and introduces novel adaptive space group constraints on both continuous and discrete state spaces. Empirically, SbCD substantially outperforms its symmetry-preserving counterpart, offering a promising proof of concept. Future work could extend SbCD by (i) enforcing dynamically varying Wyckoff symmetry constraints on asymmetric-unit fractional coordinates and (ii) exploring more physically realistic transition pathways within its symmetry-breaking framework.

\subsection*{AI use statement}

In this work, we used generative AI tools for [assist in the writing of proofs].
We have not used generative AI tools for [help develop theoretical models or conceptual frameworks, formulate mathematical claims, provide critical ingredients for proving mathematical claims, propose or refine hypotheses,  design or provide feedback on research methodology or experiments, implement methods, interpret results],
and [Generate synthetic data sets, assist with translation, clean and reformat dataset, support qualitative and thematic data analysis] are not applicable to this work.
Additionally, we used generative AI tools for [edit a research paper to improve readability]. We have reviewed the only AI proof-writing assistance on stochastic path measure theories in Proposition 4.1. We take responsibility for the final content of this work,
including text, claims or artifacts produced with the aid of generative AI.

\subsection*{Ethics statement}

SbCD holds the potential to accelerate materials discovery by efficiently exploring large chemical and crystallographic design spaces, potentially enabling advances in energy storage, catalysis, semiconductors, and medicine. At the same time, it may also facilitate the discovery of harmful materials, highlighting the need for responsible use, safety screening, and rigorous experimental validation.

\subsection*{Reproducibility statement}

In the Appendix, we provide detailed descriptions of SbCD’s training and sampling algorithms, complete proofs of the theoretical results, and citations for all datasets used in the experiments. We also include a comprehensive specification of the model architecture, hyperparameters, and computational resources. { We publicly released both the source code and the model weights to support further development by the community.}

\bibliography{iclr2027_conference}
\bibliographystyle{iclr2027_conference}

\appendix
\newpage
\section{Visualization of discovered S.U.N materials}
\begin{figure}[h]
\begin{center}
\includegraphics[width=1.\linewidth]{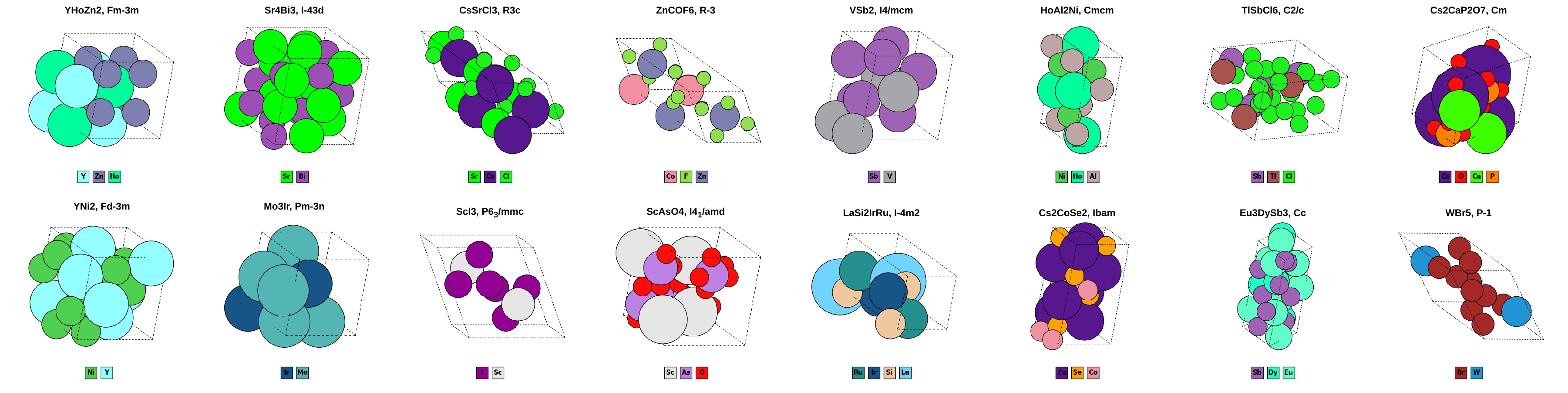}
\end{center}
\caption{S.U.N. MP-20 materials discovered by SbCD.} 
\label{fig:samples_sbcd}
\end{figure}

\begin{figure}[h]
\begin{center}
\includegraphics[width=.68\linewidth]{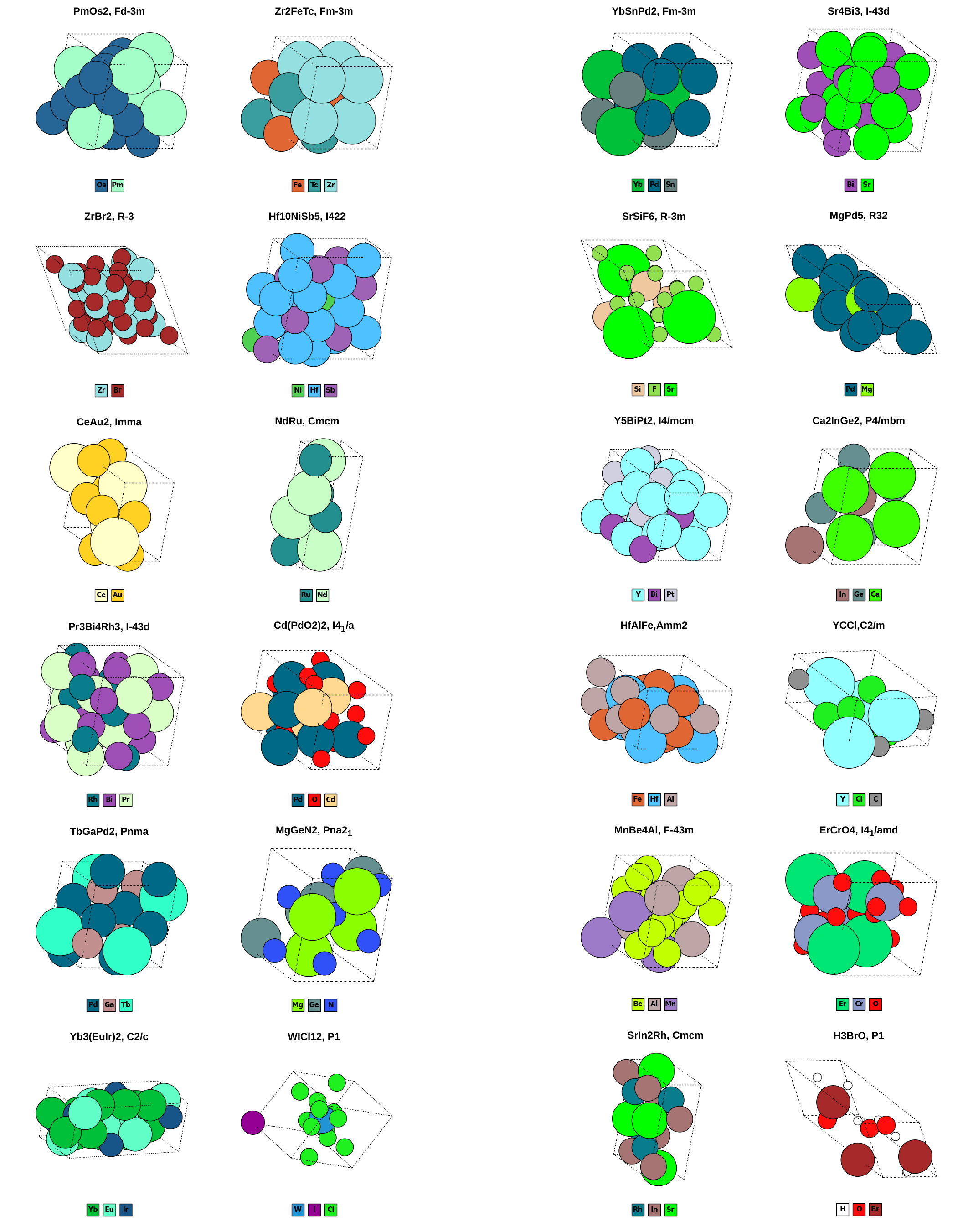}
\end{center}
\caption{Visualization of S.U.N. crystalline materials on MP-20, including their chemical formulas and space group symmetries, discovered with SbCD variants: $\text{SbCD}^{\diamondsuit}$ (left) and $\text{SbCD}^{\sharp}$ (right).}
\label{fig:samples}
\end{figure}
\begin{figure}[h]
\begin{center}
\includegraphics[width=.68\linewidth]{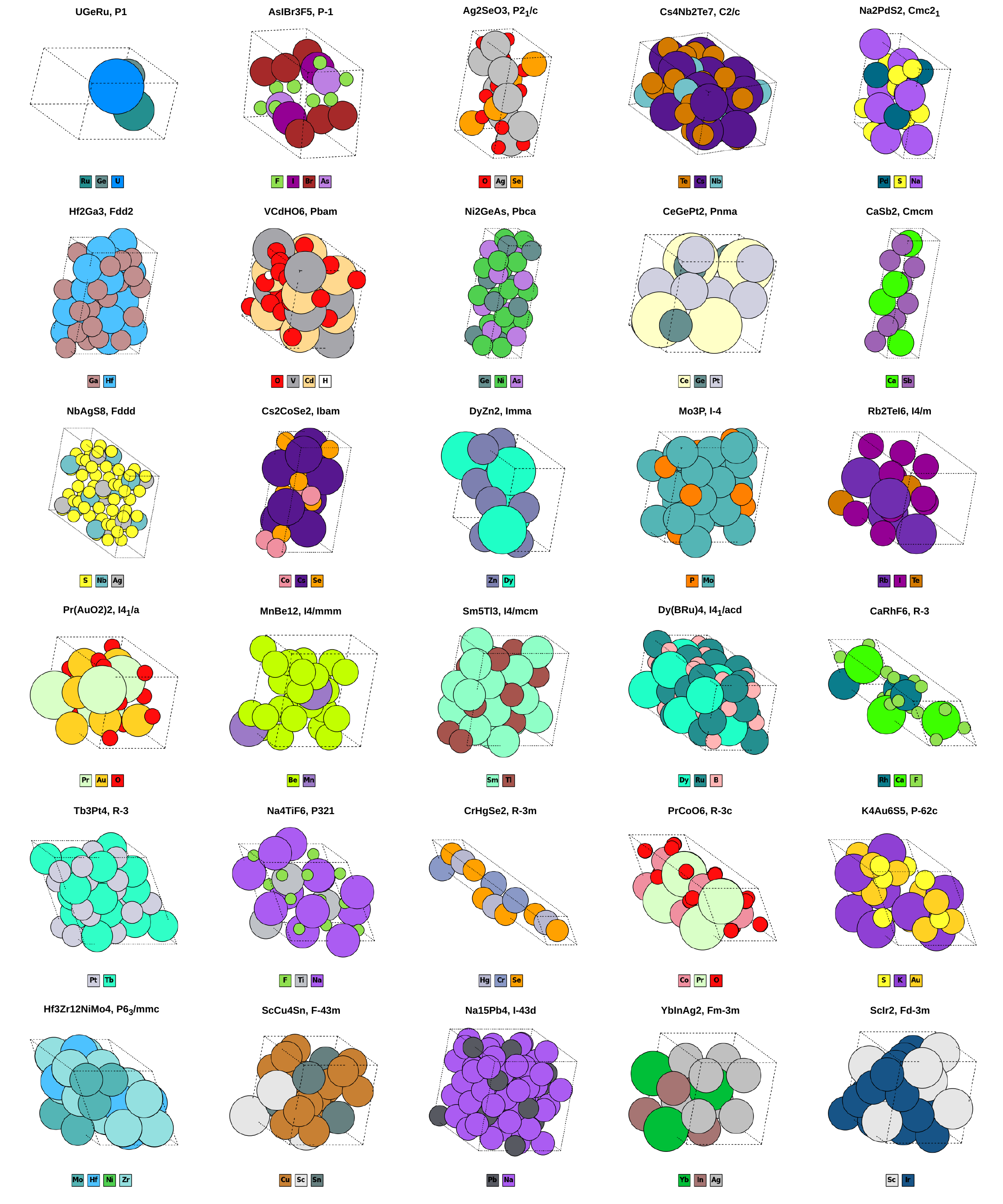}
\end{center}
\caption{S.U.N. MPTS-52 materials discovered by SbCD-XL.}
\label{fig:samples_mpts52}
\end{figure}

\newpage

\section{Future work}
\paragraph{Dynamic Wyckoff symmetry constraints on asymmetric-unit fractional coordinates.}
To our knowledge, existing approaches that enforce such constraints typically require \emph{fixed Wyckoff positions and space groups and operate on unit cells}. For example, DiffCSP++ \citep{jiao2024space} uses fixed empirical Wyckoff positions and projects noisy fractional coordinates onto the corresponding Wyckoff subspaces, ensuring that the generated coordinates satisfy the prescribed symmetry constraints. SGFM \citep{puny2025space} introduces space-group-conditional flow matching, applying frame averaging to unit-cell fractional coordinates using the predefined symmetry operations of a given space group. More recently, SGEquiDiff \citep{chang2025space} adopts a multi-stage training strategy in which earlier stages sample Wyckoff positions, which are then held fixed to constrain fractional-coordinate generation in subsequent stages. In contrast, SbCD jointly evolves space groups, Wyckoff positions, and fractional coordinates on asymmetric units. This setting poses an additional challenge because the associated Wyckoff symmetry constraints can change at every diffusion step. Enforcing these dynamically varying constraints on asymmetric-unit fractional coordinates remains, to our knowledge, an open problem that we leave for future work.

\paragraph{Physically realistic symmetry-breaking transition pathways.} SbCD currently models direct transitions from higher-symmetry space groups to the lowest-symmetry space group, $\mathrm{P1}$. Future work could explore more physically realistic pathways by redesigning the instantaneous rate matrix $\mathbf{\Lambda}_t$ (Proposition~\ref{prop:jump}) alongside adaptive lattice and site-symmetry constraints. This would allow structures to evolve through intermediate space groups toward the $\mathrm{P1}$ prior, potentially capturing symmetry-breaking dynamics more faithfully. Even in its current form, our physics-motivated framework enables generative models to learn complete crystal symmetry data distributions, offers flexibility and improves performance in generating crystals under minimal symmetry assumptions.

\section{Theoretical proofs}
\subsection{Proposition~\ref{prop:path_kl_decomp}}
\begin{proposition*}
Let $\mathbb{Q}$, $\hat{\mathbb{Q}}$, and $\mathbb{P}^{\theta}$ denote the forward, exact reverse, and learned reverse path measures of the continuous-time Markov chain. The negative log-likelihood on crystal structures $\mathcal{M}_0$ is bounded above by the expected Kullback--Leibler divergence over the reverse trajectory:
\begin{equation}
    -\log p_{\theta}(\mathcal{M}_0) \leq \mathbb{E}_{q_{T|0}}\Bigl[ D_{\mathrm{KL}}\bigl( \hat{\mathbb{Q}}^{\mathcal{M}_T \rightarrow \mathcal{M}_0} \,\big\|\, \mathbb{P}^{\theta, \mathcal{M}_T} \bigr) \Bigr] + \mathrm{Const.} \nonumber
\end{equation}
The divergence decomposes additively into space group, lattice, site symmetry, atom type, and fractional coordinate components:
\begin{align}
    D_{\mathrm{KL}}\bigl( \hat{\mathbb{Q}}^{\mathcal{M}_T \rightarrow \mathcal{M}_0} \,&\big\|\, \mathbb{P}^{\theta, \mathcal{M}_T} \bigr) = \underbrace{\colorbox{white!10}{$\displaystyle D_{\mathrm{KL}}\bigl( \hat{\mathbb{Q}}_{\mathbf{A}} \,\big\|\, \mathbb{P}^{\theta}_{\mathbf{A}} \bigr) 
    + D_{\mathrm{KL}}\bigl( \hat{\mathbb{Q}}_{\mathbf{F}} \,\big\|\, \mathbb{P}^{\theta}_{\mathbf{F}} \bigr)$}}_{\textsc{I}:~\text{Atom Types \& Fractional Coordinates}} \nonumber \\
    &+ \underbrace{\colorbox{white!20}{$\displaystyle D_{\mathrm{KL}}\bigl( \hat{\mathbb{Q}}_{G} \,\big\|\, \mathbb{P}^{\theta}_G \bigr)$}}_{\textsc{II}:~\text{Space Group}}
    + \underbrace{\colorbox{white!20}{$\displaystyle \mathbb{E}_{\hat{\mathbb{Q}}_G}\Bigl[ 
        D_{\mathrm{KL}}\bigl( \hat{\mathbb{Q}}_{\mathbf{k}\mid G} \,\big\|\, \mathbb{P}^{\theta}_{\mathbf{k}\mid G} \bigr) 
        + D_{\mathrm{KL}}\bigl( \hat{\mathbb{Q}}_{\mathbf{S}\mid G} \,\big\|\, \mathbb{P}^{\theta}_{\mathbf{S}\mid G} \bigr)
    \Bigr]$}}_{\textsc{III}:~\text{Lattice \& Site Symmetry}} \nonumber
\end{align}
\end{proposition*}

\textit{Proof:} Let $\mathbb{Q}$ denote the forward CTMC path measure, $\mathbb{\hat{Q}}$ the exact reverse path measure, and $\mathbb{P}^{\theta}$ the learned reverse path measure. Let $\mathcal{P}=C([0,T], \mathbb{M})$ denote the path space on $\mathbb{M}$. For an assymmetric-unit crystal representation  $\mathcal{M}=\{G, \mathbf{A}, \mathbf{k}, \mathbf{S}, \mathbf{F}\}$, we denote by $\mathcal{M}_t$ the noisy structure at diffusion timestep $t$, and by $W_s$ the reverse diffusion path. Following \citet{campbell2022a}, we derive the negative log-likelihood of the model. 
\scalebox{.8}{
\begin{minipage}{\linewidth}
\begin{align}
    -\log p_{\theta}(\mathcal{M}_0) &= -\log\int_{\mathbb{M}}p_{\text{ref}}(d\mathcal{M}_T)\int_{\mathcal{P}}\mathbf{1}_{\{W_T=\mathcal{M}_0\}}\mathbb{P}^{\theta, \mathcal{M}_T}(dw) \nonumber\\ 
    &= -\log\int_{\mathbb{M}}q_{T|0}(d\mathcal{M}_T|\mathcal{M}_0)\int_{\mathcal{P}}\mathbf{1}_{\{W_T=\mathcal{M}_0\}}\mathbb{\hat{Q}}^{\mathcal{M}_T}(dw) \frac{dp_{\text{ref}}}{dq_{T|0}(\cdot|\mathcal{M}_0)}(\mathcal{M}_T)\frac{d\mathbb{P}^{\theta, \mathcal{M}_T}}{d\mathbb{\hat{Q}}^{\mathcal{M}_T}}(w)\nonumber\\
    &= -\log\int_{\mathbb{M}}q_{T|0}(d\mathcal{M}_T|\mathcal{M}_0)\int_{\mathcal{P}}\mathbb{\hat{Q}}^{\mathcal{M}_T}(dw|W_T=\mathcal{M}_0) \frac{dp_{\text{ref}}}{dq_{T|0}(\cdot|\mathcal{M}_0)}(\mathcal{M}_T)\frac{d\mathbb{P}^{\theta, \mathcal{M}_T}}{d\mathbb{\hat{Q}}^{\mathcal{M}_T}}(w)\mathbb{\hat{Q}}^{\mathcal{M}_T}(\{W_T=\mathcal{M}_0\})
\end{align}
\end{minipage}
}

Where $\mathbb{\hat{Q}}^{\mathcal{M}_T}, \mathbb{P}^{\theta, \mathcal{M}_T}$ denote the exact and learned reverse path measures starting from $\mathcal{M}_T$, respectively. In the second equation, we apply a change of measure, and in the third equation, we apply a bridge decomposition of $\mathbf{1}_{\{W_T=\mathcal{M}_0\}}\mathbb{\hat{Q}}^{\mathcal{M}_T}(dw)$. Specifically, this term is decomposed into the bridge path measure, $\mathbb{\hat{Q}}^{\mathcal{M}_T}(dw|W_T=\mathcal{M}_0)$ and the probability $\mathbb{\hat{Q}}^{\mathcal{M}_T}(\{W_T=\mathcal{M}_0\})$, which represents a probability that a reverse path starting from $\mathcal{M}_T$ terminates at $\mathcal{M}_0$. We then define the joint probability measure $\mu(d\mathcal{M}_T,dw)$ and the integrand $f(\mathcal{M}_T,w)$ as follows:
\begin{align}
    \mu(d\mathcal{M}_T,dw)=q_{T|0}(d\mathcal{M}_t|\mathcal{M}_0)\mathbb{\hat{Q}}^{\mathcal{M}_T}(dw|W_T=\mathcal{M}_0) \nonumber\\
    f(\mathcal{M}_T,w)=\frac{dp_{\text{ref}}}{dq_{T|0}(\cdot|\mathcal{M}_0)}(\mathcal{M}_T)\frac{d\mathbb{P}^{\theta,\mathcal{M}_T}}{d\mathbb{\hat{Q}}^{\mathcal{M}_T}}(w)\mathbb{\hat{Q}}^{\mathcal{M}_T}(\{W_T=\mathcal{M}_0\})
\end{align}
We can rewrite the negative log-likelihood:
\begin{align}
    -\log p_{\theta}(\mathcal{M}_0) = - \log \int f\left(\mathcal{M}_T, w\right)\mu(d\mathcal{M}_T, dw)
\end{align}
By Jensen's inequality ($-\log$ is convex, so $-\log \mathbb{E}[X] \leq \mathbb{E}[-\log X]$)
\begin{align}
    -\log p_{\theta}(\mathcal{M}_0) \leq \int - \log f(\mathcal{M}_T,w)  \mu(d\mathcal{M}_T, dw)
\end{align}
$$
\text{where} \quad -\log f(\mathcal{M}_T,w) = -\log \frac{dp_{\text{ref}}}{dq_{T|0}(\cdot|\mathcal{M}_0)}(\mathcal{M}_T) - \log \frac{d\mathbb{P}^{\theta,\mathcal{M}_T}}{d\mathbb{\hat{Q}}^{\mathcal{M}_T}}(w) - \log\mathbb{\hat{Q}}^{\mathcal{M}_T}(\{W_T=\mathcal{M}_0\}) 
$$
The last term is independent of the model parameter $\theta$, and the first term does not depend on the path realization $w$. Therefore, taking the expectation with respect to $\mu$ yields
\begin{align}
    -\log p_{\theta}(\mathcal{M}_0)
    &\leq
    \mathbb{E}_{\mu}\left[
    -\log
    \frac{d\mathbb{P}^{\theta,\mathcal{M}_T}}
    {d\widehat{\mathbb{Q}}^{\mathcal{M}_T}}(w)
    \right]
    + C_0 \nonumber\\
    &=
    \mathbb{E}_{q_{T|0}(\cdot|\mathcal{M}_0)}
    \left[
    D_{\mathrm{KL}}\Bigl(
    \widehat{\mathbb{Q}}^{\mathcal{M}_T\rightarrow \mathcal{M}_0}
    \,\|\, \mathbb{P}^{\theta,\mathcal{M}_T}
    \Bigr)
    \right]
    + C,
\end{align}
where
$$
C_0 = \mathbb{E}_{q_{T|0}(\cdot|\mathcal{M}_0)} \left[ -\log \widehat{\mathbb{Q}}^{\mathcal{M}_T}(\{W_T=\mathcal{M}_0\}) -\log \frac{dp_{\text{ref}}}{dq_{T|0}(\cdot|\mathcal{M}_0)}(\mathcal{M}_T) \right],
$$

After rewriting the bridge expectation as a KL divergence, the bridge-normalization term cancels with the corresponding term in $C_0$.
$$ 
C = \mathbb{E}_{q_{T|0}(\cdot|\mathcal{M}_0)} \left[ -\log\frac{dp_{\text{ref}}}{dq_{T|0}(\cdot|\mathcal{M}_0)}(\mathcal{M}_T) \right].
$$
Finally,
$
\widehat{\mathbb{Q}}^{\mathcal{M}_T\rightarrow \mathcal{M}_0} := \widehat{\mathbb{Q}}^{\mathcal{M}_T}(\cdot|W_T=\mathcal{M}_0)
$
denotes the bridge measure connecting $\mathcal{M}_T$ and $\mathcal{M}_0$.

We assume that the bridge measure factorizes according to the dependency structure of the crystal representation, with $G$ governing $\mathbf{k}$ and $\mathbf{S}$, while $\mathbf{A}$ and $\mathbf{F}$ are modeled independently. Under the same factorization for the learned reverse measure, the KL chain rule yields the following decomposition.

\scalebox{.8}{
\begin{minipage}{\linewidth}
\begin{align}
    \mathbb{\hat{Q}}^{\mathcal{M}_T\rightarrow\mathcal{M}_0}(dG, d\mathbf{A}, d\mathbf{k}, d\mathbf{S}, d\mathbf{F}) =\mathbb{\hat{Q}}_G^{\mathcal{M}_T\rightarrow\mathcal{M}_0}(dG)\mathbb{\hat{Q}}^{\mathcal{M}_T\rightarrow\mathcal{M}_0}_{\mathbf{k}|G}(d\mathbf{k}|G)\cdot\mathbb{\hat{Q}}_{\mathbf{S}|G}^{\mathcal{M}_T\rightarrow\mathcal{M}_0}(d\mathbf{S}|G)\mathbb{\hat{Q}}^{\mathcal{M}_T\rightarrow\mathcal{M}_0}_{\mathbf{A}}(d\mathbf{A})\mathbb{\hat{Q}}^{\mathcal{M}_T\rightarrow\mathcal{M}_0}_{\mathbf{F}}(d\mathbf{F}) \nonumber \\
    \mathbb{P}^{\theta,\mathcal{M}_T}(dG, d\mathbf{A}, d\mathbf{k}, d\mathbf{S}, d\mathbf{F}) =\mathbb{P}_{G}^{\theta,\mathcal{M}_T}(dG)\cdot\mathbb{P}_{\mathbf{k}|G}^{\theta,\mathcal{M}_T}(d\mathbf{k}|G)\mathbb{P}^{\theta,\mathcal{M}_T}_{\mathbf{S}|G}(d\mathbf{S}|G)\mathbb{P}^{\theta,\mathcal{M}_T}_{\mathbf{A}}(d\mathbf{A})\mathbb{P}_{\mathbf{F}}^{\theta,\mathcal{M}_T}(d\mathbf{F})
\end{align}
\end{minipage}
}

We apply the KL chain rule that yields: % Principle of Minimum Cross-Entropy

\scalebox{.9}{
\begin{minipage}{\linewidth}
\begin{align}
    D_{\mathrm{KL}}\left(\mathbb{\hat{Q}}^{\mathcal{M}_T\rightarrow\mathcal{M}_0}\|\mathbb{P}^{\theta, \mathcal{M}_T}\right) &= D_{\mathrm{KL}}\left(\mathbb{\hat{Q}}_{G}^{\mathcal{M}_T\rightarrow\mathcal{M}_0}\|\mathbb{P}^{\theta, \mathcal{M}_T}_G\right) \nonumber\\
    &+ \mathbb{E}_{\mathbb{\hat{Q}}_G^{\mathcal{M}_T\rightarrow\mathcal{M}_0}}\Bigl[D_{\mathrm{KL}}\left(\mathbb{\hat{Q}}^{\mathcal{M}_T\rightarrow\mathcal{M}_0}_{\mathbf{k}|G}\|\mathbb{P}^{\theta, \mathcal{M}_T}_{\mathbf{k}|G}\right) 
    + D_{\mathrm{KL}}\left(\mathbb{\hat{Q}}^{\mathcal{M}_T\rightarrow\mathcal{M}_0}_{\mathbf{S}|G}\|\mathbb{P}^{\theta, \mathcal{M}_T}_{\mathbf{S}|G}\right)\Bigr] \nonumber\\
    &+  D_{\mathrm{KL}}\left(\mathbb{\hat{Q}}^{\mathcal{M}_T\rightarrow\mathcal{M}_0}_{\mathbf{A}}\|\mathbb{P}^{\theta, \mathcal{M}_T}_{\mathbf{A}}\right) + D_{\mathrm{KL}}\left(\mathbb{\hat{Q}}^{\mathcal{M}_T\rightarrow\mathcal{M}_0}_{\mathbf{F}}\|\mathbb{P}^{\theta,\mathcal{M}_T}_{\mathbf{F}}\right)
\end{align}
\end{minipage}
}

% We conclude the proof. 
Now, we consider a discretized version of the continuous-time exact reverse bridge path measure on $\mathbf{A}, \mathbf{F}$:
\begin{align}
    \mathbb{\hat{Q}}_{\mathbf{A}}^{\mathcal{M}_T \rightarrow \mathcal{M}_0} (dw) \Rightarrow q(\mathbf{A}_{t_1}, \mathbf{A}_{t_2}, \dots, \mathbf{A}_{t_{N-1}} | \mathbf{A}_0, \mathbf{A}_T) \nonumber\\
    \mathbb{\hat{Q}}_{\mathbf{F}}^{\mathcal{M}_T \rightarrow \mathcal{M}_0} (dw) \Rightarrow q(\mathbf{F}_{t_1}, \mathbf{F}_{t_2}, \dots, \mathbf{F}_{t_{N-1}} | \mathbf{F}_0, \mathbf{F}_T)
\end{align}

And, the learned reverse process continuous-time path measure can be discretized as:
\begin{align}
    \mathbb{{P}}^{\theta, \mathcal{M}_T}_{\mathbf{A}}(dw) \Rightarrow p_{\theta}( \mathbf{A}_{t_{N-1}},\dots,\mathbf{A}_{t_{0}}|\mathcal{M}_T)=\prod_{j=1}^N p_{\theta}(\mathbf{A}_{t_{j-1}}|\mathcal{M}_{t_j}) \mathrm{,\;where\;\mathcal{M}_{t_j}\; contains\; \mathbf{A}_{t_{j}}} \nonumber\\
    \mathbb{{P}}^{\theta, \mathcal{M}_T}_{\mathbf{F}}(dw) \Rightarrow p_{\theta}( \mathbf{F}_{t_{N-1}},\dots,\mathbf{F}_{t_{0}}|\mathcal{M}_T)=\prod_{j=1}^N p_{\theta}(\mathbf{F}_{t_{j-1}}|\mathcal{M}_{t_j}) \mathrm{,\;where\;\mathcal{M}_{t_j}\; contains\; \mathbf{F}_{t_{j}}}
\end{align}
We obtain the discretized Radon–Nikodym derivative
\begin{align}
    \log \frac{d\mathbb{\hat{Q}}_{\mathbf{A}}^{\mathcal{M}_T \rightarrow \mathcal{M}_0}}{d\mathbb{P}_{\mathbf{A}}^{\theta,\mathcal{M}_T}}(w) \Rightarrow \sum_{j=1}^{N} \log \frac{q(\mathbf{A}_{t_{j-1}}|\mathbf{A}_0, \mathbf{A}_{t_j})}{p_{\theta}(\mathbf{A}_{t_{j-1}}|\mathcal{M}_{t_j})} \nonumber\\
    \log \frac{d\mathbb{\hat{Q}}_{\mathbf{F}}^{\mathcal{M}_T \rightarrow \mathcal{M}_0}}{d\mathbb{P}_{\mathbf{F}}^{\theta,\mathcal{M}_T}}(w) \Rightarrow \sum_{j=1}^{N} \log \frac{q(\mathbf{F}_{t_{j-1}}|\mathbf{F}_0, \mathbf{F}_{t_j})}{p_{\theta}(\mathbf{F}_{t_{j-1}}|\mathcal{M}_{t_j})}
\end{align}

These discretized Radon–Nikodym derivatives recover the KL terms appearing in the discrete-time evidence upper bound of \citet{sohl2015deep}, which underlies diffusion models in continuous-state spaces \citep{ho2020denoising} and discrete-state spaces \citep{austin2021structured}. Similarly, we obtain the discretized Radon–Nikodym derivatives for $\mathbf{k}$ and $\mathbf{S}$:
\begin{align}
    &\log \frac{d\mathbb{\hat{Q}}_{\mathbf{k}|G}^{\mathcal{M}_T \rightarrow \mathcal{M}_0}}{d\mathbb{P}_{\mathbf{k}|G}^{\theta,\mathcal{M}_T}}(w) \Rightarrow \sum_{j=1}^{N} \log \frac{q(\mathbf{k}_{t_{j-1}}|\mathbf{k}_0, \mathbf{k}_{t_j}, G_{t_j},G_{t_{j-1}})}{p_{\theta}(\mathbf{k}_{t_{j-1}}|\mathcal{M}_{t_j}, G_{t_{j-1}})} \mathrm{,\;where\;\mathcal{M}_{t_j}\; contains\; G_{t_{j}}\; and\; \mathbf{k}_{t_{j}}} \nonumber\\
    &\log \frac{d\mathbb{\hat{Q}}_{\mathbf{S}|G}^{\mathcal{M}_T \rightarrow \mathcal{M}_0}}{d\mathbb{P}_{\mathbf{S}|G}^{\theta,\mathcal{M}_T}}(w) \Rightarrow \sum_{j=1}^{N} \log \frac{q(\mathbf{S}_{t_{j-1}}|\mathbf{S}_0, \mathbf{S}_{t_j}, G_{t_j},G_{t_{j-1}})}{p_{\theta}(\mathbf{S}_{t_{j-1}}|\mathcal{M}_{t_j},G_{t_{j-1}})} \mathrm{,\;where\;\mathcal{M}_{t_j}\; contains\; G_{t_{j}}\; and\; \mathbf{S}_{t_{j}}}  \nonumber
\end{align}

Finally, the remaining term $D_{\mathrm{KL}}\left(\mathbb{\hat{Q}}_{G}^{\mathcal{M}_T\rightarrow\mathcal{M}_0}\|\mathbb{P}^{\theta, \mathcal{M}_T}_G\right)$ is optimized using  CTMC Markov jump diffusion from \citet{Feller1949OnTT,campbell2022a}.

We conclude the proof.

\subsection{Proposition~\ref{prop:jump}}
\begin{proposition*}
Consider the discrete state space of $230$ crystallographic space groups drawn from the six crystal families. Let $\mathbf{\Pi}, \mathbf{\Lambda}_t \in \mathbb{R}^{230 \times 230}$ denote the time-invariant transition probability matrix and the instantaneous rate matrix, respectively, defined element-wise as follows:
\begin{align}
\mathbf{\Pi}(\widetilde{G},G)=
\begin{cases}
0 & \widetilde{G} = \mathrm{P1} \\
0 & G = \widetilde{G} \\
1 & G = \mathrm{P1} \neq  \widetilde{G} \\
0 & \text{otherwise}.
\end{cases}
\qquad
\mathbf{\Lambda}_t(\widetilde{G},G)=
\begin{cases}
0 & \widetilde{G} = \mathrm{P1} \\
-\lambda(t) & G = \widetilde{G} \\
\lambda(t) & G = \mathrm{P1} \neq \widetilde{G} \\
0 & \text{otherwise}. \nonumber
\end{cases}
\end{align}
   
Then, $\mathbf{\Lambda}_t$ and $\mathbf{\Lambda}_s$ commute for all $t,s$. Consequently, the forward transition probability of being in state $G$ at time $t$, given state $\widetilde{G}$ at time $\widetilde{t} < t$, $q_{t|\widetilde{t}}(G|\widetilde{G})$, admits the matrix exponential form:
\begin{align}
q_{t|\widetilde{t}}(G|\widetilde{G}) = \left(\mathbf{D}\exp\Bigr[\mathbf{\Sigma}\int_{\widetilde{t}}^t\lambda(s)ds\Bigl]\mathbf{D}^{-1}\right)_{\widetilde{G},G} \quad \text{with} \quad
\frac{\mathbf{\Lambda}_t}{\lambda(t)} =
\mathbf{D}\mathbf{\Sigma}\mathbf{D}^{-1} \nonumber
\end{align}
\end{proposition*}

\textit{Proof:} We consider a continuous-time Markov process with time-dependent transition rates. Let $q_{t|\widetilde{t}}(G|\widetilde{G})$ denote the transition probability to be in the space group $G$ at time $t$, if the process was in the space group $\widetilde{G}$ at the earlier time $\widetilde{t} <t$. We start from the Kolmogorov forward equation \citep{Feller1949OnTT} in the jump-rate form:

\begin{align}
    \frac{\partial}{\partial t} q_{t|\widetilde{t}}(G|\widetilde{G}) = -\lambda_t(G)q_{t|\widetilde{t}}(G|\widetilde{G}) + \sum_{G'}\lambda_t(G')\mathbf{\Pi}_t(G',G)q_{t|\widetilde{t}}(G'|\widetilde{G})
\end{align} 
Here,
\begin{itemize}
    \item $\lambda_t(G')$ is the total jump rate out of state $G'$ at time $t$.
    \item $\mathbf{\Pi}_t(G', G)$ is the probability that, given a jump occurs from the space group $G'$ at time $t$, the process jumps to the space group $G$ (and $\Pi_t(G', G')=0$).
\end{itemize}

We recall the \textit{instantaneous} transition rate (generator) matrix $\mathbf{\Lambda}_t(G',G)$ with entries:
\begin{align}
    \mathbf{\Lambda}_t(G',G) = \lambda_t(G')\mathbf{\Pi}_{t}(G',G) - \delta_{G'G}\lambda_t(G') =\left\{\begin{array}{cc}
    \lambda_t(G')\mathbf{\Pi}_{t}(G',G) &  G \ne G' \\
    -\lambda_t(G') & G = G'
    \end{array}\right.
\end{align}
Apply the generator matrix to the forward equation, we have:

\begin{align}
    \frac{\partial}{\partial t} q_{t|\widetilde{t}}(G|\widetilde{G}) &= -\lambda_t(G)q_{t|\widetilde{t}}(G|\widetilde{G}) + \sum_{G'}\lambda_t(G')\mathbf{\Pi}_t(G',G)q_{t|\widetilde{t}}(G'|\widetilde{G})  \nonumber\\
    &= \mathbf{\Lambda}_t(G,G)q_{t|\widetilde{t}}(G|\widetilde{G}) + \sum_{G'}\lambda_t(G')\mathbf{\Pi}_t(G',G)q_{t|\widetilde{t}}(G'|\widetilde{G}) \nonumber\\
    &=  \mathbf{\Lambda}_t(G,G)q_{t|\widetilde{t}}(G|\widetilde{G}) + \sum_{G'\ne G} \mathbf{\Lambda}_t(G',G)q_{t|\widetilde{t}}(G'|\widetilde{G}) \nonumber\\
    &= \sum_{G'} \mathbf{\Lambda}_t(G',G)q_{t|\widetilde{t}}(G'|\widetilde{G}) = \left[q_{t|\widetilde{t}}\mathbf{\Lambda}_t\right]_{\widetilde{G},G}
\end{align}
In the matrix form, the forward equation is equivalent to the matrix ODE:
\begin{align}
    \frac{\partial}{\partial t}q_{t|\widetilde{t}}=q_{t|\widetilde{t}}\mathbf{\Lambda}_t
\end{align}
Integrating both sides gives:
\begin{align}
    q_{t|\widetilde{t}}=I + \int_{\widetilde{t}}^tq_{t|\widetilde{t}}\mathbf{\Lambda}_sds
\end{align}
Iterating the integral gives the Dyson series:
\begin{align}
    q_{t|\widetilde{t}} &=I + \int_{\widetilde{t}}^t\mathbf{\Lambda}_{s_1}ds_1 + \int_{\widetilde{t}}^t\int_{\widetilde{t}}^{s_1}\mathbf{\Lambda}_{s_1}\mathbf{\Lambda}_{s_2}ds_1ds_2 + \ldots \nonumber\\
    &= I + \sum_{n=1}^{\infty} \int_{\widetilde{t}}^t\int_{\widetilde{t}}^{s_1}\cdots\int_{\widetilde{t}}^{s_{n-1}}\mathbf{\Lambda}_{s_n}\cdots \mathbf{\Lambda}_{s_1}ds_n\cdots ds_1
\end{align}

with the ordering $\widetilde{t} \leq s_n \leq \cdots \leq s_1 \leq t$. Now, assuming that the instantaneous transition rate matrix $\mathbf{\Lambda}_{s_i}$ commutes with $\mathbf{\Lambda}_{s_j}$ for all $s_i,s_j$, i.e., $\mathbf{\Lambda}_{s_i}\mathbf{\Lambda}_{s_j}=\mathbf{\Lambda}_{s_j}\mathbf{\Lambda}_{s_i}$. With this assumption, the integrand is symmetric in $(s_1, s_2, \cdots, s_n)$. The integration can be further simplified to the matrix exponential form:
\begin{align}
    \label{eq:exp}
    q_{t|\widetilde{t}}&=I +  \sum_{n=1}^{\infty} \frac{1}{n!}\left(\int_{\widetilde{t}}^t\mathbf{\Lambda}_{s}ds\right)^n \nonumber\\
    &=\exp\left(\int_{\widetilde{t}}^t\mathbf{\Lambda}_{s}ds\right)
\end{align}

We now consider the space-group \textit{relative} transition probability $\mathbf{\Pi}_t$ that is constant over time:
\begin{align}
\mathbf{\Pi}_t(\widetilde{G},G)=\mathbf{\Pi}(\widetilde{G},G)=
\begin{cases}
0 & \widetilde{G} = \mathrm{P1} \\
0 & G = \widetilde{G} \\
1 & G = \mathrm{P1} \neq  \widetilde{G} \\
0 & \text{otherwise}.
\end{cases}
\end{align}

This design forces every jump to the lowest symmetry space group $\mathrm{P1}$. The instantaneous rate matrix thus can be obtained by applying its definition with $\mathbf{\Pi}_t$ above:
\begin{align}
\mathbf{\Lambda}_t(\widetilde{G},G)=\lambda(t)\mathbf{\Pi}^{+}(\widetilde{G},G)=
\begin{cases}
0 & \widetilde{G} = \mathrm{P1} \\
-\lambda(t) & G = \widetilde{G} \\
\lambda(t) & G = \mathrm{P1} \neq \widetilde{G} \\
0 & \text{otherwise}.
\end{cases}  \text{with} \quad
\mathbf{\Pi}^{+}(\widetilde{G},G)=
\begin{cases}
0 & \widetilde{G} = \mathrm{P1} \\
-1 & G = \widetilde{G} \\
1 & G = \mathrm{P1} \neq \widetilde{G} \\
0 & \text{otherwise}.
\end{cases}
\end{align}

We can easily verify the commutativity of the instantaneous rate matrix for all $t,s$:
\begin{align}
    \mathbf{\Lambda}_t\mathbf{\Lambda}_s=\lambda(t)\mathbf{\Pi}^{+}\lambda(s)\mathbf{\Pi}^{+}=\lambda(t)\lambda(s)\mathbf{\Pi}^{+}\mathbf{\Pi}^{+}=\mathbf{\Lambda}_s\mathbf{\Lambda}_t
\end{align}

Thus, we can apply Equation \ref{eq:exp} to get the analytical form of space-group transition probabilities.

\begin{align}
    q_{t|\widetilde{t}} &= \exp \left(\int_{\widetilde{t}}^t \Lambda_s ds\right) \nonumber\\
        &= \exp \left(\mathbf{\Pi}^{+}\int_{\widetilde{t}}^t \lambda(s) ds\right) \nonumber\\
        &= \mathbf{D} \exp \left(\Sigma \int_{\widetilde{t}}^t \lambda(s) ds \right) \mathbf{D}^{-1}
\end{align}
with the eigen-decomposition of $\mathbf{\Pi}^{+}=\mathbf{D}\Sigma\mathbf{D}^{-1}$

We conclude the proof.

\subsection{Corollary~\ref{cor:tau}}
\begin{corollary*}
Let a crystal initially reside in a space group $\widetilde{G}$ and be observed in a lower-symmetry space group $G \prec \widetilde{G}$ at time $t$. Consider a constant transition rate, $\lambda(t) = \lambda \; \forall t$, then the posterior distribution of the holding time $\tau \in [0, t]$ is $\tau = -\frac{1}{\lambda}\log(1 - u(1 - e^{-\lambda t}\bigr)), \;\text{where} \; u \sim \mathcal{U}(0,1)$. Consequently, the symmetry-breaking time window can be calculated as $w = t - \tau$.
\end{corollary*}

\textit{Proof:} Consider a stochastic system with two states, $\widetilde{G}$ and $G$. The system is initialized in state $\widetilde{G}$ and is observed to be in state $G$ at time $t$. The posterior distribution of the holding time $\tau$ (the transition event $\widetilde{G} \rightarrow G$) is derived via Bayes' theorem as follows:

\begin{align}
p(\tau \mid G, \widetilde{G}) = \frac{P(G \mid \tau, \widetilde{G}) \cdot P(\tau \mid \widetilde{G})}{P(G \mid \widetilde{G})}.
\end{align}

The likelihood term equals one if the transition occurred before the observation time $t$:
\begin{align}
P(G \mid \tau, \widetilde{G}) = \mathbf{1}[\tau < t].
\end{align}

The standard survival distribution gives the prior density of the absorption time:
\begin{align}
p(\tau) = P(\tau \mid \widetilde{G}) = \lambda(\tau) \exp\left(-\int_0^\tau \lambda(u) \, du\right).
\end{align}

The evidence (marginal likelihood) evaluates to:
\begin{align}
P(G \mid \widetilde{G}) &= \int_0^\infty P(G \mid \tau) \, p(\tau \mid \widetilde{G}) \, d\tau \nonumber\\
&= \int_0^t 1 \cdot p(\tau) \, d\tau \nonumber\\
&= \int_0^t \lambda(\tau) \exp\left(-\int_0^\tau \lambda(u) \, du\right) d\tau \nonumber\\
&= 1 - \exp\left(-\int_0^t \lambda(u) \, du\right).
\end{align}

Substituting these components, the posterior density is:
\begin{align}
p(\tau \mid G, \widetilde{G}) = \frac{\mathbf{1}[\tau < t] \cdot \lambda(\tau) \exp\left(-\int_0^\tau \lambda(u) \, du\right)}{1 - \exp\left(-\int_0^t \lambda(u) \, du\right)}.
\end{align}

In the specific case of a constant transition rate $\lambda(t) = \lambda$, this simplifies to a truncated exponential distribution:
\begin{align}
p(\tau \mid G, \widetilde{G}) = \frac{\lambda e^{-\lambda \tau}}{1 - e^{-\lambda t}}, \quad \tau \in [0, t].
\end{align}

The corresponding cumulative distribution function (CDF) is:
\begin{align}
F(\tau) = \frac{1 - e^{-\lambda \tau}}{1 - e^{-\lambda t}}.
\end{align}

By applying the inverse transform sampling method, samples of $\tau$ can be generated using:
\begin{align}
\tau = -\frac{1}{\lambda}\log\left(1 - u(1 - e^{-\lambda t})\right), \quad u \sim \mathcal{U}(0,1).
\end{align}
The symmetry-breaking time window is thus defined as $w=t-\tau$.

We conclude the proof.

\subsection{Proposition~\ref{prop:k}}

\begin{proposition*}
Consider a forward diffusion process with a symmetry-breaking event at time $\tau$, transitioning from space group $\widetilde{G}$ to $G$. From Lemma~\ref{lemma:k}, let $(\mathfrak{m}, \mathfrak{m_b})$ and $(\widetilde{\mathfrak{m}}, \mathfrak{\widetilde{m}_b})$ denote the mask--bias pairs associated with $G$ and $\widetilde{G}$, respectively. Then, the continuous-state forward diffusion kernel applied to the lattice parameters $\mathbf{k}$ is given analytically by:
\begin{align}
    \mathbf{k}_t = \widetilde{\mathfrak{m}}\odot&\left(\sqrt{\bar{\alpha}_t}\,\mathbf{k}_0+\sqrt{1-\bar{\alpha}_t}\,\widetilde{\boldsymbol{\epsilon}}\right) +\mathfrak{\widetilde{m}_b} + \mathfrak{m}\odot\mathfrak{\widetilde{m}_b}\sqrt{\frac{\bar{\alpha}_t}{\bar{\alpha}_{t-w_{\mathbf{k}}}}}\nonumber \\
    &-\left(\widetilde{\mathfrak{m}}\odot\sqrt{1-\frac{\bar{\alpha}_t}{\bar{\alpha}_{t-w_{\mathbf{k}}}}}\,\widetilde{\boldsymbol{\epsilon}} + \mathfrak{\widetilde{m}_b}\right) + \left(\mathfrak{m}\odot\sqrt{1-\frac{\bar{\alpha}_t}{\bar{\alpha}_{t-w_{\mathbf{k}}}}}\,\boldsymbol{\epsilon} + \mathfrak{m_b} \right), \nonumber
\end{align}
where $w_{\mathbf{k}} = t - \tau$.
The corresponding reverse-time sampling step is simulated as:
\begin{align}
    \mathbf{k}_{t-1} = \widetilde{\mathfrak{m}}\odot\Bigl(\frac{\sqrt{\bar{\alpha}_{t-1}}}{1-\bar{\alpha}_t}(1-\alpha_t)\,\hat{\mathbf{k}}^{\theta}_0(\mathcal{M}_t) + \mathfrak{m}\odot\frac{\sqrt{\alpha_t}}{1-\bar{\alpha}_t}(1-\bar{\alpha}_{t-1})\,\mathbf{k}_t + \mathfrak{m_b} + \sigma_t\boldsymbol{\epsilon}\Bigr) + \mathfrak{\widetilde{m}_b}. \nonumber
\end{align}
\end{proposition*}

\textit{Proof:} The set of symmetric basis matrices \(\{\mathbf{B}_i\}_{i=1}^6\), derived by \citet{jiao2024space}, is given as follows.

\[ \begin{aligned} \mathbf{B}_1 &= \begin{pmatrix} 0 & 1 & 0 \\ 1 & 0 & 0 \\ 0 & 0 & 0 \end{pmatrix}, & \mathbf{B}_2 &= \begin{pmatrix} 0 & 0 & 1 \\ 0 & 0 & 0 \\ 1 & 0 & 0 \end{pmatrix}, & \mathbf{B}_3 &= \begin{pmatrix} 0 & 0 & 0 \\ 0 & 0 & 1 \\ 0 & 1 & 0 \end{pmatrix}, \\[1em] \mathbf{B}_4 &= \begin{pmatrix} 1 & 0 & 0 \\ 0 & -1 & 0 \\ 0 & 0 & 0 \end{pmatrix}, & \mathbf{B}_5 &= \begin{pmatrix} 1 & 0 & 0 \\ 0 & 1 & 0 \\ 0 & 0 & -2 \end{pmatrix}, & \mathbf{B}_6 &= \begin{pmatrix} 1 & 0 & 0 \\ 0 & 1 & 0 \\ 0 & 0 & 1 \end{pmatrix}. \end{aligned} \]

\label{ap:proofk}
We can cast the space-group-constrained diffusion process over lattice parameters proposed by \citet{jiao2024space} as a masking problem. Below, we describe the lattice masking mechanism applied to the six crystal families, ordered by increasing symmetry.

\begin{align}
    \mathrm{i}) \quad \mathrm{Triclinic:} \quad &\mathfrak{m}=[1,1,1,1,1,1] \qquad \mathfrak{m_b}=[0,0,0,0,0,0] \nonumber\\
    \mathrm{ii}) \quad \mathrm{Monoclinic:} \quad &\mathfrak{m}=[0,1,0,1,1,1] \qquad \mathfrak{m_b}=[0,0,0,0,0,0] \nonumber\\
    \mathrm{iii}) \quad \mathrm{Orthorhombic:} \quad &\mathfrak{m}=[0,0,0,1,1,1] \qquad \mathfrak{m_b}=[0,0,0,0,0,0] \nonumber\\
    \mathrm{iv}) \quad \mathrm{Tetragonal:} \quad &\mathfrak{m}=[0,0,0,0,1,1] \qquad \mathfrak{m_b}=[0,0,0,0,0,0] \nonumber\\
    \mathrm{v}) \quad \mathrm{Hexagonal:} \quad &\mathfrak{m}=[0,0,0,0,1,1] \qquad \mathfrak{m_b}=[-\log\frac{3}{4},0,0,0,0,0] \nonumber\\
    \mathrm{vi}) \quad \mathrm{Cubic:} \quad &\mathfrak{m}=[0,0,0,0,0,1] \qquad \mathfrak{m_b}=[0,0,0,0,0,0]
\end{align}

\paragraph{Masking operation axioms} Let $(\mathfrak{m}, \mathfrak{m}_b)$ and $(\widetilde{\mathfrak{m}}, \widetilde{\mathfrak{m}}_b)$ denote the mask-bias pairs associated with $G$ and $\widetilde{G}$, respectively. Suppose that $\widetilde{G}$ belongs to a crystal family with higher symmetry than that of $G$, denoted by $G \prec\widetilde{G}$. Then the following properties hold:

\begin{align}
    &\mathrm{i}) \quad \widetilde{\mathfrak{m}}\odot\mathfrak{m}=\mathfrak{m}\odot\widetilde{\mathfrak{m}}=\widetilde{\mathfrak{m}} \qquad \mathrm{ii}) \quad \mathfrak{m} \odot \mathfrak{m_b} = \widetilde{\mathfrak{m}}\odot\mathfrak{\widetilde{m}_b}=0\nonumber\\
    &\mathrm{iii}) \quad C \mathfrak{m_b} = \mathfrak{m_b} \quad \text{for any scalar value } C \qquad \mathrm{iv}) \quad \widetilde{\mathfrak{m}} \odot \mathfrak{m_b} = \mathfrak{m_b} \odot \widetilde{\mathfrak{m}} = 0\nonumber\\
\end{align}

\paragraph{Inductive proof}
At $t=\tau$, the diffusion process is in the space group $\widetilde{G}$. Applying Lemma~\ref{lemma:k}, we have:

\begin{align}
    \mathbf{k}_\tau = \widetilde{\mathfrak{m}} \odot \left(\sqrt{\bar{\alpha}_{\tau}}\mathbf{k}_0 + \sqrt{1-\bar{\alpha}_{\tau}}\boldsymbol{\epsilon}_{\tau}\right) + \mathfrak{\widetilde{m}_b}
\end{align}

For $t>\tau$, the diffusion process jumps to the space group $G$. Considering $t=\tau+1$, we have:

\begin{align}
    \mathbf{k}_{\tau+1} &= \mathfrak{m} \odot\left(\sqrt{\alpha_{\tau+1}}\mathbf{k}_{\tau} + \sqrt{1-\alpha_{\tau+1}}\boldsymbol{\epsilon}_{\tau+1}\right) + \mathfrak{m}_b \nonumber \\
    & = \mathfrak{m} \odot\left[\sqrt{\alpha_{\tau+1}}\left(\widetilde{\mathfrak{m}} \odot \left(\sqrt{\bar{\alpha}_{\tau}}\mathbf{k}_0 + \sqrt{1-\bar{\alpha}_{\tau}}\boldsymbol{\epsilon}_{\tau}\right) + \mathfrak{\widetilde{m}_b}\right) + \sqrt{1-\alpha_{\tau+1}}\boldsymbol{\epsilon}_{\tau+1}\right] + \mathfrak{m}_b \nonumber\\
    & = \mathfrak{m} \odot\left[\widetilde{\mathfrak{m}} \odot \left(\sqrt{\bar{\alpha}_{\tau+1}}\mathbf{k}_0 + \sqrt{\alpha_{\tau+1}(1-\bar{\alpha}_{\tau})}\boldsymbol{\epsilon}_{\tau}\right) + \mathfrak{\widetilde{m}_b}\sqrt{\alpha_{\tau+1}}+ \sqrt{1-\alpha_{\tau+1}}\boldsymbol{\epsilon}_{\tau+1}\right] + \mathfrak{m}_b \nonumber\\
    & =^{(\star)} \widetilde{\mathfrak{m}} \odot \left(\sqrt{\bar{\alpha}_{\tau+1}}\mathbf{k}_0 + \sqrt{\alpha_{\tau+1}(1-\bar{\alpha}_{\tau})}\boldsymbol{\epsilon}_{\tau} + \sqrt{1-\alpha_{\tau+1}}\boldsymbol{\epsilon}_{\tau} -  \sqrt{1-\alpha_{\tau+1}}\boldsymbol{\epsilon}_{\tau}\right) \nonumber\\
    & \qquad\qquad\qquad\qquad\qquad\qquad + \mathfrak{m} \odot \mathfrak{\widetilde{m}_b}\sqrt{\alpha_{\tau+1}} + \mathfrak{m} \odot \sqrt{1-\alpha_{\tau+1}}\boldsymbol{\epsilon}_{\tau+1} + \mathfrak{m}_b \nonumber\\
    & =^{(\star\star)} \widetilde{\mathfrak{m}} \odot \left(\sqrt{\bar{\alpha}_{\tau+1}}\mathbf{k}_0 + \sqrt{1-\bar{\alpha}_{\tau+1}}\boldsymbol{\epsilon}_{\tau} \right) + \mathfrak{\widetilde{m}_b} - \mathfrak{\widetilde{m}_b} - \widetilde{\mathfrak{m}} \odot \sqrt{1-\alpha_{\tau+1}}\boldsymbol{\epsilon}_{\tau} \nonumber\\
    & \qquad\qquad\qquad\qquad\qquad\qquad + \mathfrak{m} \odot \mathfrak{\widetilde{m}_b}\sqrt{\alpha_{\tau+1}} + \mathfrak{m} \odot \sqrt{1-\alpha_{\tau+1}}\boldsymbol{\epsilon}_{\tau+1} + \mathfrak{m}_b \nonumber\\
    & = \widetilde{\mathfrak{m}} \odot \left(\sqrt{\bar{\alpha}_{\tau+1}}\mathbf{k}_0 + \sqrt{1-\bar{\alpha}_{\tau+1}}\boldsymbol{\epsilon}_{\tau} \right) +  \mathfrak{\widetilde{m}_b} - \left(\widetilde{\mathfrak{m}} \odot \sqrt{1-\alpha_{\tau+1}}\boldsymbol{\epsilon}_{\tau} + \mathfrak{\widetilde{m}_b}\right)\nonumber\\
    & \qquad\qquad\qquad\qquad\qquad\qquad + \left(\mathfrak{m} \odot \sqrt{1-\alpha_{\tau+1}}\boldsymbol{\epsilon}_{\tau+1} + \mathfrak{m}_b\right) +  \mathfrak{m} \odot \mathfrak{\widetilde{m}_b}\sqrt{\alpha_{\tau+1}} \nonumber\\
    & = \widetilde{\mathfrak{m}} \odot \left(\sqrt{\bar{\alpha}_{\tau+1}}\mathbf{k}_0 + \sqrt{1-\bar{\alpha}_{\tau+1}}\boldsymbol{\epsilon}_{\tau} \right) +  \mathfrak{\widetilde{m}_b} - \left(\widetilde{\mathfrak{m}} \odot \sqrt{1-\frac{\bar{\alpha}_{\tau+1}}{\bar{\alpha}_{\tau}}}\boldsymbol{\epsilon}_{\tau} + \mathfrak{\widetilde{m}_b}\right)\nonumber\\
    & \qquad\qquad\qquad\qquad\qquad\qquad + \left(\mathfrak{m} \odot \sqrt{1-\frac{\bar{\alpha}_{\tau+1}}{\bar{\alpha}_{\tau}}}\boldsymbol{\epsilon}_{\tau+1} + \mathfrak{m}_b\right) +  \mathfrak{m} \odot \mathfrak{\widetilde{m}_b}\sqrt{\frac{\bar{\alpha}_{\tau+1}}{\bar{\alpha}_{\tau}}}
\end{align}

In $(\star)$, we use $\mathfrak{m} \odot \widetilde{\mathfrak{m}} = \widetilde{\mathfrak{m}}$. In $(\star\star)$, we use the fact that the sum of two independent Gaussian random variables with variances $\alpha_{\tau+1}(1-\bar{\alpha}_{\tau})$ and $1-\alpha_{\tau+1}$ is Gaussian with variance $1-\bar{\alpha}_{\tau+1}$.

Similarly, for $t=\tau +2$, we can estimate $\mathbf{k}_{\tau+2}$ as follows:

\begin{align}
    \mathbf{k}_{\tau+2} &= \mathfrak{m} \odot\left(\sqrt{\alpha_{\tau+2}}\mathbf{k}_{\tau+1} + \sqrt{1-\alpha_{\tau+2}}\boldsymbol{\epsilon}_{\tau+2}\right) + \mathfrak{m}_b \nonumber\\
    &=^{(\dag)} \mathfrak{m} \odot \Bigl[\sqrt{\alpha_{\tau+2}}\widetilde{\mathfrak{m}} \odot \left(\sqrt{\bar{\alpha}_{\tau+1}}\mathbf{k}_0 + \sqrt{1-\bar{\alpha}_{\tau+1}}\boldsymbol{\epsilon}_{\tau} \right) \nonumber\\
    &\qquad - \sqrt{\alpha_{\tau+2}}\widetilde{\mathfrak{m}} \odot \sqrt{1-\alpha_{\tau+1}}\boldsymbol{\epsilon}_{\tau}+\sqrt{\alpha_{\tau+2}}\mathfrak{m} \odot \mathfrak{\widetilde{m}_b}\sqrt{\alpha_{\tau+1}} + \sqrt{\alpha_{\tau+2}}\mathfrak{m} \odot \sqrt{1-\alpha_{\tau+1}}\boldsymbol{\epsilon}_{\tau+1} \nonumber\\
    &\qquad + \sqrt{\alpha_{\tau+2}}\mathfrak{m}_b + \sqrt{1-\alpha_{\tau+2}}\boldsymbol{\epsilon}_{\tau+2}\Bigr] + \mathfrak{m}_b \nonumber\\
    &=^{(\dag\dag)} \widetilde{\mathfrak{m}} \odot \left(\sqrt{\bar{\alpha}_{\tau+2}}\mathbf{k}_0 + \sqrt{\alpha_{\tau+2}(1-\bar{\alpha}_{\tau+1})}\boldsymbol{\epsilon}_{\tau}\right) - \widetilde{\mathfrak{m}} \odot \sqrt{\alpha_{\tau+2}(1-\alpha_{\tau+1})}\boldsymbol{\epsilon}_{\tau}  \nonumber\\
    &\qquad + \mathfrak{m} \odot \sqrt{\alpha_{\tau+2}(1-\alpha_{\tau+1})}\boldsymbol{\epsilon}_{\tau+1} +\mathfrak{m} \odot \mathfrak{\widetilde{m}_b}\sqrt{\alpha_{\tau+1}\alpha_{\tau+2}} + \mathfrak{m}\odot \sqrt{1-\alpha_{\tau+2}}\boldsymbol{\epsilon}_{\tau+2} + \mathfrak{m}_b \nonumber\\
    &= \widetilde{\mathfrak{m}} \odot \left(\sqrt{\bar{\alpha}_{\tau+2}}\mathbf{k}_0 + \sqrt{\alpha_{\tau+2}(1-\bar{\alpha}_{\tau+1})}\boldsymbol{\epsilon}_{\tau}\right) + \widetilde{\mathfrak{m}} \odot \sqrt{1-\alpha_{t+2}}\boldsymbol{\epsilon}_{\tau} \nonumber\\
    &\qquad- \widetilde{\mathfrak{m}} \odot \sqrt{1-\alpha_{t+2}}\boldsymbol{\epsilon}_{\tau} - \widetilde{\mathfrak{m}} \odot \sqrt{\alpha_{\tau+2}(1-\alpha_{\tau+1})}\boldsymbol{\epsilon}_{\tau} \nonumber\\
    &\qquad + \mathfrak{m} \odot \sqrt{\alpha_{\tau+2}(1-\alpha_{\tau+1})}\boldsymbol{\epsilon}_{\tau+1} + \mathfrak{m}\odot \sqrt{1-\alpha_{\tau+2}}\boldsymbol{\epsilon}_{\tau+2} \nonumber\\
    &\qquad +\mathfrak{m} \odot \mathfrak{\widetilde{m}_b}\sqrt{\alpha_{\tau+1}\alpha_{\tau+2}} + \mathfrak{m}_b \nonumber\\
    &=^{(\dag\dag\dag)} \widetilde{\mathfrak{m}} \odot \left(\sqrt{\bar{\alpha}_{\tau+2}}\mathbf{k}_0 + \sqrt{(1-\bar{\alpha}_{\tau+2})}\boldsymbol{\epsilon}_{\tau}\right) \nonumber\\
    &\qquad- \widetilde{\mathfrak{m}} \odot \sqrt{(1-\alpha_{\tau+1}\alpha_{\tau+2})}\boldsymbol{\epsilon}_{\tau} \nonumber\\
    &\qquad + \mathfrak{m}\odot \sqrt{1-\alpha_{\tau+1}\alpha_{\tau+2}}\boldsymbol{\epsilon}_{\tau+2} \nonumber\\
    &\qquad +\mathfrak{m} \odot \mathfrak{\widetilde{m}_b}\sqrt{\alpha_{\tau+1}\alpha_{\tau+2}} + \mathfrak{m}_b \nonumber\\
    &= \widetilde{\mathfrak{m}} \odot \left(\sqrt{\bar{\alpha}_{\tau+2}}\mathbf{k}_0 + \sqrt{1-\bar{\alpha}_{\tau+2}}\boldsymbol{\epsilon}_{\tau}\right) + \mathfrak{\widetilde{m}_b}
    - \left( \widetilde{\mathfrak{m}} \odot \sqrt{1-\alpha_{\tau+1}\alpha_{\tau+2}}\boldsymbol{\epsilon}_{\tau} + \mathfrak{\widetilde{m}_b} \right) \nonumber\\
    &\qquad + \left(\mathfrak{m}\odot \sqrt{1-\alpha_{\tau+1}\alpha_{\tau+2}}\boldsymbol{\epsilon}_{\tau+2} + \mathfrak{m}_b \right) +\mathfrak{m} \odot \mathfrak{\widetilde{m}_b}\sqrt{\alpha_{\tau+1}\alpha_{\tau+2}} \nonumber\\
    &= \widetilde{\mathfrak{m}} \odot \left(\sqrt{\bar{\alpha}_{\tau+2}}\mathbf{k}_0 + \sqrt{1-\bar{\alpha}_{\tau+2}}\boldsymbol{\epsilon}_{\tau}\right) + \mathfrak{\widetilde{m}_b}
    - \left( \widetilde{\mathfrak{m}} \odot \sqrt{1-\frac{\bar{\alpha}_{\tau+2}}{\bar{\alpha}_{\tau}}}\boldsymbol{\epsilon}_{\tau} + \mathfrak{\widetilde{m}_b} \right) \nonumber\\
    &\qquad + \left(\mathfrak{m}\odot \sqrt{1-\frac{\bar{\alpha}_{\tau+2}}{\bar{\alpha}_{\tau}}}\boldsymbol{\epsilon}_{\tau+2} + \mathfrak{m}_b \right) +\mathfrak{m} \odot \mathfrak{\widetilde{m}_b}\sqrt{\frac{\bar{\alpha}_{\tau+2}}{\bar{\alpha}_{\tau}}}
\end{align}

In $(\dag)$, we substitute the expression for $\mathbf{k}_{\tau+1}$ obtained in step $(\star)$. In $(\dag\dag)$, we use $\mathfrak{m} \odot \widetilde{\mathfrak{m}} = \widetilde{\mathfrak{m}}$ and $\mathfrak{m} \odot \mathfrak{m_b} = 0$. In $(\dag\dag\dag)$, we use the fact that the sum of two independent Gaussian random variables with variances $\alpha_{\tau+2}(1-\bar{\alpha}_{\tau+1})$ and $1-\alpha_{\tau+2}$ is Gaussian with variance $1-\bar{\alpha}_{\tau+2}$, and that the sum of two independent Gaussian random variables with variances $\alpha_{\tau+2}(1-\alpha_{\tau+1})$ and $1-\alpha_{\tau+2}$ is Gaussian with variance $1-\alpha_{\tau+1}\alpha_{\tau+2}$.
For a general case $t > \tau$, and $w=t-\tau$, we have:

\begin{align}
    \mathbf{k}_{t} &= \widetilde{\mathfrak{m}} \odot \left(\sqrt{\bar{\alpha}_{t}}\mathbf{k}_0 + \sqrt{1-\bar{\alpha}_{t}}\boldsymbol{\epsilon}_{\tau}\right) + \mathfrak{\widetilde{m}_b}
    - \left( \widetilde{\mathfrak{m}} \odot \sqrt{1-\frac{\bar{\alpha}_{t}}{\bar{\alpha}_{t-w}}}\boldsymbol{\epsilon}_{\tau} + \mathfrak{\widetilde{m}_b} \right) \nonumber\\
    &\qquad + \left(\mathfrak{m}\odot \sqrt{1-\frac{\bar{\alpha}_{t}}{\bar{\alpha}_{t-w}}}\boldsymbol{\epsilon}_{t} + \mathfrak{m}_b \right) +\mathfrak{m} \odot \mathfrak{\widetilde{m}_b}\sqrt{\frac{\bar{\alpha}_{t}}{\bar{\alpha}_{t-w}}}
\end{align}

Sampling is performed using DDIM samplers from \citet{song2020denoising}. The reverse process proceeds toward increasing symmetry levels, from $G$ to $\widetilde{G}$.
\begin{align}
    \textbf{k}_{t-1}&=\mathfrak{\widetilde{m}}\odot\left(\frac{\sqrt{\bar{\alpha}_{t-1}}}{1-\bar{\alpha_t}}\left(1-\alpha_t\right)\mathbf{\hat{k}}^{\theta}_0(\mathcal{M}_t)+(\mathfrak{m}\odot\mathbf{k}_t + \mathfrak{m_b})\frac{\sqrt{\alpha_t}}{1-\bar{\alpha}_t}\left(1-\bar{\alpha}_{t-1}\right) + \sigma_t\boldsymbol{\epsilon}\right) + \mathfrak{\widetilde{m}_b} \nonumber\\
    &=\mathfrak{\widetilde{m}}\odot\left(\frac{\sqrt{\bar{\alpha}_{t-1}}}{1-\bar{\alpha_t}}\left(1-\alpha_t\right)\mathbf{\hat{k}}^{\theta}_0(\mathcal{M}_t)+\mathfrak{m}\odot\frac{\sqrt{\alpha_t}}{1-\bar{\alpha}_t}\left(1-\bar{\alpha}_{t-1}\right)\mathbf{k}_t + \mathfrak{m_b} + \sigma_t\boldsymbol{\epsilon}\right) + \mathfrak{\widetilde{m}_b}
\end{align}
We apply $C\mathfrak{m_b}=\mathfrak{m_b}$ for any scalar value $C$,i.e., the biases remain constant across crystal families, and $\mathbf{\hat{k}}^{\theta}_0(\mathcal{M}_t)$ denotes the data-prediction parameterization of a neural network.

We conclude the proof.

\subsection{Proposition~\ref{prop:marginal_s}}
\label{ap:proofs}
\begin{proposition*}
  Let $\mathfrak{\widetilde{g}}, \mathfrak{g} \in \Delta^{81-1}$ be the marginal site-symmetry priors associated with the space groups $\widetilde{G}$ and $G$, respectively. Let $\bar{\alpha}_t$ be a strictly decreasing noise schedule with symmetry-breaking time $\tau$ satisfying $0<\bar{\alpha}_t, \bar{\alpha}_{\tau} \leq 1$. For $t>\tau$, define
    \begin{align}
        \boldsymbol{\pi}_t = \nu_t\mathfrak{\widetilde{g}} + (1-\nu_t)\mathfrak{g}=\frac{\bar{\alpha}_t(1-\bar{\alpha}_{t- w_{\mathbf{S}}})\mathfrak{\widetilde{g}} + (\bar{\alpha}_{t- w_{\mathbf{S}}}-\bar{\alpha}_t)\mathfrak{g}}{\bar{\alpha}_{t- w_{\mathbf{S}}} (1- \bar{\alpha}_t)} \mathrm{\quad where\;} w_{\mathbf{S}}=t -\tau,\; \nu_t \in [0,1] \nonumber
    \end{align}
    Then, $\boldsymbol{\pi}_t$ lies in the convex hull of the two priors, $\boldsymbol{\pi}_{\tau}=\mathfrak{\widetilde{g}},\; \lim_{t\rightarrow T}\boldsymbol{\pi}_{t}=\mathfrak{g}$ , and the trajectory $\{\boldsymbol{\pi}_t\}_{t>\tau}$ moves monotonically along the straight line segment connecting $\mathfrak{\widetilde{g}}$ to $\mathfrak{g}$ inside the simplex.
\end{proposition*}

\textit{Proof:} We introduce a \textit{non-stationary prior mixture process} that defines \textit{a forward process} which changes its target distribution midway through corruption.
Let denote $\widetilde{\mathfrak{g}}$ and $\mathfrak{g}$ the marginal site symmetry prior of space group $\widetilde{G}$ and $G$, respectively. We still assume that $\widetilde{G}$ is a higher-symmetry space group than $G$, i.e., $G \prec \widetilde{G}$, and recall that $\tau$ is the symmetry-breaking event; we rewrite the forward categorical diffusion from \citet{austin2021structured} as follows.
\begin{align}
    q\left(\mathbf{S}_{t}|\mathbf{S}_{0}, \tau, G, \widetilde{G}\right) &= \mathrm{Cat}\left(\mathbf{S}_{t}; \mathbf{S}_{0}\bar{\mathbf{Q}}_{t,\tau}^{\widetilde{G} \to G}\right)
\end{align}

We have the following transition matrices within a space group.
\begin{align}
    \bar{Q}_{\tau\leftarrow0}^{\widetilde{G}} = \bar{\alpha}_{\tau}I + (1-\bar{\alpha}_{\tau})\mathbf{1}({\widetilde{\mathfrak{g}}})^T
\end{align}
\begin{align}
    \bar{Q}_{t\leftarrow\tau}^{G} = \frac{\bar{\alpha}_{t}}{\bar{\alpha}_{\tau}}I + (1-\frac{\bar{\alpha}_{t}}{\bar{\alpha}_{\tau}})\mathbf{1}({\mathfrak{g}})^T
\end{align}

Then,
\begin{align}
    \bar{\mathbf{Q}}_{t}^{\widetilde{G} \to G} = \bar{Q}_{\tau\leftarrow0}^{\widetilde{G}}\bar{Q}_{t\leftarrow\tau}^{G} &= \Bigl(\bar{\alpha}_{\tau}I + (1-\bar{\alpha}_{\tau})\mathbf{1}({\widetilde{\mathfrak{g}}})^T\Bigr)\Bigl(\frac{\bar{\alpha}_{t}}{\bar{\alpha}_{\tau}}I + (1-\frac{\bar{\alpha}_{t}}{\bar{\alpha}_{\tau}})\mathbf{1}({\mathfrak{g}})^T\Bigr) \nonumber\\
    &= \bar{\alpha}_tI + (\bar{\alpha}_{\tau} - \bar{\alpha}_t)\mathbf{1}({\mathfrak{g}})^T + (1-\bar{\alpha}_{\tau})\frac{\bar{\alpha}_{t}}{\bar{\alpha}_{\tau}}\mathbf{1} ({\widetilde{\mathfrak{g}}})^T \nonumber\\
    &+ (1-\bar{\alpha}_{\tau})(1-\frac{\bar{\alpha}_{t}}{\bar{\alpha}_{\tau}})\mathbf{1} ({\mathfrak{g}})^T, \nonumber \qquad \text{as} \quad ({\widetilde{\mathfrak{g}}})^T\mathbf{1}=1 \\
    &=\bar{\alpha}_tI + \frac{1}{\bar{\alpha}_\tau}\Bigr[\bar{\alpha}_t(1-\bar{\alpha}_\tau)\mathbf{1}(\mathfrak{\widetilde{\mathfrak{g}}})^T + (\bar{\alpha}_\tau-\bar{\alpha}_t)\mathbf{1}(\mathfrak{g})^T \Bigl] \nonumber \\
     &= \bar{\alpha}_tI + (1-\bar{\alpha}_t)\mathbf{1}(\boldsymbol{\pi}_t)^T \nonumber\\
\text{where} \qquad & \boldsymbol{\pi}_t = \frac{\bar{\alpha}_t(1-\bar{\alpha}_\tau)\mathfrak{\widetilde{\mathfrak{g}}} + (\bar{\alpha}_\tau-\bar{\alpha}_t)\mathfrak{g}}{\bar{\alpha}_\tau (1- \bar{\alpha}_t)}
\end{align}

 $\boldsymbol{\pi}_t$ is a symmetry-breaking (SB) prior that smoothly interpolates between two given site-symmetry priors. %There are two extreme edge cases, $t \rightarrow T, \bar{\alpha}_t \rightarrow 0$, then $\boldsymbol{\pi}_t = \mathfrak{g}$, the site-symmetry prior of the target space group $G$, and $t<\tau$, there is no space group transition, i.e., $\mathfrak{g} = \widetilde{\mathfrak{g}}$, thus $\boldsymbol{\pi}_t = \mathfrak{\widetilde{\mathfrak{g}}}$.

\textit{Convexity} 

We rewrite the SB prior as a convex hull of the two priors
\begin{align}
    \boldsymbol{\pi}_t = \nu_t\mathfrak{\widetilde{\mathfrak{g}}} + (1-\nu_t)\mathfrak{g}, \qquad \text{with} \quad 0\leq \nu_t = \frac{\bar{\alpha}_t(1-\bar{\alpha}_{\tau})}{\bar{\alpha}_{\tau}(1-\bar{\alpha}_t)} \leq 1
\end{align}  

The convexity property ensures that $\boldsymbol{\pi}_t$ is a valid categorical distribution that lies within the probability simplex. The forward kernel thus remains row-stochastic and requires no renormalization trick. 

\textit{Boundary consistency}

For $t\leq\tau$, there is no transition between priors, i.e. $\mathfrak{g}=\widetilde{\mathfrak{g}}$, thus $
\boldsymbol{\pi}_{t\leq\tau} = \mathfrak{\widetilde{\mathfrak{g}}}
$. And for $t\rightarrow T, \bar{\alpha}_t \rightarrow 0, \nu_t \rightarrow0$, which leads to $\lim_{t\rightarrow T}  \boldsymbol{\pi}_t=\mathfrak{g}$. And, the path connects two priors continuously. 

\textit{Monotone geometric path}

By differentiating $\boldsymbol{\pi}_{t}$ w.r.t time $t$, we have

\begin{align}
    \frac{d\boldsymbol{\pi}_{t}}{dt} = \frac{d\nu_t}{dt}(\mathfrak{\widetilde{\mathfrak{g}}}-\mathfrak{g})
\end{align}

Since $\bar{\alpha}_t$ is decreasing, $\nu_t$ is decreasing, thus $\frac{d\nu_t}{dt} <0$ and the path moves monotonically toward $\mathfrak{g}$.

We conclude the proof.

\subsection{Corollary~\ref{coro:s}}

\begin{corollary*}
Let the discrete-state diffusion transition matrices of \citet{austin2021structured} be redefined under the symmetry-breaking formulation as:
\begin{align}
\mathbf{Q}_{t} = \alpha_tI + (1-\alpha_t)\mathbf{1}(\mathfrak{g})^T, \qquad
 \mathbf{\bar{Q}}_{t-1} = \bar{\alpha}_{t-1}I + (1-\bar{\alpha}_{t-1})\mathbf{1}(\mathfrak{\widetilde{g}})^T \nonumber \\
  \bar{\mathbf{Q}}_{t}^{\widetilde{G} \to G} = \bar{\alpha}_tI + (1-\bar{\alpha}_t)\mathbf{1}(\boldsymbol{\pi}_t)^T \mathrm{\; with\;} \mathfrak{g}, \mathfrak{\widetilde{g}}, \boldsymbol{\pi}_t \mathrm{\; in\; Proposition~\ref{prop:marginal_s}\;} \nonumber
\end{align}
Then, the forward and reverse sampling processes on the site symmetry $\mathbf{S}$ can be derived as follows:
\begin{align}
\mathbf{S}_{t} \sim \mathrm{Cat}\left(\mathbf{S}_{t}; \mathbf{S}_{0}\bar{\mathbf{Q}}_{t}^{\widetilde{G} \to G}\right) \qquad \mathbf{S}_{t-1} \sim \mathrm{Cat}\left(\mathbf{S}_{t-1}; \frac{\mathbf{S}_{t}\mathbf{Q}^T_{t}\odot\mathbf{\hat{S}}_{0}^{\theta}(\mathcal{M}_t)\mathbf{\bar{Q}}_{t-1}}{\mathbf{\hat{S}}_{0}^{\theta}(\mathcal{M}_t)\bar{\mathbf{Q}}_{t}^{\widetilde{G} \to G} \mathbf{S}_{t}^T}\right) \nonumber 
\end{align}
\end{corollary*}

\textit{Proof:} The forward noise kernel follows from the proof in Appendix~\ref{ap:proofs}, which we recall below:
\begin{align}
    & q\left(\mathbf{S}_{t}|\mathbf{S}_{0}, \tau, G, \widetilde{G}\right) = \mathrm{Cat}\left(\mathbf{S}_{t}; \mathbf{S}_{0}\bar{\mathbf{Q}}_{t,\tau}^{\widetilde{G} \to G}\right) \nonumber\\
    & \mathrm{where\;} \bar{\mathbf{Q}}_{t}^{\widetilde{G} \to G} =  \bar{\alpha}_tI + (1-\bar{\alpha}_t)\mathbf{1}(\boldsymbol{\pi}_t)^T \nonumber\\
    & \mathrm{and\;} \boldsymbol{\pi}_t = \frac{\bar{\alpha}_t(1-\bar{\alpha}_\tau)\mathfrak{\widetilde{\mathfrak{g}}} + (\bar{\alpha}_\tau-\bar{\alpha}_t)\mathfrak{g}}{\bar{\alpha}_\tau (1- \bar{\alpha}_t)} =\nu_t\mathfrak{\widetilde{\mathfrak{g}}} + (1-\nu_t)\mathfrak{g}, \text{\;with} \quad 0\leq \nu_t = \frac{\bar{\alpha}_t(1-\bar{\alpha}_{\tau})}{\bar{\alpha}_{\tau}(1-\bar{\alpha}_t)} \leq 1
\end{align}

We rewrite the reverse sampling kernel of \citet{austin2021structured} under the symmetry-breaking formulation, in which the process proceeds toward higher symmetry orders in crystals. Specifically, we assume that the process jumps from space group $G_t=G$ to space group $G_{t-1}=\widetilde{G}$, where $G \prec \widetilde{G}$. The true posterior can be written as
\begin{align}
q(\mathbf{S}_{t-1}\mid \mathbf{S}_{t}, \mathbf{S}_{0}, G_{t-1}, G_t)
=
\frac{
q(\mathbf{S}_t\mid \mathbf{S}_{t-1}, G_t)
q(\mathbf{S}_{t-1}\mid \mathbf{S}_{0},G_{t-1})
}{
q(\mathbf{S}_t\mid \mathbf{S}_0,G_t)
}.
\end{align}

We then parameterize the model posterior as follows:
\begin{align}
p_{\theta}(\mathbf{S}_{t-1}\mid \mathcal{M}_t, G_{t-1})
=
\frac{
q(\mathbf{S}_t\mid \mathbf{S}_{t-1}, G_t)
\tilde{p}_{\theta}(\mathbf{S}_{t-1}\mid \mathcal{M}_t, G_{t-1})
}{
\sum_{\mathbf{S}_{t-1}}
q(\mathbf{S}_t\mid \mathbf{S}_{t-1}, G_t)
\tilde{p}_{\theta}(\mathbf{S}_{t-1}\mid \mathcal{M}_t, G_{t-1})
}.
\end{align}

This defines a valid distribution for any non-negative, unnormalized prior $\tilde{p}_{\theta}$ over $\mathbf{S}_{t-1}$. The true posterior is recovered by setting
\begin{align}
\tilde{p}_{\theta}(\cdot\mid \mathcal{M}_t, G_{t-1})
=
q(\cdot\mid \mathbf{S}_{0}, G_{t-1}).
\end{align}
Since $q(\mathbf{S}_{t-1}\mid \mathbf{S}_{0}, G_{t-1})$ is a closed-form forward kernel, we parameterize the model to predict $\mathbf{S}_0$, denoted by $\hat{\mathbf{S}}_{0}^{\theta}(\mathcal{M}_t)$. Then,
\begin{align}
\tilde{p}_{\theta}(\cdot\mid \mathcal{M}_t, G_{t-1})
=
q(\cdot\mid \hat{\mathbf{S}}_{0}^{\theta}(\mathcal{M}_t), G_{t-1}).
\end{align}

Thus, we can parameterize the model posterior as:
\begin{align}
    p_{\theta}(\mathbf{S}_{t-1}\mid \mathcal{M}_t, G_{t-1})
=
\frac{
q(\mathbf{S}_t\mid \mathbf{S}_{t-1}, G_t)
q(\mathbf{S}_{t-1}\mid \hat{\mathbf{S}}_{0}^{\theta}(\mathcal{M}_t), G_{t-1})
}{
\sum_{\mathbf{S}_{t-1}}
q(\mathbf{S}_t\mid \mathbf{S}_{t-1}, G_t)
q(\mathbf{S}_{t-1}\mid \hat{\mathbf{S}}_{0}^{\theta}(\mathcal{M}_t), G_{t-1})
}.
\end{align}

Alternatively, we can rewrite the model posterior using the corresponding space groups defined above.

\begin{align}
p_{\theta}(\mathbf{S}_{t-1}\mid \mathcal{M}_t, \widetilde{G})
=
\frac{
q(\mathbf{S}_t\mid \mathbf{S}_{t-1}, G)
q(\mathbf{S}_{t-1}\mid \hat{\mathbf{S}}_{0}^{\theta}(\mathcal{M}_t), \widetilde{G})
}{
\sum_{\mathbf{S}_{t-1}}
q(\mathbf{S}_t\mid \mathbf{S}_{t-1}, G)
q(\mathbf{S}_{t-1}\mid \hat{\mathbf{S}}_{0}^{\theta}(\mathcal{M}_t), \widetilde{G})
}.
\end{align}

We now analyze each component:

(i)
\begin{align}
    &q(\mathbf{S}_t\mid \mathbf{S}_{t-1}, G), \mathrm{\; equivalent\; to.\;} \mathbf{S}_t \sim \mathrm{Cat}(\mathbf{S}_t; \mathbf{S}_{t-1}\mathbf{Q}_t) \Rightarrow \mathbf{S}_{t-1} \sim \mathrm{Cat}( \mathbf{S}_{t-1};  \mathbf{S}_{t}\mathbf{Q}_t^T) \nonumber\\
    & \mathrm{where\quad} \mathbf{Q}_{t} = \alpha_tI + (1-\alpha_t)\mathbf{1}(\mathfrak{g})^T
\end{align}

(ii)
\begin{align}
    & q(\mathbf{S}_{t-1}\mid \hat{\mathbf{S}}_{0}^{\theta}(\mathcal{M}_t), \widetilde{G}) \mathrm{\; equivalent\; to.\;} \mathbf{S}_{t-1} \sim \mathrm{Cat}(\mathbf{S}_{t-1}; \hat{\mathbf{S}}_{0}^{\theta}(\mathcal{M}_t)\mathbf{\bar{Q}}_{t-1})\nonumber\\
    & \mathrm{where\quad} \mathbf{\bar{Q}}_{t-1} = \bar{\alpha}_tI + (1-\bar{\alpha}_t)\mathbf{1}(\mathfrak{\widetilde{g}})^T
\end{align}

(iii) The scalar normalization term.
\begin{align}
    &\sum_{\mathbf{S}_{t-1}}
q(\mathbf{S}_t\mid \mathbf{S}_{t-1}, G)
q(\mathbf{S}_{t-1}\mid \hat{\mathbf{S}}_{0}^{\theta}(\mathcal{M}_t), \widetilde{G}) \nonumber\\
&\mathrm{with\quad} q(\mathbf{S}_t\mid \mathbf{S}_{t-1}, G) \Rightarrow  \mathbf{S}_{t-1} \sim \mathrm{Cat}(\mathbf{S}_{t-1}; \mathbf{S}_{t}\mathbf{Q}_t^T) \nonumber\\
&\mathrm{and\quad} q(\mathbf{S}_{t-1}\mid \hat{\mathbf{S}}_{0}^{\theta}(\mathcal{M}_t), \widetilde{G}) \Rightarrow \mathbf{S}_{t-1} \sim \mathrm{Cat}(\mathbf{S}_{t-1}; \hat{\mathbf{S}}_{0}^{\theta}(\mathcal{M}_t)\mathbf{\bar{Q}}_{t-1}) \nonumber\\
&\mathrm{\text{the outer sum over $\mathbf{S}_{t-1}$ can be written as an inner product as follows:}}\nonumber\\
&\qquad \hat{\mathbf{S}}_{0}^{\theta}(\mathcal{M}_t)\mathbf{\bar{Q}}_{t-1} (\mathbf{S}_{t}\mathbf{Q}_t^T)^T = \hat{\mathbf{S}}_{0}^{\theta}(\mathcal{M}_t)\mathbf{\bar{Q}}_{t-1}\mathbf{Q}_t\mathbf{S}^T_{t}=\mathbf{\hat{S}}_{0}^{\theta}(\mathcal{M}_t)\bar{\mathbf{Q}}_{t}^{\widetilde{G} \to G} \mathbf{S}_{t}^T
\end{align}

From (i), (ii), (iii), we obtain the reverse sampling process on the site symmetry $\mathbf{S}$ as:

\begin{align}
    \mathbf{S}_{t-1} \sim \mathrm{Cat}\left(\mathbf{S}_{t-1}; \frac{\mathbf{S}_{t}\mathbf{Q}^T_{t}\odot\mathbf{\hat{S}}_{0}^{\theta}(\mathcal{M}_t)\mathbf{\bar{Q}}_{t-1}}{\mathbf{\hat{S}}_{0}^{\theta}(\mathcal{M}_t)\bar{\mathbf{Q}}_{t}^{\widetilde{G} \to G} \mathbf{S}_{t}^T}\right)
\end{align}

We conclude the proof.

\section{Experimental setup}
\label{app:exp}

\subsection{Oriented site-symmetry symbols}
\label{app:orient_ss}
We provide the list of oriented site-symmetry symbols used for one-hot encoding of site symmetries.

\texttt{\{-3.: 0, ..m: 1, 6/mm2/m: 2, m.mm: 3, ..2: 4, 4/mm.m: 5, .m.: 6, .-3m: 7, 32.: 8, mm2..: 9, \
    -6mm2m: 10, -3..: 11, 2.22: 12, mmm..: 13, 3..: 14, m.m2: 15, mmm.: 16, -43m: 17, 6/m..: 18, 432: 19,\
    .2.: 20, m..: 21, 4..: 22, 2mm.: 23, 4/mmm: 24, m2.: 25, 23.: 26, 222: 27, m.2m: 28, m-3m: 29, 2m.: 30,\
    3mm: 31, -3m2/m: 32, .3m: 33, 222.: 34, -32/m: 35, 3m.: 36, 222..: 37, 622: 38, -4m2: 39, 22.: 40, -4..: 41,\
    2: 42, -4m.2: 43, mmm: 44, 3m: 45, .m: 46, -622m2: 47, 2/m2/m.: 48, 32: 49, ..2/m: 50, 2mm: 51, -6..: 52, 4m.m: 53,\
    2.mm: 54, .32: 55, 6mm: 56, m: 57, 422: 58, mm2: 59, 4mm: 60, 1: 61, -42m: 62, m-3.: 63, 2/m..: 64, 4/m..: 65, -42.m: 66,\
    42.2: 67, m2m.: 68, mm.: 69, 6..: 70, 2/m: 71, -32/m.: 72, .2/m.: 73, .-3.: 74, 322: 75, .3.: 76, 3.: 77, 2..: 78, m2m: 79,\
    -1: 80\}}

\subsection{Stable, uniqueness, and novelty (S.U.N) metrics}
Generated crystalline materials should ideally be thermodynamically stable. To assess stability, we calculate their energies and compare them against competing phases on a convex hull constructed from the lowest-energy materials at each composition. Following prior work \citep{levy2025symmcd, miller2024flowmm}, we utilize a pretrained CHGNet model \citep{deng2023chgnet} to estimate crystal energies and use the convex hull derived from the Materials Project \citep{riebesell2023matbench}. Among the stable generated structures, we compute the number of unique stable structures and assess their novelty with respect to the training dataset. For uniqueness and novelty evaluation, we utilize $\mathrm{StructureMatcher}$ \citep{ong2013python} to determine structural similarity.

\subsection{Wyckoff position retrieval}
Since the proposed site-symmetry representation directly maps to a valid point group, we do not need to project the generated site symmetries onto the closest point group, unlike in \citet{levy2025symmcd}. However, multiple Wyckoff positions may share the same underlying oriented site-symmetry symbol, even though they correspond to different sets of symmetry-equivalent sites. As a result, the oriented site-symmetry symbol alone is not sufficient to uniquely determine the appropriate Wyckoff position for a generated fractional coordinate. We therefore compare the generated coordinate against all candidate Wyckoff positions of a given space group with the matching oriented site symmetry and select the one whose symmetry-equivalent sites lie closest to the generated coordinate. In practice, this nearest-site search is performed using the $\mathrm{search\_closest\_wp}$ function from $\mathrm{PyXtal}$ \citep{fredericks2021pyxtal}, and distances are evaluated under periodic boundary conditions in fractional-coordinate space.

\subsection{Architecture}
We adopt the equivariant graph neural network-based denoiser from DiffCSP/++ \citep{jiao2023crystal,jiao2024space}. For site-symmetry prediction, $\text{SbCD}^{\sharp}$ follows the approach from \citet{levy2025symmcd}, which predicts the probabilities of 13 symmetry elements over each of 15 possible axes, since both employ the same site-symmetry representation.
In contrast, SbCD and $\text{SbCD}^{\diamondsuit}$ modify the existing architecture to predict probabilities over 81 possible oriented site-symmetry symbols only. For space groups, we also use a one-hot encoding over 230 discrete space groups instead of the complex space-group encoding scheme proposed by \citet{levy2025symmcd}.

\begin{table*}[th]
\caption{Hyperparameters used in the main experiments of Table~\ref{tab:res} on MP-20.}
\label{tab:hyper}
\begin{small}
\begin{tabular}{l ccc}
\toprule
 Hyperparameter & SbCD &  $\text{SbCD}^{\sharp}$ & $\text{SbCD}^{\diamondsuit}$  \\
\midrule
  $\mathrm{numLayer}$ & $6$ & $8$ & $6$ \\
  $\mathrm{hiddenDim}$ & $1024$ & $1024$ & $512$  \\
  $\mathrm{embedDim}$ & $512$ & $512$ & $256$  \\
  $\mathrm{learnRate}$ & $1\times10^{-3}$ & $1\times10^{-3}$ & $1\times10^{-3}$ \\
  $\mathrm{optim}$ & $\mathrm{Adam}$  & $\mathrm{Adam}$  & $\mathrm{Adam}$ \\
  $\mathrm{scheduler}$ & $\mathrm{ReduceLROnPlateau}$  & $\mathrm{ReduceLROnPlateau}$  & $\mathrm{ReduceLROnPlateau}$ \\
  $\mathrm{batchSize}$ &  $256$   & $256$  &  $256$  \\
  $\mathrm{numEpoch}$ &  $3000$   & $3000$  &  $3000$  \\
  $\lambda$ & $2\times10^{-3}$ & $1.5\times10^{-3}$ &  $1.5\times10^{-3}$ \\
  $l_{\mathbf{F}}, l_{\mathbf{k}}, l_{\mathbf{A}}, l_{\mathbf{S}}, l_G, l_{\mathrm{jump}}$ & $\{3, 10, 1,1, 4, 1\}$ & $\{3, 10, 1,12, 4, 1\}$ & $\{3, 10, 1,1, 4, 1\}$ \\
\bottomrule
\end{tabular}
\end{small}
\end{table*}
\begin{table*}[th]
\caption{Hyperparameters used in the main experiments of Table~\ref{tab:res_mpts52} on MPTS-52.}
\label{tab:hyper_mpts52}
\begin{small}
\begin{tabular}{l ccc}
\toprule
 Hyperparameter & SbCD-S&  SbCD-L & SbCD-XL  \\
\midrule
  $\mathrm{numLayer}$ & $6$ & $6$ & $8$ \\
  $\mathrm{hiddenDim}$ & $512$ & $1024$ & $1024$  \\
  $\mathrm{embedDim}$ & $256$ & $512$ & $512$  \\
  $\mathrm{learnRate}$ & $1\times10^{-3}$ & $1\times10^{-3}$ & $1\times10^{-3}$ \\
  $\mathrm{optim}$ & $\mathrm{Adam}$  & $\mathrm{Adam}$  & $\mathrm{Adam}$ \\
  $\mathrm{scheduler}$ & $\mathrm{ReduceLROnPlateau}$  & $\mathrm{ReduceLROnPlateau}$  & $\mathrm{ReduceLROnPlateau}$ \\
  $\mathrm{batchSize}$ &  $256$   & $256$  &  $256$  \\
  $\mathrm{numEpoch}$ &  $2000$   & $2000$  &  $2000$  \\
  $\lambda$ & $2\times10^{-3}$ & $2\times10^{-3}$ &  $2\times10^{-3}$ \\
  $l_{\mathbf{F}}, l_{\mathbf{k}}, l_{\mathbf{A}}, l_{\mathbf{S}}, l_G, l_{\mathrm{jump}}$ & $\{3, 10, 1,1, 4, 1\}$ & $\{3, 10, 1,1, 4, 1\}$ & $\{3, 10, 1,1, 4, 1\}$ \\
\bottomrule
\end{tabular}
\end{small}
\end{table*}

\subsection{Hyperparameter search}
We conduct a hyperparameter search over architectural parameters, including the number of layers, $\mathrm{numLayer} \in \{6,8\}$, the hidden dimension, $\mathrm{hiddenDim} \in \{512, 1024\}$, and the symmetry embedding dimension (for site symmetry and space groups), $\mathrm{embedDim} \in \{256, 512\}$. Moreover, we ablate the learning rate, $\mathrm{learnRate} \in \{5\times10^{-4}, 1\times10^{-3}\}$, and the transition rate, $\lambda \in \{0.8\times10^{-3}, 1\times10^{-3}, 1.5\times10^{-3}, 2\times10^{-3}\}$. For a fair comparison, we set the diffusion schedules \citep{austin2021structured, ho2020denoising} for $\mathbf{k}, \mathbf{S}, \mathbf{A},$ and $\mathbf{F}$ similarly to \citet{levy2025symmcd}. Tables~\ref{tab:hyper} and~\ref{tab:hyper_mpts52} summarize the configuration used in the main experiments.
% including loss weights

For the $\text{SymmCD}^{\ddag}$ baselines reported in Tables~\ref{tab:res} and~\ref{tab:res_mpts52}, we retrained the models using the hyperparameters and experimental settings recommended by \citet{levy2025symmcd}, as no official pretrained model weights are publicly available.

\subsection{Computational analysis}
SbCD variants utilize the asymmetric-unit structure as \citet{levy2025symmcd}, which is notably memory efficient for crystal generation. Moreover, SbCD and $\text{SbCD}^{\diamondsuit}$ employ a simplified site-symmetry representation, further reducing memory overhead and improving training efficiency. Table~\ref{tab:compute} compares the computational analysis of these models at a batch size of 256 on an NVIDIA GeForce RTX 3060.
\begin{table}[t]
\caption{Comparison of memory usage, training time per epoch, and architecture size between SbCD variants at a batch size of 256 on an NVIDIA GeForce RTX 3060 with 20 CPU cores.}
\label{tab:compute}
\centering
\resizebox{.85\linewidth}{!}{
\begin{tabular}{l ccc}
\toprule
Method & Memory(MB)$(\downarrow$) & Training time per epoch (s)($\downarrow$)  & Number of parameters (M)($\downarrow$)\\
\midrule
% SymmCD \citep{levy2025symmcd}& $7731$ & $81$ & $61.5$\\
$\text{SbCD}^{\sharp}$ & $7731$ & $81$ & $61.5$\\
$\text{SbCD}^{\diamondsuit}$ &  $4323$ & $34$ & $14.8$\\
SbCD & $6775$ &  $62$ & $47.9$ \\
\bottomrule
\end{tabular}
}
\end{table}

\subsection{Training and sampling algorithms}
We provide the training and sampling procedures of SbCD in Algorithms~\ref{alg:trainSbCD} and~\ref{alg:sampleSbCD}, respectively. Notably, SbCD initiates the sampling process from the minimal symmetry prior, namely the space group $\mathrm{P1}$ with its trivial site symmetry $\mathrm{1}$.

\begin{algorithm}[h]
\caption{Sampling of SbCD}
\begin{algorithmic}[1]
    \Require Trained model $\theta$.
    \State ${\color{BlueGreen} G_T} \leftarrow \mathrm{P1}$ \Comment{Lowest symmetry space group}
    \State $\mathbf{S}_T \leftarrow \mathbf{1}$ \Comment{Trivial site symmetry of $\mathrm{P1}$}
    \State $\mathbf{k}_T \sim \mathcal{N}(0,I),\quad \mathbf{F}_T \sim \mathcal{U}(0,I),\quad \mathbf{A}_T \sim p_{\mathrm{marginal}}(\mathrm{A})$
    \For{t = T to 1}
    \State ${\color{VioletRed} G_{t-1}} = {\color{BlueGreen} G_t} + \sum_{i=1}^{230} \mathbf{P}_i \Bigl(G_{i}-{\color{BlueGreen} G_t}\Bigr) \; \mathrm{with} \; \mathbf{P}_i = \mathrm{Poisson}\Bigl(\hat{\Lambda}^{\theta}_t({\color{BlueGreen} G_t},G_{i})\Bigr)$  \Comment{Lemma~\ref{lem:tau}}
    \State{$/\ast\mathrm{Proposition~\ref{prop:k}}\ast/$}
    \State $\mathbf{k}_{t-1} \leftarrow \mathfrak{{\color{VioletRed}m}}\odot\left(\frac{\sqrt{\bar{\alpha}_{t-1}}}{1-\bar{\alpha_t}}\left(1-\alpha_t\right)\mathbf{\hat{k}}^{\theta}_0(\mathcal{M}_t)+\mathfrak{{\color{BlueGreen}m}}\odot\frac{\sqrt{\alpha_t}}{1-\bar{\alpha}_t}\left(1-\bar{\alpha}_{t-1}\right)\mathbf{k}_t + \mathfrak{{\color{BlueGreen}m_b}} + \sigma_t\boldsymbol{\epsilon}\right) + \mathfrak{{\color{VioletRed}m_b}}$
    \State $
        \mathbf{S}_{t-1} \sim \mathrm{Cat}\left(\mathbf{S}_{t-1}; \frac{\mathbf{S}_{t}\mathbf{Q}^T_{t}\odot\mathbf{\hat{S}}_{0}^{\theta}(\mathcal{M}_t)\mathbf{\bar{Q}}_{t-1}}{\mathbf{\hat{S}}_{0}^{\theta}(\mathcal{M}_t)\bar{\mathbf{Q}}_{t}^{{\color{VioletRed} G_{t-1}} \to {\color{BlueGreen} G_t}} \mathbf{S}_{t}^T}\right) \mathrm{\; with\;} \mathbf{Q}_{t}, \mathbf{\bar{Q}}_{t-1}, \bar{\mathbf{Q}}_{t}^{{\color{VioletRed} G_{t-1}} \to {\color{BlueGreen} G_t}} \mathrm{\; in\; Corollary~\ref{coro:s}\;} \nonumber$
    \State $\mathbf{F}_{t-1} \leftarrow \mathrm{follow \; \text{\citet{jiao2023crystal}}}, \mathbf{A}_{t-1} \leftarrow \mathrm{follow \; \text{\citet{austin2021structured, levy2025symmcd}}}$
    \EndFor
    \State
    \Return Generated crystal unit cell: $\mathbf{k}_{0}$, $G_0$, and replicating $\mathbf{F}_{0}, \mathbf{A}_{0}$ via $\mathbf{S}_{0}$. 
\end{algorithmic}
\label{alg:sampleSbCD}
\end{algorithm}

\section{Additional results}
\label{app:addres}
\subsection{Symmetry-breaking time window} 
Table~\ref{tab:resw} presents the full results of the ablation study on the symmetry-breaking time window, partially illustrated in Figure~\ref{fig:wsize}. As noted, SbCD variants with a fixed time window achieve high structural validity, indicating that such a schedule effectively preserves physically plausible atomic configurations. However, their property statistics and S.U.N. scores are significantly worse than the results in Table~\ref{tab:res}, obtained using the theoretically derived posterior symmetry-breaking time window. This suggests that while a fixed schedule improves structural validity, the adaptive posterior schedule provides a better trade-off between structural plausibility and overall generation quality.

\begin{table}[h]
\centering
\caption{Ablation results with symmetry-breaking time window $w=1$.}
\label{tab:resw}
\resizebox{.7\linewidth}{!}{
\begin{tabular}{lcc cc| cc| cc}
\toprule
\multirow{2}{*}{Method} & \multicolumn{2}{c}{\textbf{Validity} (\%)($\uparrow$)} & \multicolumn{2}{c}{\textbf{Coverage} (\%)($\uparrow$)} & \multicolumn{2}{c}{\textbf{Property statistics} ($\downarrow$)} & \multicolumn{2}{c}{\textbf{CHGNet} $E^{\text{hull}} < 0$ (\%)($\uparrow$)}\\
\cmidrule(lr){2-5} \cmidrule(lr){6-7} \cmidrule(lr){8-9}
 & $\mathrm{Struct.}$ & $\mathrm{Comp.}$ & $\mathrm{Recall}$ & $\mathrm{Precision}$ & $\mathrm{d_{\rho}}$ &  $\#\; \mathrm{elem.}$ & $\mathrm{Stable}$ & $\mathrm{S.U.N}$ \\
\midrule
SbCD & $94.27$ & $81.48$ & $99.26$ & $99.01$ & $1.66$ & $0.18$ & $7.38$ & $5.25$\\
$\text{SbCD}^{\sharp}$ & $98.13$ & $ 84.00$ & $93.38$ & $98.89$ & $1.3$ & $0.21$ & $7.44$ & $4.74$\\
\bottomrule
\end{tabular}
}
\end{table}

\subsection{Statistical significance} 
Due to the high computational cost of validation, particularly for evaluating the stability of sampled crystals, we note that several baselines, including FlowMM, DiffCSP, DiffCSP++, CDVAE, SGFM, etc., report results from a single evaluation over 10000 samples. Using this large sample size, we evaluate SbCD with {three models trained independently with different random seeds. For each model, we perform three sampling runs.} Across these evaluations in Table~\ref{tab:stat}, we observe that SbCD performs consistently on { each model, suggesting that its results are robust to sampling variance. The model with seed 1000 yields slightly better results and outperforms the state-of-the-art baseline CrysLLMGen \citep{khastagir2025llm} on both metastability and M.S.U.N metrics.}
\begin{table}[h]
\centering
\caption{ Statistical robustness of SbCD across three models independently trained on MP-20.}
\label{tab:stat}
\resizebox{.9\linewidth}{!}{
\begin{tabular}{lcc cc| ccc| cc | cc}
\toprule
\multirow{2}{*}{ Seed} & \multicolumn{2}{c}{\textbf{Validity} (\%)($\uparrow$)} & \multicolumn{2}{c}{\textbf{Coverage} (\%)($\uparrow$)} & \multicolumn{3}{c}{\textbf{Property statistics} ($\downarrow$)} & \multicolumn{2}{c}{\textbf{CHGNet} $E^{\text{hull}} < 0$ (\%)($\uparrow$)} & \multicolumn{2}{c}{ \textbf{CHGNet} $E^{\text{hull}} < 0.1$ (\%)($\uparrow$)} \\
\cmidrule(lr){2-5} \cmidrule(lr){6-8} \cmidrule(lr){9-10} \cmidrule(lr){11-12}
 & $\mathrm{Struct.}$ & $\mathrm{Comp.}$ & $\mathrm{Recall}$ & $\mathrm{Precision}$ & $\mathrm{d_{\rho}}$ &  $\#\; \mathrm{elem.}$ & $\mathrm{d_{sg}}$ & $\mathrm{Stable}$ & $\mathrm{S.U.N}$  &  $\mathrm{MetaStable}$ &  $\mathrm{M.S.U.N}$ \\
\midrule
  42 & $89.81\,\text{\tiny{($\pm$0.02)}}$ & $90.44\,\text{\tiny{($\pm$0.21)}}$ & $99.48\,\text{\tiny{($\pm$0.07)}}$ & $98.18\,\text{\tiny{($\pm$0.46)}}$ & $0.1910\,\text{\tiny{($\pm$0.0123)}}$ & $0.0395\,\text{\tiny{($\pm$0.0086)}}$ & $0.3168\,\text{\tiny{($\pm$0.0114)}}$ & $11.273\,\text{\tiny{($\pm$0.012)}}$ & $7.207\,\text{\tiny{($\pm$0.098)}}$ &  $59.743\,\text{\tiny{($\pm$1.443)}}$ &   $34.003\,\text{\tiny{($\pm$0.577)}}$\\
  1000 &   $94.79\,\text{\tiny{($\pm$0.17)}}$  &  $88.29\,\text{\tiny{($\pm$0.03)}}$  &   $99.45\,\text{\tiny{($\pm$0.10)}}$ &   $99.29\,\text{\tiny{($\pm$0.03)}}$ &   $0.1116\,\text{\tiny{($\pm$0.0226)}}$ &    $0.0522\,\text{\tiny{($\pm$0.0144)}}$&    $0.3132\,\text{\tiny{($\pm$0.0004)}}$ &   $12.547\,\text{\tiny{($\pm$0.150)}}$ &   $7.893\,\text{\tiny{($\pm$0.081)}}$&   $64.427\,\text{\tiny{($\pm$0.375)}}$ &  $36.290\,\text{\tiny{($\pm$0.069)}}$ \\
   12345 &  $89.38\,\text{\tiny{($\pm$0.05)}}$ &  $89.53\,\text{\tiny{($\pm$0.15)}}$  &   $99.45\,\text{\tiny{($\pm$0.07)}}$ &    $98.58\,\text{\tiny{($\pm$0.04)}}$ &   $0.1969\,\text{\tiny{($\pm$0.0298)}}$ &   $0.0194\,\text{\tiny{($\pm$0.0044)}}$ &   $0.3069\,\text{\tiny{($\pm$0.0013)}}$&  $11.820\,\text{\tiny{($\pm$0.087)}}$ &  $7.233\,\text{\tiny{($\pm$0.116)}}$ &  $58.687\,\text{\tiny{($\pm$0.317)}}$ &  $32.923\,\text{\tiny{($\pm$0.462)}}$ \\
   CrysLLMGen (7B) &    $99.94$ &   $93.55$ &    $98.52$ &  $99.84$ &   $0.972$ &  $0.272$  &  $-$ &   $16.79$ &  $9.21$ &   $62.02$ &  $35.94$\\ 
\bottomrule
\end{tabular}
}
\end{table}

\paragraph{Figure~\ref{fig:stable} clarification.} Due to the high computational cost of structure relaxation and energy-above-hull evaluation, we conduct the ablation study on 3,000 generated crystalline materials.

\subsection{Number of function evaluations (NFEs)} 
We compare SbCD and SymmCD under different numbers of function evaluations (NFEs) during sampling on the MP-20 benchmark. Since pretrained SymmCD weights are not publicly available, we retrain SymmCD using the training and evaluation protocol of \citet{levy2025symmcd} to ensure a fair comparison. We evaluate structural validity ($\mathrm{Struct.}$), compositional validity ($\mathrm{Comp.}$), and their product ($\mathrm{Validity}$) over 3,000 generated crystals. The results are summarized in Figure~\ref{fig:speed}, with the corresponding numerical values reported in Tables~\ref{tab:nfestruct},~\ref{tab:nfecomp}, and~\ref{tab:nfevalid}. We observe that, in the low-NFE regime, SbCD generates significantly more valid crystal structures than its symmetry-preserving counterpart, SymmCD. This demonstrates the benefit of learning space groups as part of crystal generative modeling and suggests the potential of SbCD for efficient few-step crystal generation.

\begin{table}[h]
\centering
\caption{Structure Validity}
\label{tab:nfestruct}
\resizebox{.4\linewidth}{!}{
\begin{tabular}{l ccccc}
\toprule
\multirow{2}{*}{Method} & \multicolumn{5}{c}{\textbf{NFEs}}\\
\cmidrule(lr){2-6}
 & $50$ & $100$ & $250$ & $500$ & $1000$ \\
\midrule
$\text{SymmCD}^{\ddag}$ & $31.83$ & $63.13$ & $80.53$ & $80.93$ & $87.27$ \\
SbCD & \cellcolor{Green!30} $85.00$ & \cellcolor{Green!30} $87.17$ & \cellcolor{Green!30} $89.73$ & \cellcolor{Green!30} $91.67$ & \cellcolor{Green!30} $89.63$ \\
\bottomrule
\end{tabular}
}
\end{table}
\begin{table}[h]
\centering
\caption{Composition Validity}
\label{tab:nfecomp}
\resizebox{.4\linewidth}{!}{
\begin{tabular}{l ccccc}
\toprule
\multirow{2}{*}{Method} & \multicolumn{5}{c}{\textbf{NFEs}}\\
\cmidrule(lr){2-6}
 & $50$ & $100$ & $250$ & $500$ & $1000$ \\
\midrule
$\text{SymmCD}^{\ddag}$ & $44.93$ & $70.30$ & \cellcolor{Green!30} $82.83$ & $84.70$ & $85.40$ \\
SbCD & \cellcolor{Green!30} $76.74$ & \cellcolor{Green!30} $78.33$ & $81.43$ & \cellcolor{Green!30} $90.60$ & \cellcolor{Green!30} $89.73$ \\
\bottomrule
\end{tabular}
}
\end{table}
\begin{table}[h]
\centering
\caption{Validity}
\label{tab:nfevalid}
\resizebox{.4\linewidth}{!}{
\begin{tabular}{l ccccc}
\toprule
\multirow{2}{*}{Method} & \multicolumn{5}{c}{\textbf{NFEs}}\\
\cmidrule(lr){2-6}
 & $50$ & $100$ & $250$ & $500$ & $1000$ \\
\midrule
$\text{SymmCD}^{\ddag}$ & $25.03$ & $49.97$ & $66.63$ & $68.67$ & $74.53$ \\
SbCD & \cellcolor{Green!30} $65.23$ & \cellcolor{Green!30} $68.77$ & \cellcolor{Green!30} $73.73$ & \cellcolor{Green!30} $83.43$ & \cellcolor{Green!30} $80.30$ \\
\bottomrule
\end{tabular}
}
\end{table}

\subsection{Analysis of space-group distributions}

\begin{figure}[h]
\begin{center}
\includegraphics[width=1.\linewidth]{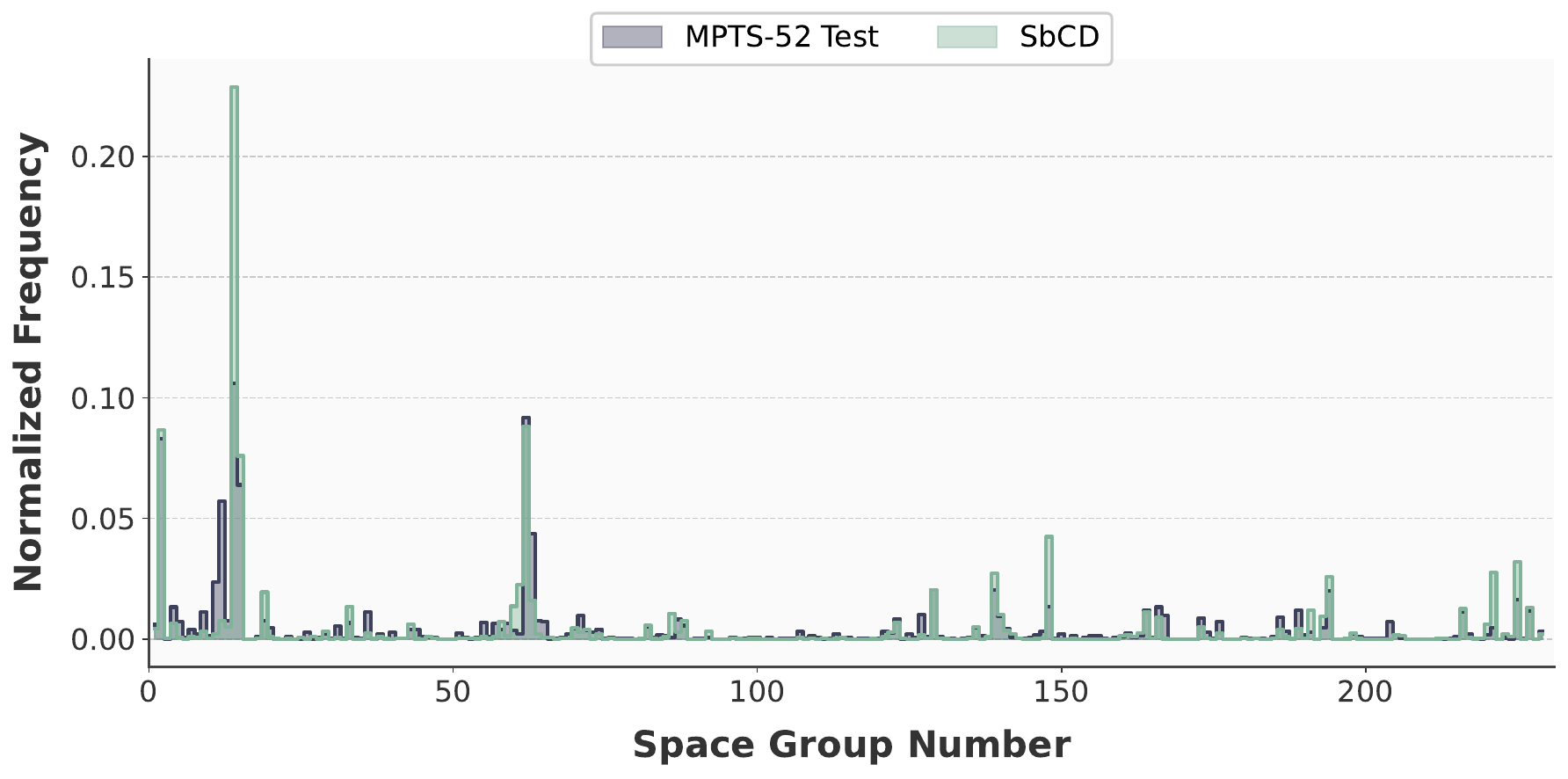}
\end{center}
\caption{Space group distributions of the MPTS-52 test set and crystals sampled by SbCD.}
\label{fig:sg_mpts52}
\end{figure}

\begin{figure}[h]
\begin{center}
\includegraphics[width=1.\linewidth]{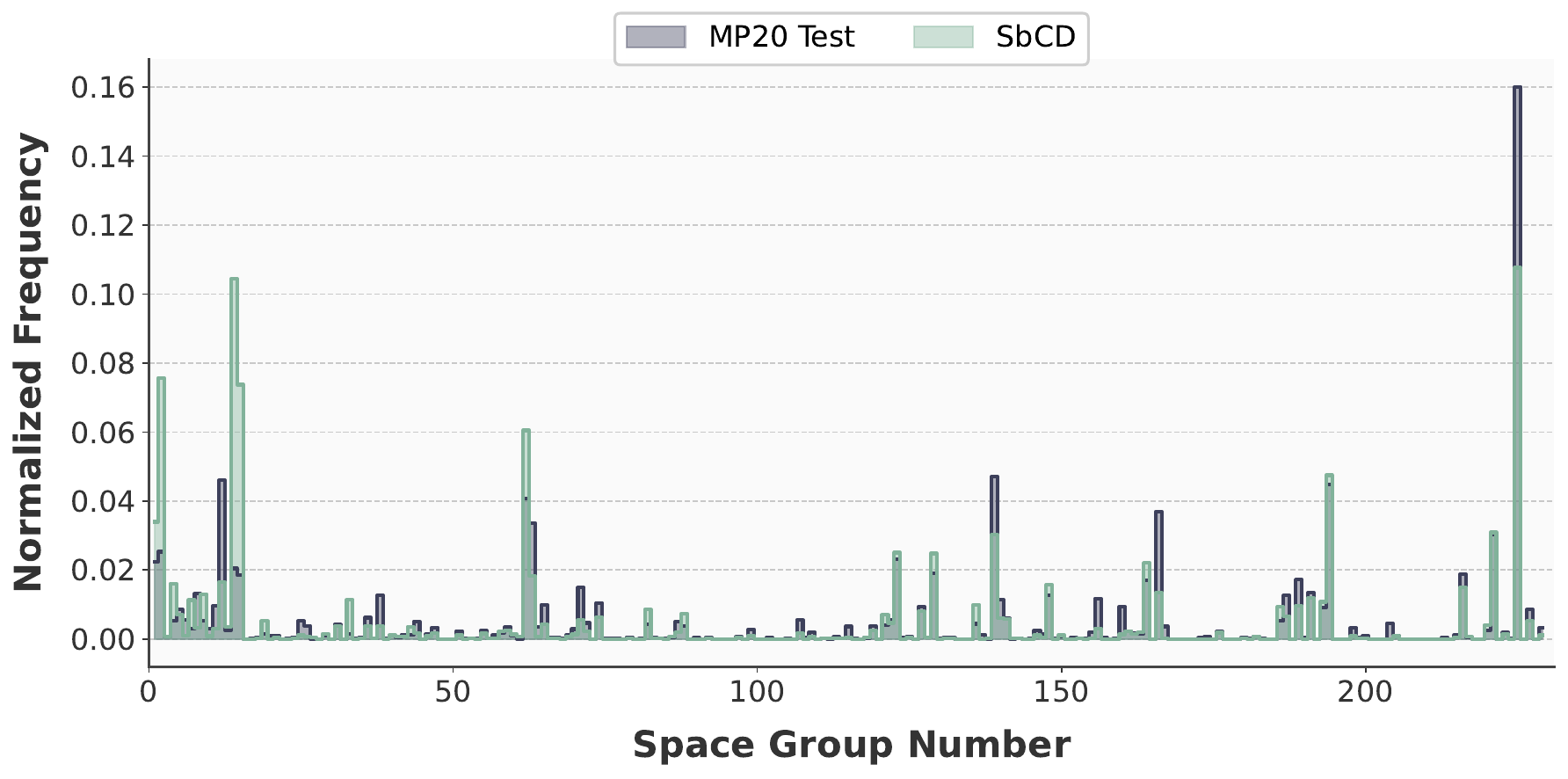}
\end{center}
\caption{Space group distributions of the MP-20 test set and crystals sampled by SbCD.}
\label{fig:sg_mp20}
\end{figure}

In Figure~\ref{fig:dsg}, we observe that DiffCSP++ and SymmCD directly condition on space groups sampled from the training data distribution, which naturally encourages generated samples to have space-group distributions close to the data distribution. SbCD variants achieve comparable values $d_{sg}$ to DiffCSP++ and SymmCD, while outperforming DiffCSP, FlowMM, and CDVAE, which do not explicitly model or utilize crystal symmetries during inference. Furthermore, Figures~\ref{fig:sg_mp20} and~\ref{fig:sg_mpts52} show that SbCD variants generate structures spanning a broad range of space groups observed in the data distribution. We also provide some examples of SbCD-generated crystal structures across different space groups in Figures~\ref{fig:samples_sbcd},~\ref{fig:samples}, and~\ref{fig:samples_mpts52}.

We additionally evaluate SbCD's performance on the top 10 most common space groups among generated structures in Table~\ref{tab:top10}. Interestingly, $\text{SbCD}^{\sharp}$, which adopts the site-symmetry representation from \citet{levy2025symmcd}, achieves higher structural validity on the top 10 most common space groups compared with its average validity across all space groups. In contrast, $\text{SbCD}^{\diamondsuit}$ and SbCD, which use our proposed site-symmetry representation, generate more valid structures across less frequent space groups, which compensates for their slightly lower structural validity on the most common space groups. Most property metrics worsen for the top 10 most common space groups, due to the limited coverage of these space groups in the data distribution. Nevertheless, we observe that structures generated from the most common space groups tend to exhibit higher stability.

\begin{table}[h]
\centering
\caption{De novo generation results averaged across the 10 most representative space groups in crystalline materials generated on the MP-20 dataset.}
\label{tab:top10}
\resizebox{1.\linewidth}{!}{
\begin{tabular}{lcc cc| ccc| cc}
\toprule
\multirow{2}{*}{Method} & \multicolumn{2}{c}{\textbf{Validity} (\%)($\uparrow$)} & \multicolumn{2}{c}{\textbf{Coverage} (\%)($\uparrow$)} & \multicolumn{3}{c}{\textbf{Property statistics} ($\downarrow$)} & \multicolumn{2}{c}{\textbf{CHGNet} $E^{\text{hull}} < 0$ (\%)($\uparrow$)}\\
\cmidrule(lr){2-5} \cmidrule(lr){6-8} \cmidrule(lr){9-10}
 & $\mathrm{Struct.}$ & $\mathrm{Comp.}$ & $\mathrm{Recall}$ & $\mathrm{Precision}$ & $\mathrm{d_{\rho}}$ &  $\#\; \mathrm{elem.}$ & $\mathrm{d_{sg}}$ & $\mathrm{Stable}$ & $\mathrm{S.U.N}$ \\
\midrule
$\text{SbCD}^{\sharp}$ & $89.81$ & $88.32$ & $99.51$ & $97.31$ & $0.09$ & $0.04$ & $0.22$ & $10.74$ & $7.31$\\
$\text{SbCD}^{\sharp}$ (10 SGs) & $97.98$ & $89.46$ & $96.97$ & $99.45$ & $0.13$ & $0.045$ & $0.51$ & $12.84$ & $9.60$\\
\midrule
$\text{SbCD}^{\diamondsuit}$& $88.25$ & $87.47$ & $99.44$ & $97.71$ &  $0.16$ & $0.07$ & $0.28$ &  $10.23$ & $7.67$\\
$\text{SbCD}^{\diamondsuit}$ (10 SGs) & $83.78$ & $88.46$ & $94.66$ & $96.63$ & $0.37$ & $0.12$ & $0.55$ & $10.33$ & $8.17$\\
\midrule
SbCD & $89.80$ & $90.32$ & $99.44$ & $97.91$ & $0.18$ & $0.03$ & $0.32$ & $11.28$ & $7.15$\\
SbCD (10 SGs) & $86.60$ & $91.02$ & $97.71$ & $98.51$ & $0.23$ & $0.042$ & $0.54$ & $11.76$ & $7.79$\\
\bottomrule
\end{tabular}
}
\end{table}

\subsection{Histogram of the energy above the convex hull $E_{hull}$}

\begin{figure}[h]
\begin{center}
\includegraphics[width=1.\linewidth]{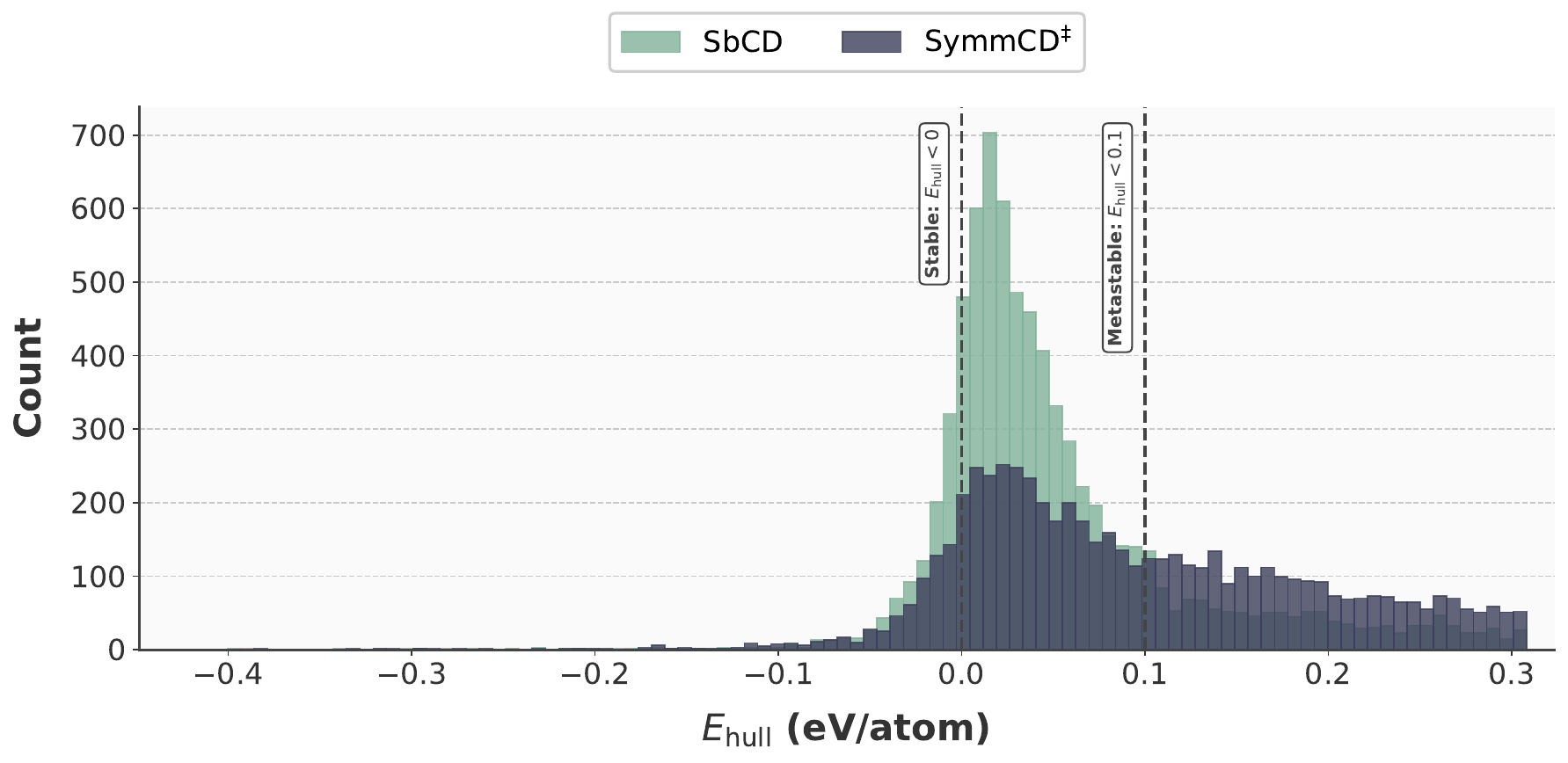}
\end{center}
\caption{Comparison of the $E_{\mathrm{hull}}$ distributions on the MP-20 dataset.}
\label{fig:ehull_mp20}
\end{figure}

\begin{figure}[h]
\begin{center}
\includegraphics[width=1.\linewidth]{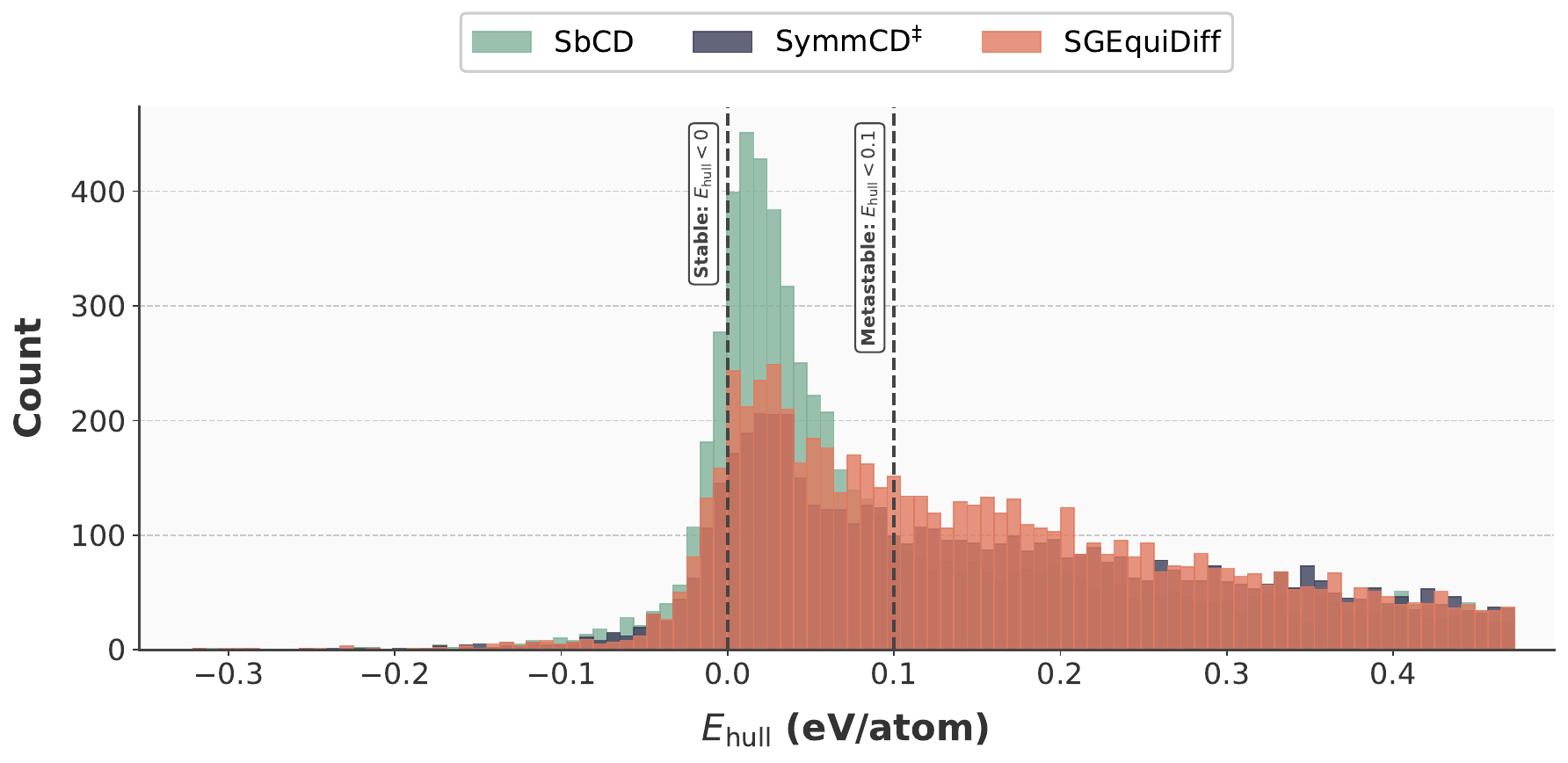}
\end{center}
\caption{Comparison of the $E_{\mathrm{hull}}$ distributions on the MPTS-52 dataset.}
\label{fig:ehull_mpts52}
\end{figure}

Figures~\ref{fig:ehull_mp20} and~\ref{fig:ehull_mpts52} visualize the distributions of $E_{\mathrm{hull}}$ computed after relaxation with CHGNET~\citep{deng2023chgnet} for different methods. As shown, SbCD performs competitively, generating a significantly higher proportion of structures that are thermodynamically stable ($E_{\mathrm{hull}} < 0$) or metastable ($E_{\mathrm{hull}} < 0.1$).

\begin{algorithm}[h]
\caption{Training of SbCD}
\label{alg:trainSbCD}
\begin{algorithmic}[1]
    \Require Model $\theta$, transition-rate function $\lambda$, dataset of crystal asymmetric units $\mathcal{D}$
    
    \While{not converged}
        \State Sample a clean asymmetric unit and diffusion timestep
        \[
            \mathcal{M}_0 = (\mathbf{F}_0, \mathbf{A}_0, \mathbf{k}_0, \mathbf{S}_0, {\color{VioletRed} G_0}) \sim \mathcal{D},
            \qquad
            t \sim \mathcal{U}(0,T)
        \]
        \State Sample $\tau$ and $w$ via Corollary~\ref{cor:tau}

        \begin{align}
        &\tau = -\frac{1}{\lambda}\log\left(1 - u\left(1 - e^{-\lambda t}\right)\right), \quad u \sim \mathcal{U}(0,1) \nonumber\\
        &w_{\mathbf{S}} = w_\mathbf{k}= t -\tau \nonumber
        \end{align}
        \State Sample the space-group state at time $t$ via Proposition~\ref{prop:jump}
        \[
           {\color{BlueGreen} G_t}  \sim q_{t|0}({\color{BlueGreen} G_t}|{\color{VioletRed} G_0}) \rightarrow  G_{t+1} \sim\Pi_t({\color{BlueGreen} G_t}, G_{t+1})
        \]
        
        \State Corrupt $\mathbf{k}_0$ using the symmetry-breaking continuous kernel via Proposition~\ref{prop:k}

        \begin{align}
            \mathbf{k}_t = \mathfrak{\color{VioletRed}m}\odot&\left(\sqrt{\bar{\alpha_t}}\mathbf{k}_0+\sqrt{1-\bar{\alpha_t}}{\color{VioletRed}\boldsymbol{\epsilon}}\right) +{\color{VioletRed}\mathfrak{m_b}} + {\color{BlueGreen}\mathfrak{m}}\odot{\color{VioletRed}\mathfrak{m_b}}\sqrt{\frac{\bar{\alpha}_t}{\bar{\alpha}_{t-w_{\mathbf{k}}}}}\nonumber \\
            &-\left({\color{VioletRed}\mathfrak{m}}\odot\sqrt{1-\frac{\bar{\alpha}_t}{\bar{\alpha}_{t-w_{\mathbf{k}}}}}{\color{VioletRed}\boldsymbol{\epsilon}} + {\color{VioletRed}\mathfrak{m_b}}\right) + \left({\color{BlueGreen}\mathfrak{m}}\odot\sqrt{1-\frac{\bar{\alpha}_t}{\bar{\alpha}_{t-w_{\mathbf{k}}}}}{\color{BlueGreen}\boldsymbol{\epsilon}} + {\color{BlueGreen}\mathfrak{m_b}} \right) \nonumber \\
            \mathrm{with\;} {\color{VioletRed}\epsilon}, {\color{BlueGreen}\epsilon} \sim &\mathcal{N}(0,I) \nonumber
        \end{align}
        
        \State Corrupt $\mathbf{S}_0$ using the symmetry-breaking categorical kernel in Proposition~\ref{prop:marginal_s}.

        \begin{align}
            & \mathbf{S}_t = \mathrm{Cat}(\mathbf{S}_t; \mathbf{S}_0\bar{\mathbf{Q}}_{t}^{{\color{VioletRed}G_0} \to {\color{BlueGreen}G_t}}), \quad \bar{\mathbf{Q}}_{t}^{{\color{VioletRed}G_0} \to {\color{BlueGreen}G_t}} = \bar{\alpha}_tI + (1-\bar{\alpha}_t)\mathbf{1}(\boldsymbol{\pi}_t)^T \nonumber\\
            &\mathrm{with \;}\boldsymbol{\pi}_t =\frac{\bar{\alpha}_t(1-\bar{\alpha}_{t- w_{\mathbf{S}}})\mathfrak{\color{VioletRed}g} + (\bar{\alpha}_{t- w_{\mathbf{S}}}-\bar{\alpha}_t)\mathfrak{\color{BlueGreen}g}}{\bar{\alpha}_{t- w_{\mathbf{S}}} (1- \bar{\alpha}_t)} \nonumber
        \end{align}
        
        \State Corrupt $\mathbf{F}_0$ with wrapped Gaussian noises \citep{jiao2023crystal}, yielding $\mathbf{F}_t$
        
        \State Corrupt $\mathbf{A}_0$ with categorical noises \citep{austin2021structured, levy2025symmcd}, yielding $\mathbf{A}_t$.
        
        \State Evaluate the training losses on the corrupted samples $\mathcal{M}_t, \mathcal{M}_{t+1}$, defined as:
        \begin{align}
            &\mathcal{M}_t = (\mathbf{F}_t, \mathbf{A}_t, \mathbf{k}_t, \mathbf{S}_t, {\color{BlueGreen} G_t}) \nonumber\\
            &\mathcal{M}_{t+1} = (\mathbf{F}_{t+1}, \mathbf{A}_{t+1}, \mathbf{k}_{t+1}, \mathbf{S}_{t+1}, G_{t+1})\; \mathrm{by\; repeating\; steps\; 5,6,7,8} \nonumber
        \end{align}
        \begin{align}
            &\mathcal{L}_{\mathbf{k}} \leftarrow \mathrm{MSE}(\mathbf{\hat{k}}_0^{\theta}(\mathcal{M}_t), \mathbf{k}_0), \; \mathcal{L}_{\mathbf{S}} \leftarrow \mathrm{CE}(\mathbf{\hat{S}}_0^{\theta}(\mathcal{M}_t), \mathbf{S}_0), \; \mathcal{L}_{\mathbf{A}} \leftarrow \mathrm{CE}(\mathbf{\hat{A}}_0^{\theta}(\mathcal{M}_t), \mathbf{A}_0)\nonumber\\
            &\mathcal{L}_G \leftarrow \mathrm{CE}(\hat{G}_0^{\theta}(\mathcal{M}_t), {\color{VioletRed} G_0}) \nonumber\\
             &\mathcal{L}_{\mathrm{jump}} \leftarrow -\hat{\mathbf{\Lambda}}_t^{\theta} ({\color{BlueGreen} G_t},{\color{BlueGreen} G_t}) + \mathbf{\Lambda}_t ({\color{BlueGreen} G_t},{\color{BlueGreen} G_t}) \log (\hat{\mathbf{\Lambda}}_t^{\theta} (G_{t+1}, {\color{BlueGreen} G_t})), \mathrm{\quad Lemma~\ref{lemma:jump}} \nonumber\\
             &\qquad \mathrm{where\;} \hat{\mathbf{\Lambda}}_t^{\theta} ({\color{BlueGreen} G_t},{\color{BlueGreen} G_t})=-\sum_{G \neq {\color{BlueGreen} G_t}} \hat{\mathbf{\Lambda}}_t^{\theta} ({\color{BlueGreen} G_t},G)\nonumber\\
             &\qquad \mathrm{with\; } \hat{\mathbf{\Lambda}}_t^{\theta} ({\color{BlueGreen} G_t},G) = \mathbf{\Lambda}_t (G,{\color{BlueGreen} G_t}) \sum_{{\color{VioletRed} G_0}} \frac{q_{t|0}(G|{\color{VioletRed} G_0})}{q_{t|0}({\color{BlueGreen} G_t}|{\color{VioletRed} G_0})}p_{0|t}^{\theta}({\color{VioletRed} G_0}|\mathcal{M}_t) \nonumber\\
             &\qquad\mathrm{and\; } \hat{\mathbf{\Lambda}}_t^{\theta} (G_{t+1},{\color{BlueGreen} G_t}) = \mathbf{\Lambda}_t ({\color{BlueGreen} G_t},G_{t+1}) \sum_{{\color{VioletRed} G_0}} \frac{q_{t|0}( {\color{BlueGreen} G_t}|{\color{VioletRed} G_0})}{q_{t|0}(G_{t+1}|{\color{VioletRed} G_0})}p_{0|t}^{\theta}({\color{VioletRed} G_0}|\mathcal{M}_{t+1}) \nonumber\\
             &\mathcal{L}_{\mathbf{F}} \leftarrow \mathrm{follow \; \citep{jiao2023crystal}} \nonumber
        \end{align}
        
        \State Update $\theta$ by descending the full objective weighted by $\{l_{\mathbf{k}}, l_{\mathbf{S}}, l_{\mathbf{A}}, l_G, l_{\mathbf{F}}, l_{\mathrm{jump}} \}$
        \[
            \mathcal{L}
            =
            l_{\mathbf{k}} \mathcal{L}_{\mathbf{k}}
            + l_{\mathbf{S}} \mathcal{L}_{\mathbf{S}}
            + l_{\mathbf{A}} \mathcal{L}_{\mathbf{A}}
            + l_G \mathcal{L}_G
            + l_{\mathrm{jump}} \mathcal{L}_{\mathrm{jump}}
            + l_{\mathbf{F}} \mathcal{L}_{\mathbf{F}}.
        \]
    \EndWhile
\end{algorithmic}
\end{algorithm}

\end{document}